\documentclass[sigconf]{acmart}
\AtBeginDocument{%
  }

\setcopyright{acmlicensed}
\copyrightyear{2026}
\acmYear{2026}
\acmDOI{XXXXXXX.XXXXXXX}
\acmConference[ACM FAccT '26]{ACM Conference on Fairness, Accountability, and Transparency}{June 25--28,
  2026}{Montréal, Canada}
\acmISBN{978-1-4503-XXXX-X/2026/06}

\usepackage{enumitem}

\usepackage{amsfonts}
\usepackage{bm}
\usepackage{xfrac}
\usepackage{cleveref}
\usepackage{algorithm,algorithmic}
\usepackage{multirow,multicol}
\usepackage{subfig,wrapfig}
\newcommand{\myref}[2]{\ref{#1}\subref{#2}}

\setboolean{ALC@noend}{true}
\usepackage{threeparttable}
\usepackage{rotating}
\usepackage{booktabs}

\def\ie{i.e.}
\def\eg{e.g.}
\def\aka{a.k.a.}

\def\etal{\emph{et al.}}

\newcommand{\tabincell}[2]{\begin{tabular}{@{}#1@{}}#2\end{tabular}}
\newcommand{\topelement}[1]{\small{#1}}

\DeclareMathOperator*{\argmax}{argmax}
\DeclareMathOperator*{\argmin}{argmin}

\newcommand{\defineq}{\triangleq}
\def\mvrho{\mathbf{wv}_\rho}

\def\rhomv{\mathbf{wv}}
\def\mvpi{\rhomv_\pi}
\def\incons{\phi_\rho}

\def\xneg{\mathop{\breve{\bm{x}}}}
\def\xpos{\mathop{\bm{a}}}
\def\xqtb{\mathop{\tilde{\bm{a}}}}

\def\probE{\mathbb{E}}

\def\insx{\bm{x}}

\def\justL{\ell_\text{bias}}
\def\justLf{\hat{\mathcal{L}}_\text{bias}}
\def\justLt{\mathcal{L}_\text{bias}}

\def\justAcc{\ell_\text{err}}

\def\justAccF{\hat{\mathcal{L}}_\text{err}}
\def\justAccT{\mathcal{L}_\text{err}}

\def\propfull{ensemble pruning via improving accuracy and fairness concurrently}
\def\propabbr{\emph{EPAF}}
\def\poepfull{Pareto optimal \propfull}
\def\poepabbr{\emph{POAF}}
\def\greedy{\propabbr{}\emph{-C}}
\def\ddismi{\propabbr{}\emph{-D}}

\def\justObjF{\hat{\mathcal{L}}}
\def\justObjT{\mathcal{L}}

\usepackage[acronym,shortcuts,xindy]{glossaries}
\renewcommand{\gls}{\ac*}

\newacronym{ml}{ML}{machine learning}
\newacronym{dl}{DL}{deep learning}
\newacronym{ai}{AI}{artificial intelligence}

\newacronym{dp}{DP}{demographic parity}
\newacronym{eo}{EO}{equalised odds}
\newacronym{eopp}{EOpp}{equality of opportunity}
\newacronym{pqp}{PP}{predictive parity}
\newacronym{cff}{CFF}{counterfactual fairness}
\newacronym{sp}{SP}{statistical parity}
\newacronym{dr}{\emph{DR}}{discriminative risk}
\newacronym{sa}{SA}{sensitive attribute}
\newacronym{hd}{HD}{Hausdorff distance}
\newacronym{sota}{SOTA}{state-of-the-art}
\newacronym{dt}{DT}{decision tree}

\newcommand{\topequation}{%
  \setlength\abovedisplayskip{1pt}
  \setlength\belowdisplayskip{1pt}
}

\newcommand{\toptabcol}{%
  \renewcommand\tabcolsep{2.74pt}%
}

\graphicspath{{dr_figs/}{dr_newfig/}}

\begin{document}

\title{Improving Fairness with Ensemble Combination: Margin-Dependent Bounds}
\author{Yijun Bian}
\email{yibi@di.ku.dk}
\orcid{0000-0002-5926-7100}
\affiliation{%
  \institution{University of Copenhagen}
  \city{Copenhagen}
  \country{Denmark}
}



\begin{abstract}
The concern about hidden discrimination in \gls{ml} models is growing, as their widespread real-world applications increasingly impact human lives. 
Various techniques, including commonly used group fairness measures and several fairness-aware ensemble-based methods, have been developed to enhance fairness. 
However, existing fairness measures typically focus on only one aspect---either group or individual fairness, and the compatibility difficulty among these measures indicates a possibility of remaining biases even when one of them is satisfied. 
Moreover, existing mechanisms to boost fairness usually present empirical results to show validity, yet few of them discuss whether fairness can be boosted with certain theoretical guarantees. 
To address these issues, we propose a fairness quality measure named `\emph{discriminative risk}' by only perturbing protected attributes in instances, to express both individual and group fairness aspects. 
Furthermore, we investigate its properties and establish the first- and second-order oracle bounds and their relaxations, which show that fairness is possibly improved via ensemble combination with margin-dependent bounds. 
The analysis is suitable for both binary and multi-class classification. 
A few ensemble pruning methods are also proposed to utilise our proposed measure and obtain both accurate and fair sub-ensembles; comprehensive experiments are conducted to evaluate the effectiveness of the proposed fairness measure and pruning methods.\looseness=-1 
\end{abstract}

\maketitle

\renewcommand\justLf{\hat{\bm{\mathit{L}}}_\text{bias}}
\renewcommand\justLt{\bm{\mathit{L}}_\text{bias}}
\renewcommand\justAccT{\bm{\mathit{L}}_\text{err}}
\renewcommand\justAccF{\hat{\bm{\mathit{L}}}_\text{err}}
\renewcommand{\justObjF}{\hat{\bm{\mathit{L}}}}
\renewcommand{\justObjT}{\bm{\mathit{L}}}

\section{Introduction}
\label{sec:intro}

Machine learning (ML) is increasingly applied in sensitive decision-making domains such as recruitment, jurisdiction, and credit evaluation. 
As \gls{ml} becomes more pervasive in real-world scenarios nowadays, concerns about the fairness and reliability of \gls{ml} models have emerged and grown. 
Discriminative models may perpetuate or even exacerbate improper human prejudices, negatively impacting both model performance and societal equity. 
Unfairness in \gls{ml} models identified in the literature primarily stems from two causes: data biases and algorithmic biases \citep{verma2018fairness}. 
Data biases mainly arise from inaccurate device measurements, erroneous reports, or historically biased human decisions, misleading \gls{ml} models to replicate them. 
Missing data can also distort the distribution of the dataset from the target population and introduce further biases. 
Algorithmic biases occur even with purely clean data. 
These biases may arise from proxy attributes for sensitive variables or from tendentious objectives of the learning algorithms themselves. 
For example, minimising aggregated prediction errors may inadvertently favour the privileged group over unprivileged minorities.

Numerous mechanisms have been proposed to mitigate biases and enhance fairness in \gls{ml} models, typically categorised as pre-processing, inprocessing, and post-processing mechanisms. 
Pre- and post-processing mechanisms normally function by manipulating input or output, while inprocessing mechanisms incorporate fairness constraints into training procedures or algorithmic objectives. 
Determining the superior approach is challenging as results vary based on applied fairness measures, datasets, and even the training-test split handling \citep{friedler2019comparative,dwork2018decoupled}. 
Various fairness measures have been developed to facilitate the design of fair ML models, such as group and individual fairness measures. 
Group fairness emphasises statistical/demographic equality among groups defined by protected or \glspl{sa}, including but not limited to \gls{dp} \citep{feldman2015certifying,gajane2017formalizing,jiang2020wasserstein}, \gls{eopp} \citep{hardt2016equality,gajane2017formalizing,haas2019price}, and \gls{pqp} \citep{chouldechova2017fair,verma2018fairness}. 
In contrast, individual fairness follows the principle that ``similar individuals should be evaluated or treated similarly,'' with similarity measured by specific distances between individuals \citep{joseph2016fairness,dwork2012fairness}. 
However, these measures often conflict, meaning that unfair outcomes may persist even when one criterion is met, such as 
the incompatibility of group fairness measures themselves and that between individual and group fairness measures \citep{barocas2023fairness,berk2021fairness,pleiss2017fairness,hardt2016equality}.

To address these challenges, we propose a novel fairness quality measure named `\emph{discriminative risk (DR)}', which reflects the discriminative degree of learners from both individual and group fairness perspectives. 
The intuition behind this is that by slightly perturbing an instance's \glspl{sa} and making the learnt model predict the perturbed data again, if the model gives a different prediction, there would exist underlying discriminative risk. 
We further investigate its properties and derive that the fairness quality of learners can possibly benefit from ensemble combination with margin-dependent bounds, inspired by a cancellation-of-errors effect of combination, where combining multiple weak learners yields a more powerful learner. 
In essence, we explore the possibility of a cancellation-of-biases effect in combination, seeking to answer the question: \emph{Will combination help mitigate discrimination even in multiple biased individual classifiers?} 
Our proposed bounds regarding fairness indicate the potential existence of a positive answer that depends on the voting margins, providing insights into theoretical supports for the fact that `fairness can be improved sometimes via ensembles by boosting performance on disadvantaged groups' \citep{ko2023fair,grgic2017fairness,schweighofer2025the,claucich2025fairness}. 

Our contributions in this work are fourfold:
(1) We propose a novel fairness measure that assesses the bias level of classifiers from both individual and group fairness sides, along with its properties. 
(2) We establish first- and second-order oracle bounds, and provide their corresponding relaxations, investigating the existence of a cancellation-of-biases effect in ensemble combination from a theoretical perspective. 
(3) We develop an ensemble pruning method by integrating the proposed measure with the concepts of domination and Pareto optimality, aiming to enhance fairness with minimal negative impact on accuracy.
(4) Comprehensive experiments are conducted to validate the proposed bounds and demonstrate the effectiveness of the proposed measure and its corresponding bounds, as well as the proposed pruning method's performance.

\section{Related Work}
\label{sec:related} 
In this section, we first introduce existing techniques to mitigate bias issues in \gls{ml} models and then summarise relevant fairness-aware ensemble-based methods in turn.

\textbf{Mechanisms to enhance fairness}\hspace{1em} 
Three typical mechanisms are employed to mitigate biases and enhance fairness in \gls{ml} models: pre-processing, inprocessing, and post-processing mechanisms. 
Pre-processing mechanisms usually manipulate features or labels of instances before they are fed into the algorithm, aiming to assimilate the distributions of unprivileged groups with those of the privileged group, thus making it harder for the algorithm to distinguish between them \citep{backurs2019scalable,samadi2018price,chierichetti2017fair,calmon2017optimized,louizos2016variational,feldman2015certifying,kamiran2012data}. 
While advantageous for their flexibility across classification tasks, pre-processing mechanisms often suffer from high uncertainty in the final accuracy. 
Post-processing mechanisms adjust output scores or decouple predictions for each group \citep{menon2018cost,dwork2018decoupled,corbett2017algorithmic,hardt2016equality}. 
Though also task-agnostic, they typically yield inferior results due to their late application in the learning process, and may compromise individual fairness through differentiated treatment. 
Inprocessing mechanisms favour directly incorporating fairness constraints during the training by incorporating penalty/regularisation terms \citep{zafar2017fairness2,zafar2017fairness1,woodworth2017learning,quadrianto2017recycling,kamishima2012fairness,kamiran2010discrimination}, while some adjust these constraints in minimax or multi-objective optimisation settings \citep{agarwal2019fair,agarwal2018reductions,zemel2013learning}. 
These mechanisms enforce explicit trade-offs between accuracy and fairness in objectives, but are closely tied to the specific \gls{ml} algorithm used. 
Choosing the optimal mechanism is context-dependent, varying with fairness measures, datasets, and even the training-test split handling \citep{friedler2019comparative,dwork2018decoupled}.\looseness=-1

\paragraph{\bf Types of fairness measures}
\label{sec:relat,2}
Various fairness measures have been proposed to facilitate the design of fair \gls{ml} models, which can be generally divided into distributive and procedural fairness measures. 
Procedural fairness concerns the fairness of decision-making processes, encompassing feature-apriori fairness, feature-accuracy fairness, and feature-disparity fairness \citep{grgic2018beyond}. 
These measures depend on features and user perceptions of fairness but may still introduce hidden biases within the data. 
Distributive fairness pertains to the fairness of decision-making outcomes (predictions), including unconscious/unawareness fairness, group fairness, individual fairness, and \gls{cff}. 
As the simplest, unawareness fairness means making predictions without explicitly using any protected \glspl{sa}, though it does not prevent biases linked to associations between unprotected and protected attributes \citep{dwork2012fairness}. 
Group fairness focuses on statistical/demographic equality among groups defined by \glspl{sa}, such as \gls{dp}, \gls{eo} \citep{hardt2016equality,haas2019price,berk2021fairness,pleiss2017fairness}, \gls{eopp}, and \gls{pqp}. 
In contrast, individual fairness operates on the principle that ``similar individuals should be evaluated or treated similarly,'' where similarity is measured by some certain distance between individuals, while the specified distance also matters a lot \citep{joseph2016fairness,dwork2012fairness}. 
Besides, \gls{cff} aims to explain the sources of discrimination and qualitative equity through causal interference tools \citep{nilforoshan2022causal,kusner2017counterfactual}.\looseness=-1

However, the group fairness measures are often hardly compatible with each other, 
for example, the occurrence between \gls{eo} and \gls{dp}, or that between equalised calibration and \gls{eo}. 
Individual and group fairness, such as \gls{dp}, are also incompatible except in the case of trivial degenerate solutions. 
Moreover, three fairness criteria---independence, separation, and sufficiency---are demonstrated not to be satisfied concurrently unless in degenerate cases \citep{barocas2023fairness}. 
Furthermore, significant attention has been paid to compromising accuracy in the pursuit of higher levels of fairness \citep{friedler2019comparative,menon2018cost,corbett2017algorithmic,bechavod2017penalizing}. 
It is widely accepted that introducing fairness constraints into an optimisation problem 
likely results in reduced accuracy compared to optimising solely for accuracy. However, some researchers have recently proposed a few unique scenarios where fairness and accuracy can be simultaneously improved \citep{pessach2021improving,wick2019unlocking}.

\paragraph{\bf Fairness-aware ensemble-based methods}
One of the primary challenges in designing fair learning algorithms is the potential trade-off between fairness and accuracy. 
Recent studies have explored various fairness-enhancing techniques and their impact on ML models, including a few methods employing typical boosting mechanisms in ensemble learning, such as AdaFair \citep{iosifidis2019adafair}, FARF \citep{zhang2021farf}, and FairGBM \citep{cruz2022fairgbm}. 
For instance, FARF and AdaFair combined different fairness-related criteria into the training phase, while FairGBM transformed non-differentiable fairness constraints into a proxy inequality constraint to facilitate gradient-based optimisation. 
Furthermore, it is recognised that fairness may emerge naturally alongside ensembling due to the performance improvement in the marginalised group(s) \citep{grgic2017fairness,ko2023fair,claucich2025fairness,schweighofer2025the}. 
Despite these advancements, there is a paucity of research addressing the theoretical guarantees of these methods in enhancing fairness, with most studies relying on empirical results to demonstrate practical effectiveness.

\section{Methodology}
\label{sec:method}

In this section, we formally study the fairness properties of ensemble methods with weighted voting.

We begin by introducing the necessary notations used throughout the paper. 
We denote a dataset by $S=\{(\bm{x}_i,y_i)\}_{i=1}^n$, where instances are independent identically distributed (i.i.d.) drawn from an input/feature-output/label space $\mathcal{X}\times\mathcal{Y}$ according to an unknown distribution $\mathcal{D}$. 
The label space $\mathcal{Y}=\{1,2,...,n_c\} (n_c\geqslant 2)$ is finite and can represent binary or multi-class classification, and the feature space $\mathcal{X}$ is arbitrary. 
An instance including sensitive or protected attributes $\xpos$ is represented as $\bm{x}\defineq (\xneg,\xpos)$, and $\xqtb$ indicates a randomly perturbed version of $\xpos$. Note that $\xpos = [a_1,...,a_{n_a}]^\mathsf{T}$ allows multiple attributes, with $n_a$ as the number of \glspl{sa}, and for each attribute, $a_i\in \mathbb{Z}_+ (1 \leqslant i \leqslant n_a)$ allows binary and multiple values. A member belonging to the marginalised group(s) will be perturbed into the privileged group, while a member of the privileged group is perturbed randomly into one of the marginalised groups. 
Taking a multi-valued sensitive attribute $a_i\in\{1,2,...,n_b\}$ for example, where $a_i=1$ represents the privileged group and $n_b$ is the number of its values, let $\tilde{a}_i$ denote a random perturbation of $a_i$. Then if $a_i=1$, it can be one of $\{2,3,...,n_b\}$ with equal probability; otherwise, $\tilde{a}_i=1$. 
Multiple sensitive attributes will be handled analogously, one by one. 
Also note that non-sensitive attributes $\xneg$ may or may not include proxy attributes that are affected somehow by \glspl{sa}. 
A hypothesis in the space of hypotheses $\mathcal{F}$ is a function $f\in\mathcal{F}: \mathcal{X}\mapsto\mathcal{Y}$, also known as a classifier or model.

\subsection{Fairness quality from both individual- and group-level aspects}
\label{method:4}

To discuss the properties of fairness, it is essential to have a proper measure that accurately reflects the prediction quality of hypotheses. 
Lots of fairness measures have been proposed (see Section~\ref{sec:relat,2}), including three commonly used ones (\ie{} \gls{dp} \citep{feldman2015certifying,gajane2017formalizing}, \gls{eopp} \citep{hardt2016equality}, and \gls{pqp} \citep{chouldechova2017fair,verma2018fairness}).%
\footnote{
In this paper, these three group fairness measures of one hypothesis $f(\cdot)$ are evaluated respectively as
\begin{subequations}
\topequation%
\begin{align}
    \mathrm{DP}(f) &= \lvert
    \mathbb{P}_\mathcal{D}[ f(\bm{x})=1 \mid a=1 ] - 
    \mathbb{P}_\mathcal{D}[ f(\bm{x})=1 \mid a=0 ] 
    \rvert \,,\label{eq:group,1}\\
    \mathrm{EOpp}(f) &= \lvert
    \mathbb{P}_\mathcal{D}[ f(\bm{x})=1 \mid a=1, y=1 ] - 
    \mathbb{P}_\mathcal{D}[ f(\bm{x})=1 \mid a=0, y=1 ] 
    \rvert \,,\label{eq:group,2}\\
    \mathrm{PP}(f) &= \lvert
    \mathbb{P}_\mathcal{D}[ y=1 \mid a=1, f(\bm{x})=1 ] - 
    \mathbb{P}_\mathcal{D}[ y=1 \mid a=0, f(\bm{x})=1 ] 
    \rvert \,,\label{eq:group,3}
\end{align}
\label{eq:group,fair}%
\end{subequations}%
where $\bm{x}=(\xneg,a)$, $y$ is the true label, and $f(\bm{x})$ is the prediction. 
Note that $a=1$ and $a=0$ mean that the instance $\bm{x}$ belongs to the privileged group and marginalised groups, respectively. 
These three fairness measures are originally defined on a single binary-valued sensitive attribute only. 
} 
However, the hard compatibility among them means that each one of them focuses only on one specific aspect of fairness (either group or individual fairness). 
Therefore, to capture the discriminative degree of hypotheses from both individual and group sides, we propose a new fairness quality measure named \emph{\gls{dr}}.

Presume that the considered dataset $S\defineq\{(\bm{x}_i,y_i)\}_{i=1}^n =\{(\xneg_i,\xpos_i,$\,\\$y_i)\}_{i=1}^n$ consists of instances containing \glspl{sa}. 
Following the principle of individual fairness that ``similar individuals should be evaluated/treated similarly'',\footnote{%
Apart from various choices of the metric, there are different formulations in the literature to mathematically capture the aforementioned intuition, such as Lipschitz mapping-based, probability Lipschitzness, and the $(\epsilon-\delta)$ language-based formulations. 
In other words, for a mapping $h:\mathcal{X}\mapsto\mathcal{Y}$, we say it satisfies individual fairness if for all possible $(\xneg,\xpos), ({\xneg}',{\xpos}') \in\mathcal{X}$, it holds any one of the following: 
(1) it is $\lambda$-Lipschitz w.r.t. appropriate metrics on the domain $\mathcal{X}$ and the codomain $\mathcal{Y}$, that is,
\begin{equation}
\topequation%
    \mathbf{d}_{\mathcal{Y}}\big(
        h(\xneg,\xpos),h({\xneg}',{\xpos}')
    \big) \leqslant \lambda\cdot
    \mathbf{d}_{\mathcal{X}}\big(
        (\xneg,\xpos),({\xneg}',{\xpos}')
    \big) \,;
    \label{eq:indiv,1}
\end{equation}%
(2) it is probability Lipschitz w.r.t. appropriate metrics on the domain and the codomain, that is, 
$%
    \mathbb{P}_\mathcal{D}\left[
    \mathbf{d}_\mathcal{Y}\big(
        h(\xneg,\xpos),h({\xneg}',{\xpos}')
    \big) / \mathbf{d}_\mathcal{X}\big(
        (\xneg,\xpos),({\xneg}',{\xpos}')
    \big) \geqslant\epsilon
    \right] \leqslant\delta \,;
    \label{eq:indiv,2}
$ or (3) it holds
$%
    \mathbf{d}_\mathcal{X}\big(
        (\xneg,\xpos),({\xneg}',{\xpos}')
    \big) \leqslant\epsilon 
    \Rightarrow
    \mathbf{d}_\mathcal{Y}\big(
        h(\xneg,\xpos),h({\xneg}',{\xpos}')
    \big) \leqslant\delta \,,
    \label{eq:indiv,3}
$ %
where we consider $\epsilon\geqslant 0$ and $\delta\geqslant 0$. 
} the treatment/evaluation on one instance should not change solely due to minor changes or perturbation in its \glspl{sa}. This indicates the existence of underlying discriminative risks if one hypothesis/classifier makes different predictions for an instance based solely on changes in \glspl{sa}. 
Consequently, the fairness quality of one hypothesis $f(\cdot)$ can be evaluated by 
\begin{equation}
\topequation
    \justL(f,\bm{x}) =
    \mathbb{I}(
        f(\xneg,\xpos) \neq
        f(\xneg,\xqtb)
    ) \,,\label{eq:1}
\end{equation}%
similarly to the $0/1$ loss. 
Note that \eqref{eq:1} is evaluated on only one instance. 
To describe this characteristic of the hypothesis on multiple instances (\aka{} from a group level), 
then the empirical discriminative risk on $S$ and the true discriminative risk of the hypothesis are expressed as 
\begin{equation}
\topequation\textstyle
    \justLf(f,S) = 
    \frac{1}{n}\sum_{i=1}^n 
    \justL(f,\bm{x}_i)
    \,,\label{eq:1a}
\end{equation}%
and
\begin{equation}
\topequation
    \justLt(f)= 
    \mathbb{E}_{(\bm{x},y)\sim\mathcal{D}}[ \justL(f,\bm{x})]
    \,,\label{eq:1b}
\end{equation}%
respectively. 
Note that \(\justL(f,\bm{x})\) quantifies instance-level discriminative risk, following the intuition that ``there will exist something unfair if similar instances are not treated similarly''; then \(\justLf(f,S)\) and \(\justLt(f)\) introduce themselves as measuring and gathering $\justL(f,\bm{x})$ of all instances over a dataset or distribution, thereby capturing a group-level perspective. The latter two aggregate this risk at the dataset/distribution level across all sensitive groups,  yet without explicit subgroup partitioning, which is advantageous. 
Overall, \gls{dr} measures the discriminative risk of one hypothesis $f(\cdot)$ from both individual and group viewpoints.

Furthermore, we argue that \emph{the empirical \gls{dr} on $S$ is an unbiased estimation of the true \gls{dr}}. 
The reason is that for one random variable $\mathsf{X}$ representing instances, $\justL(f,\bm{x})$ in \eqref{eq:1} could be viewed as a new random variable obtained by using a few fixed operations on $\mathsf{X}$, recorded as $\mathsf{Y}$. 
Then for $n$ random variables (\ie{} $\mathsf{X}_1,\mathsf{X}_2,...,\mathsf{X}_n$ representing instances) that are i.i.d., 
by operating them in the same way, we can get random variables $\mathsf{Y}_1,\mathsf{Y}_2,...,\mathsf{Y}_n$ that are i.i.d. as well. 
Then we can rewrite $\justLf(f,S)$ as $\frac{1}{n} \sum_{i=1}^n \mathsf{Y}_i$ and $\justLt(f)$ as $\mathbb{E}_{\mathsf{Y}\sim \mathcal{D}'}[\mathsf{Y}]$, where $\mathcal{D}'$ denotes the space after operating $\mathsf{X}\sim\mathcal{D}$. 
Therefore, it could be easily seen that the former is an unbiased estimation of the latter. 

It is worth noting that a model's \gls{dr} equalling zero is a {\em necessary but not sufficient condition} for the statement that ``the model is purely fair''. That is to say, there must be something discriminative within the model if $\justLf(f,S)>0$, but it would still be too soon to claim the model's fairness when $\justLf(f,S)=0$ (in other words, discrimination may persist within the model, but we just have not detected it yet). 
Despite smaller values, \gls{dr} is more intuitive than the commonly used group fairness measures (such as \gls{dp}, \gls{eopp}, and \gls{pqp}), and can better capture the differential treatment that is induced by \glspl{sa} only than them (see \cref{expt:quality}), on the premise that when \glspl{sa} are available, while computing such group fairness measures also requires the \glspl{sa}' availability. 
Additionally, we do not fix any specific \glspl{sa} in \gls{dr} on purpose, so that it can be applied to any scenarios once participants designate the relevant protected attribute(s) according to their contents or based on needs, widening its applicability.

\subsubsection{The distinction compared with existing fairness measures}

Designing \gls{dr} is motivated by both individual- and group-fairness principles, aiming to quantify bias from both individual- and population-level perspectives. It can also be viewed as a model-free analogue of causal fairness notions (\eg{} \gls{cff}\footnote{
Given a predictive problem where $A$, $X$ and $Y$ denote the protected attributes, remaining attributes, and output of interest respectively, we assume that a causal model $(U,V,F)$ is given, where $V\equiv A\cup X$. 
The \emph{counterfactual fairness} (\gls{cff}) is a postulated criterion for predictors of $Y$, that is, 
a predictor $\hat{Y}$ is counterfactual fair if under any context $X=x$ and $A=a$, 
$
    \topequation\footnotesize
    \mathbb{P}(\hat{Y}_{A\gets a}(U)=y \mid X=x, A=a) =
    \mathbb{P}(\hat{Y}_{A\gets a'}(U)=y \mid X=x, A=a) 
    \,,\label{eq:causal,cf}
$ 
for any $y$ and for any value $a'$ attainable by $A$. 
Note that here $\gets$ is a symbol from causality \citep{pearl2000models,kusner2017counterfactual}: for two observable variables $A$ and $\hat{Y}$, a counterfactual statement ``the value of $\hat{Y}$ if $A$ had taken value $a$'' is denoted by $\hat{Y}_{A\gets a}(u)$ for $\hat{Y}$ given $U=u$ where the equations for $A$ are replaced with $A=a$. 
} \citep{kusner2017counterfactual} and proxy discrimination\footnote{
We generally consider causal graphs involving a protected attribute $A$, a set of proxy variables $P$, features $X$, a predictor $R$, and sometimes an observed outcome $Y$. A causal graph is a directed, acyclic graph whose nodes represent random variables. 
A variable $V$ in a causal graph exhibits \emph{potential proxy discrimination}, if there exists a direct path from $A$ to $V$ that is blocked by a proxy variable and $V$ itself is not a proxy; 
An intervention on $P$ is denoted by $\mathrm{do}(P=p)$, and 
a predictor $R$ exhibits no \emph{proxy discrimination} based on a proxy $P$ if for all $p,p'$,
$
    \topequation
    \footnotesize
    \mathbb{P}(R\mid \mathrm{do}(P=p)) =
    \mathbb{P}(R\mid \mathrm{do}(P=p')) 
    \,.\label{eq:causal,pd}
$
} \citep{kilbertus2017avoiding}), as it probes sensitivity of predictions to changes in protected attributes, yet without explicit causal graphs or structural equations. Nevertheless, \gls{dr} differs substantively from these measures in its formulation and assumptions, as detailed below.

\paragraph{Two distinctions from individual fairness}
The Lipschitz property of individual fairness \citep{dwork2012fairness,dwork2018fairness} enforces similar predictions for similar individuals under a chosen similarity metric (\ie{} $\mathbf{d}_\mathcal{X}$ and $\mathbf{d}_\mathcal{Y}$ in~\eqref{eq:indiv,1}), whose specification is often crucial and potentially misleading when misspecified. In contrast, \gls{dr} compares an instance to its slightly perturbed version obtained by modifying only sensitive attribute(s), thereby avoiding metric-based similarity control: the two instances are treated as identical in non-sensitive features, yielding $\mathbf{d}((\xneg,\xpos),(\xneg,\xqtb))=0$. Moreover, while \eqref{eq:indiv,1} compares pairs both drawn from the observed dataset $S$, \gls{dr} compares $(\xneg,\xpos)\in S$ with a perturbed counterpart $(\xneg,\xqtb)\notin S$ in~\eqref{eq:1}. 
In short, both \gls{dr} and the Lipschitz property follow the same principle of individual fairness; however, they use different forms of expression.

\paragraph{Two distinctions from group fairness}
Group fairness typically assesses disparities between subgroups defined by \glspl{sa}, \eg{} by computing criterion differences across groups as in~\eqref{eq:group,fair}. The proposed \eqref{eq:1a} and \eqref{eq:1b} instead evaluate \gls{dr} on the dataset as a whole without explicit subgroup partitioning: whenever a classifier changes its prediction for an instance under \gls{sa} perturbations, discriminative risk exists regardless of the instance's group membership. In addition, unlike many standard group-fairness metrics that primarily target a single binary \gls{sa}, \gls{dr} naturally extends to multi-value settings and multiple \glspl{sa}, broadening its applicability.

\paragraph{Three distinctions from causal fairness}
\gls{dr} is related in spirit to causal notions such as \gls{cff} \citep{kusner2017counterfactual} and proxy discrimination \citep{kilbertus2017avoiding}, but differs in several respects. 
First, \gls{dr} is a scalar quantity computed without an explicit causal graph/model $(U,V,F)$, reducing modelling burden and misspecification risks. 
Second, \gls{cff} may entail changes in non-sensitive features via proxies, whereas \gls{dr} evaluates sensitivity by holding $\xneg$ fixed and perturbing only \glspl{sa}. 
Third, counterfactual criteria qualify fairness over all counterfactual values, whereas \gls{dr} captures risk under stochastic perturbations implemented by sampling $\xqtb$ with probability $p\in[0,1]$. 

\paragraph{Similarities with existing measures}
\gls{dr} follows the individual-fairness intuition by comparing $(\xneg,\xpos)$ to $(\xneg,\xqtb)$ that differ only in \glspl{sa}, treating them as sufficiently similar. 
Aggregating $\justL(f,\bm{x})$ over a dataset/distribution yields a demographic/statistical view consistent with group fairness. 
If desired, one can also compute a subgroup-difference analogue as in \eqref{eq:group,fair}, like 
\begin{equation}
\topequation
    \justLt^\prime(f)=| 
    \mathbb{E}_{(\bm{x},y)\sim\mathcal{D} |\bm{a}=1}[ \justL(f,\bm{x}) ]-
    \mathbb{E}_{(\bm{x},y)\sim\mathcal{D} |\bm{a}=0}[ \justL(f,\bm{x}) ]
    | \,,\label{eq:dr,prime}
\end{equation}
though \emph{DR} does not require subgroup partitioning.

\subsection{Oracle bounds of discriminative risk for weighted voting}
\label{subsec:bound}
\label{method:1,2}

In this subsection, we analyse the role that ensemble combination plays in fairness improvement. 
Here we consider \emph{weighted voting} as the ensemble combination, where the prediction by an ensemble of $m$ individual classifiers, parameterised by a weight vector $\rho= [w_1,...,w_m]^\mathsf{T} \in[0,1]^m$, 
is given by $\mvrho(\bm{x})= \argmax_{y\in\mathcal{Y}} \sum_{j=1}^m w_j\cdot$\,\\$\mathbb{I}( f_j(\bm{x})=y )$, such that $\sum_{j=1}^m w_j =1$, and $w_j$ is the weight of individual classifier $f_j(\cdot)$.  
It could be viewed as $w_j=\sfrac{1}{m}$ for all $j\in\{1,2,...,m\}$ in plurality voting and majority voting, meaning that all individual classifiers will be taken into account equally. Note that ties will be resolved arbitrarily, and both parametric and non-parametric models can serve as individual classifiers.

For brevity, we record $\sum_{j=1}^m w_j\mathbb{I}(f_j(\insx)=y)$ as $q_y(\insx)$, and $y^* \defineq \mvrho(\insx)$. Then we define the \emph{top-1 voting margin} as 
\begin{equation}
\topequation
\textstyle
\gamma_\rho(\insx) \defineq 
q_{y^*}(\insx) - \max_{y\neq y^*} q_y(\insx)
\in [0,1] \,.
\end{equation}
If the ensemble exhibits discrimination risk, it means that the leading gap in the prediction class has been reversed, in other words, at least half of the votes for the prediction class will be altered. That is to say, the total weight of the inconsistency between the original prediction and the perturbed prediction has to hold
\begin{equation}
\topequation 
\incons(\insx) \defineq 
\mathbb{E}_\rho[\justL(f,\insx)] =
\mathbb{E}_\rho[ \mathbb{I}\big(
    f(\xneg,\xpos) \neq f(\xneg,\xqtb)
\big) ] \geqslant 
\tfrac{1}{2} \gamma_\rho(\insx) 
\,.
\end{equation}
Therefore, we have the discriminative risk of one ensemble
\begin{equation}
\topequation
\justL(\mvrho,\insx) \leqslant 
\mathbb{I}\big( 
\incons(\insx) \geqslant 
\tfrac{1}{2} \gamma_\rho(\insx) \big) 
\,,\label{eq:marg,bnd}
\end{equation}
and can discuss some bounds concerning fairness for the weighted vote, inspired by the work of Masegosa~\etal{} \citep{masegosa2020second}. 

Following the notations described in \cref{method:4}, the observation in \eqref{eq:marg,bnd} leads to our proposed first- and second-order oracle bounds for the fairness quality of weighted vote. 
Note that there are no more assumptions except the aforementioned notations in the following theorems. 
For brevity, $\mathbb{E}_{(\bm{x},y)\sim\mathcal{D}}[\cdot]$ and $\mathbb{E}_{f\sim\rho}[\cdot]$ may be respectively abbreviated as $\mathbb{E}_{\mathcal{D}}[\cdot]$ and $\mathbb{E}_{\rho}[\cdot]$ when the context is unambiguous. 
Also, note that a \emph{prominent difference} between our work and the work of Masegosa~\etal{} \cite{masegosa2020second} is that they investigate the expected risk of accuracy, with no relationship with margins.

\subsubsection{Oracle bounds regarding fairness for weighted voting}
\label{method:1}
\begin{theorem}[First-order oracle bound]
\label{thm:1}
\begin{equation}
\topequation
    \justLt(\mvrho) \leqslant
    2\mathbb{E}_\mathcal{D} \left[
        \tfrac{ \incons(\insx) }{ \gamma_\rho(\insx) }
    \right]
    \,.\label{eq:2}
\end{equation}
\end{theorem}
\begin{proof}
We have
$ \justLt(\mvrho) 
    = \mathbb{E}_\mathcal{D}[ \justL(\mvrho,\bm{x}) ]
    \leqslant \mathbb{P}_\mathcal{D}\big( 
        \incons(\insx) \geqslant 
        \tfrac{1}{2} \gamma_\rho(\insx) 
    \big)
    \,.\nonumber
$ %
By applying Markov's inequality to random variable $Z= \tfrac{ \incons(\insx) }{ \gamma_\rho(\insx) }$, we get
$ \justLt(\mvrho) \leqslant 
    \mathbb{P}_\mathcal{D} \left(
        \tfrac{ \incons(\insx) }{ \gamma_\rho(\insx) }
        \geqslant 0.5
    \right) 
    \leqslant
    2\mathbb{E}_\mathcal{D} \left[
        \tfrac{ \incons(\insx) }{ \gamma_\rho(\insx) }
    \right] 
    \,.\nonumber%
$ %
\end{proof}

\begin{theorem}[Second-order oracle bound]
\label{thm:3}
In multi-class classification
\begin{equation}
\topequation
    \justLt(\mvrho) \leqslant
    4\mathbb{E}_\mathcal{D}\left[
        \tfrac{ \incons(\insx)^2 }{ \gamma_\rho(\insx)^2 }
    \right]
    \,.\label{eq:5}
\end{equation}%
\end{theorem}
\begin{proof}
By applying second-order Markov's inequality to $Z= \tfrac{ \incons(\insx) }{ \gamma_\rho(\insx) }$, we obtain
$ \justLt(\mvrho) \leqslant 
    \mathbb{P}_\mathcal{D}\left(
        \tfrac{ \incons(\insx) }{ \gamma_\rho(\insx) }
        \geqslant 0.5
    \right) 
    \leqslant 
    \mathbb{P}_\mathcal{D}\Big( \Big( 
        \tfrac{ \incons(\insx) }{ \gamma_\rho(\insx) }
    \Big)^2 \geqslant 0.25 \Big)
    \leqslant
    4\mathbb{E}_\mathcal{D}\left[
        \tfrac{ \incons(\insx)^2 }{ \gamma_\rho(\insx)^2 }
    \right] 
    \,.\nonumber
$ %
\end{proof}

Furthermore, we can also obtain an alternative second-order bound based on Chebyshev-Cantelli inequality, presented in Theorem~\ref{thm:4}. 

\begin{theorem}[C-tandem oracle bound]
\label{thm:4}
If $\mathbb{E}_\mathcal{D}\left[ \incons(\insx)/\gamma_\rho(\insx) \right] <\frac{1}{2}$\,, then
\begin{equation}
\topequation
    \justLt(\mvrho) \leqslant
    \tfrac{%
        \mathbb{E}_\mathcal{D}[ 
        \incons(\insx)^2 / \gamma_\rho(\insx)^2 ]
        -\mathbb{E}_\mathcal{D}[ 
        \incons(\insx) / \gamma_\rho(\insx) ]^2
    }{%
        \mathbb{E}_\mathcal{D}[ 
        \incons(\insx)^2 / \gamma_\rho(\insx)^2 ]
        -\mathbb{E}_\mathcal{D}[ 
        \incons(\insx) / \gamma_\rho(\insx) ]
        +\frac{1}{4}
    }\,.\label{eq:6}
\end{equation}%
\end{theorem}%
\begin{proof}[Proof of \cref{thm:4}]
Let the Chebyshev-Cantelli inequality be applied to $\sfrac{ \incons(\insx) }{ \gamma_\rho(\insx) }$ and we can get
\begin{subequations}
\topequation
\begin{align}
    \justLt(\mvrho) 
    &\leqslant
    \mathbb{P}_\mathcal{D}\big(
        \incons(\insx) 
        \geqslant \tfrac{1}{2} \gamma_\rho(\insx)
    \big) \nonumber\\
    &= \mathbb{P}_\mathcal{D}\left(
        \tfrac{ \incons(\insx) }{ \gamma_\rho(\insx) }
        -\mathbb{E}_\mathcal{D}\left[
            \tfrac{ \incons(\insx) }{ \gamma_\rho(\insx) }
        \right] \geqslant \tfrac{1}{2}
        -\mathbb{E}_\mathcal{D}\left[
            \tfrac{ \incons(\insx) }{ \gamma_\rho(\insx) }
        \right] 
    \right) \nonumber\\
    &= \tfrac{ 
        \mathbb{E}_\mathcal{D}\left[ 
            \frac{ \incons(\insx)^2 }{ \gamma_\rho(\insx)^2 }
        \right]
        -\mathbb{E}_\mathcal{D}\left[
            \frac{ \incons(\insx) }{ \gamma_\rho(\insx) }
        \right] ^2
    }{
        \left( \tfrac{1}{2}-\mathbb{E}_\mathcal{D}\left[
            \frac{ \incons(\insx) }{ \gamma_\rho(\insx) }
        \right] \right)^2 +
        \mathbb{E}_\mathcal{D}\Big[
            \frac{ \incons(\insx)^2 }{ \gamma_\rho(\insx)^2 }
        \Big]
        -\mathbb{E}_\mathcal{D}\left[
            \frac{ \incons(\insx) }{ \gamma_\rho(\insx) }
        \right] ^2
    } \nonumber\,. 
\end{align}%
\end{subequations}%
Then \eqref{eq:6} follows from here.
\end{proof}

Up to now, we have gotten the first- and second-order oracle bounds of the fairness quality for the weighted vote, which will help us further investigate fairness. 
Through these theorems, we can see that the \gls{dr} of an ensemble can be bounded by a constant times the quotient of the \gls{dr} of the individual classifiers and the top-1 voting margin, which suggests that a cancellation-of-biases effect---depending on voting margins as well---is not as obvious as the cancellation-of-errors effect in ensemble combination. 

Furthermore, it is worth noting that, despite the similar names of ``first- and second-order oracle bounds'' from our inspiration \citep{masegosa2020second}, the essences of our bounds are distinct from theirs. 
Specifically, their work investigates the bounds for generalisation error and is relevant to neither fairness issues nor voting margins. 
In other words, their bounds are based on the 0/1 loss $\justAcc(f,\bm{x})= \mathbb{I}(f(\bm{x})\neq y)$, while ours are built upon $\justL(f,\bm{x})$ in \eqref{eq:1} and $\gamma_\rho(\insx)$ in \eqref{eq:marg,bnd}.

\subsubsection{Relaxation of these oracle bounds}
\label{method:5}

To increase readability, we add an additional verifiable low-margin quality control. Assume that there exist $\gamma_0\in(0,1]$ and $\eta\in[0,1]$, such that $\mathbb{P}_\mathcal{D}( \gamma_\rho(\insx) <\gamma_0 )\leqslant \eta$, where $\gamma_0$ be estimated on the validation set after training and $\eta$ is the proportion of low-marginal-value samples. Then we have two corollaries.

\begin{corollary}[Relaxation of the first-order oracle bound]
\label{col:5}
\begin{equation}
\topequation
\justLt(\mvrho) \leqslant
\tfrac{2}{\gamma_0} \mathbb{E}_\rho[ \justLt(f) ]
+\eta \,.
\end{equation}
\end{corollary}
\begin{proof}
The dataset $S$ can be divided into $S_G=\{ (\insx,y)\in S\mid \gamma_\rho(\insx) \geqslant\gamma_0 \}$ and $S_B=\{ (\insx,y)\in S\mid \gamma_\rho(\insx) <\gamma_0 \}$, whereas $\mathbb{P}(\insx\in S_B)\leqslant \eta$. 
For any $\insx\in S_G$, we have $\mathbb{P}_\mathcal{D}\left( \sfrac{\incons(\insx)}{\gamma_\rho(\insx)} \geqslant 0.5 \right) \leqslant \mathbb{P}_\mathcal{D}\left( \sfrac{\incons(\insx)}{\gamma_0} \geqslant 0.5 \right) \leqslant \tfrac{2}{\gamma_0} \mathbb{E}_\rho[ \mathbb{P}_\mathcal{D}(\justL(f,\insx)) ] =\tfrac{2}{\gamma_0}\mathbb{E}_\rho[ \justLt(f) ]$. 
For any $\insx\in S_B$, we have $\mathbb{P}_\mathcal{D}\left( \sfrac{\incons(\insx)}{\gamma_\rho(\insx)} \geqslant 0.5 \right) \leqslant 1$. 
Together, we obtain that $\justLt(\mvrho) \leqslant \mathbb{E}_\mathcal{D}[ \tfrac{2}{\gamma_0}$\, $\mathbb{E}_\rho[\justLt(f)] \cdot \mathbb{P}(\insx\in S_G) ]+\mathbb{E}_\mathcal{D}[ 1\cdot \mathbb{P}(\insx\in S_B) ] \leqslant \tfrac{2}{\gamma_0}\mathbb{E}_\rho[\justLt(f)]+\eta$. 
\end{proof}

To investigate the bound deeper, we introduce here the tandem fairness quality of two hypotheses $f(\cdot)$ and $f'(\cdot)$ on one instance $(\bm{x},y)$, adopting the idea of the tandem loss \citep{masegosa2020second}, by 
\begin{equation}
    \topequation
    \justL(f\!,f'\!,\bm{x}) \!=
    \mathbb{I}\big(\big(
        f(\xneg,\xpos) \!\neq\!
        f(\xneg,\xqtb)
    \big) \land \big(
        f'\!(\xneg,\xpos) \!\neq\!
        f'\!(\xneg,\xqtb)
    \big)\big) \,.\label{eq:3}
\end{equation}%
The tandem fairness quality counts a discriminative decision on the instance $(\bm{x},y)$ if and only if both $f(\cdot)$ and $f'\!(\cdot)$ give a discriminative prediction on it. 
Note that in the degeneration case $\justL(f,f,\bm{x}) = \justL(f,\bm{x})$.
Then the expected tandem fairness quality is defined by $\justLt(f,f')= \mathbb{E}_\mathcal{D}[ \justL(f,f',\bm{x}) ]$. 
Lemma~\ref{lem:2} relates the expectation of the second moment of the standard fairness quality to the expected tandem fairness quality. 
Note that $\mathbb{E}_{f\sim\rho,f'\sim\rho}[\justLt(f,f')]$ for the product distribution $\rho\times\rho$ over $\mathcal{F}\times\mathcal{F}$ can be abbreviated as $\mathbb{E}_{\rho^2}[\justLt(f,f')]$ for brevity.

\begin{lemma}
\label{lem:2}
In multi-class classification,
\begin{equation}
\topequation
    \mathbb{E}_\mathcal{D}[
    \mathbb{E}_\rho[
        \justL(f,\bm{x})
    ]^2]=
    \mathbb{E}_{\rho^2}[
        \justLt(f,f')]
    \,.\label{eq:4}
\end{equation}
\end{lemma}
\begin{proof}
We have
\begin{subequations}
\topequation
\begin{align}
    & \mathbb{E}_\mathcal{D}[
    \mathbb{E}_\rho[
        \justL(f,\bm{x})
    ]^2] \nonumber\\
    =& \mathbb{E}_\mathcal{D}[
    \mathbb{E}_\rho[
    \mathbb{I}(
        f(\xneg,\xpos) \neq
        f(\xneg,\xqtb)
    )] \mathbb{E}_\rho[
    \mathbb{I}(
        f'(\xneg,\xpos) \neq
        f'(\xneg,\xqtb)
    )]] \nonumber\\
    =& \mathbb{E}_\mathcal{D}[
    \mathbb{E}_{\rho^2}[
    \mathbb{I}(
        f(\xneg,\xpos) \neq
        f(\xneg,\xqtb)
    ) \mathbb{I}(
        f'(\xneg,\xpos) \neq
        f'(\xneg,\xqtb)
    )]] \nonumber\\
    =& \mathbb{E}_\mathcal{D}[
    \mathbb{E}_{\rho^2}[
    \mathbb{I}((
        f(\xneg,\xpos) \ne
        f(\xneg,\xqtb)
    )\!\land\!(
        f'(\xneg,\xpos) \ne
        f'(\xneg,\xqtb)
    ))]] \nonumber\\
    =& \mathbb{E}_{\rho^2}[
    \mathbb{E}_\mathcal{D}[
    \mathbb{I}((
        f(\xneg,\xpos) \ne
        f(\xneg,\xqtb)
    )\!\land\!(
        f'(\xneg,\xpos) \ne
        f'(\xneg,\xqtb)
    ))]] \nonumber \,.
\end{align}%
\end{subequations}%
Then we can get the left hand side of the above equation equals $\probE_{\rho^2}[ \probE_{\mathcal{D}}[ \justL(f,f',\bm{x}) ]]$, that is, $\probE_{\rho^2}[ \justLt(f,f') ]$. 
\end{proof}

\begin{corollary}[Relaxation of the second-order oracle bound]
\label{col:6}
\begin{equation}
\topequation
\justLt(\mvrho) \leqslant
\tfrac{4}{\gamma_0^2} \mathbb{E}_{\rho^2}[\justLt(f,f')] 
+\eta \,.
\end{equation}   
\end{corollary}
\begin{proof}
The dataset $S$ is still divided into $S_G$ and $S_B$, whereas $\mathbb{P}(S_B)\leqslant\eta$. 
For any $\insx\in S_G$, we have $\mathbb{P}_\mathcal{D}( \sfrac{\incons(\insx)^2}{\gamma_\rho(\insx)^2} \geqslant $\, $0.25 ) \leqslant \mathbb{P}_\mathcal{D}\left( \sfrac{\incons(\insx)^2}{\gamma_0^2} \geqslant 0.25 \right) \leqslant \tfrac{4}{\gamma_0^2} \mathbb{P}_\mathcal{D}(\incons(\insx)^2) =\tfrac{4}{\gamma_0^2} \mathbb{E}_{\rho^2}[\justLt(f,f')]$. 
For any $\insx\in S_B$, we have $\mathbb{P}_\mathcal{D}\left( \sfrac{\incons(\insx)^2}{\gamma_\rho(\insx)^2} \geqslant 0.25 \right) \leqslant 1$. Together, we obtain that $\justLt(\mvrho)\leqslant \mathbb{E}_\mathcal{D}[ \tfrac{4}{\gamma_0^2} \mathbb{E}_{\rho^2}[\justLt(f,f')] \cdot\mathbb{P}(S_G) ]+\mathbb{E}_\mathcal{D}[ 1\cdot\mathbb{P}(S_B) ]\leqslant \tfrac{4}{\gamma_0^2} \mathbb{E}_{\rho^2}[\justLt(f,f')]+\eta$. 
\end{proof}

\subsubsection{PAC bounds for the weighted voting}%
\label{method:2}
All oracle bounds described above 
are expectations that can only be estimated on finite samples instead of being calculated precisely. 
They could be transformed into empirical bounds via PAC-bound analysis as well to ease the difficulty of giving a theoretical guarantee of the performance on any unseen data, which we discuss in this subsection. 
Based on Hoeffding's inequality, we can deduct generalisation bounds presented below.

\begin{theorem}
\label{thm:n5}
For any $\delta\in(0,1)$, with probability at least $(1\!-\!\delta)$ over a random draw of $S$ with a size of $n$, for a single hypothesis $f(\cdot)$, 
\begin{equation}
\topequation
    \justLt(f) \leqslant \justLf(f,S) 
    +\sqrt{\tfrac{1}{2n}\ln\tfrac{1}{\delta}}
    \,.\label{eq:n8}
\end{equation}
\end{theorem}
\begin{proof}
According to Hoeffding's inequality, for any $\varepsilon >0$, we have
\begin{equation}
    \topequation
    \mathbb{P}(
        \justLt(f)- \justLf(f,S)
    \geqslant \varepsilon
    )\leqslant e^{-2n\varepsilon^2}
    \,.\label{eq:n10}%
\end{equation}%
Let $\delta\defineq e^{-2n\varepsilon^2} \!\in(0,1)$, 
we can obtain $\varepsilon\!= \sqrt{\sfrac{1}{(2n)}\!\ln(\sfrac{1}{\delta})}$. 
Then with probability at least $(1\!-\!\delta)$ we have \eqref{eq:n8}. 
\end{proof}

\begin{theorem}
\label{thm:5}
For any $\delta\in(0,1)$, with probability at least $(1-\delta)$ over a random draw of $S$ with a size of $n$, for all distributions $\rho$ on $\mathcal{F}$,
\begin{equation}
    \topequation
    \justLt(\mvrho) \leqslant \justLf(\mvrho{})
    +\sqrt{ \tfrac{1}{2n}
    \ln\tfrac{
        |\mathcal{F}|
    }{\delta} }
    \,.\label{eq:8}
\end{equation}%
\end{theorem}
\begin{proof}
We have known that for one single hypothesis $f(\cdot)$, and for any $\varepsilon\!>\!0$, it holds \eqref{eq:n10}. 
Then for a hypotheses set $\mathcal{F}$, we will have
$
   \mathbb{P}\big(
      \exists\, \rho\in[0,1]^{|\mathcal{F}|} \!:
      \justLt(\mvrho)- \justLf(\mvrho,S)
      \geqslant\varepsilon
   \big) 
   \leqslant \textstyle
   \sum_{f\in\mathcal{F}} \mathbb{P}\big(
      \justLt(f)- \justLf(f,S)
      \geqslant\varepsilon
   \big) 
   \leqslant \textstyle
   \sum_{j\in\mathcal{F}} e^{-2n\varepsilon^2}
   = |\mathcal{F}|e^{-2n\varepsilon^2} 
   \,.
$ 
Let $\delta\defineq |\mathcal{F}|e^{-2n\varepsilon^2}$, then we get $\varepsilon\!= \sqrt{\smash[b]{ \sfrac{1}{(2n)}\ln(\sfrac{|\mathcal{F}|}{\delta}) }}$. 
Thus with probability at least $(1\!-\!\delta)$ we have \eqref{eq:8}. 
\end{proof}


\subsection{Application in constructing fairer ensembles with little accuracy degradation}
\label{method:3}

We leverage the above bounds to construct fairer ensembles while controlling accuracy loss. 
We evaluate the accuracy quality of a hypothesis $f(\cdot)$ by the 0/1 loss $\justAcc(f,\bm{x})$, the empirical loss by $\justAccF(f,S)= \tfrac{1}{n}\sum_{i=1}^n \justAcc(f,\bm{x})$, and the expected loss by $\justAccT(f)= \mathbb{E}_\mathcal{D}[ \justAcc(f,\bm{x})]$, respectively. 
To jointly account for fairness and accuracy, we adopt a bi-objective view and seek Pareto-efficient sub-ensembles \citep{qian2015pareto}.

\paragraph{Pareto dominance.} 
For two sub-objectives to be minimised, $\mathcal{G}=(\justAccT,\justLt)$ as the objective, a probability distribution $\rho$ weakly dominates $\pi$ if 
$\justAccT(\mvrho)\le \justAccT(\mvpi)$ and $\justLt(\mvrho)\le \justLt(\mvpi)$, denoted as $\succeq_\mathcal{G}$, and dominates if at least one inequality is strict, denoted as $\succ_\mathcal{G}$. Note that both $\rho$ and $\pi$ are on $\mathcal{F}$ and dependent of $S$.\looseness=-1

\emph{An alternative scalar objective.}\hspace{1em}
We also consider a symmetric weighted-sum objective for comparing hypotheses,
\begin{equation}
\topequation 
\justObjT(f,f')=
\lambda\tfrac{\justAccT(f)+\justAccT(f')}{2}+(1-\lambda)\justLt(f,f') \,,
\label{eq:10}
\end{equation}
where $\lambda\in(0,1)$ as a regularisation factor trades off accuracy and fairness. 
In the degeneration case $\justObjT(f,f)= \lambda\justAccT(f)+ (1-\lambda)\justLt(f)$ when two hypotheses $f(\cdot)$ and $f'(\cdot)$ are identical. 
For a weighted vote $\mvrho$, this yields
\begin{equation}
\topequation
\justObjT(\mvrho)=
\lambda\mathbb{E}_\rho[\justAccT(f)]+(1-\lambda)\mathbb{E}_{\rho^2}[\justLt(f,f')] \,,
\label{eq:11}
\end{equation}
which we aim to minimise.

\paragraph{\poepfull{} (\poepabbr{})}
Based on Pareto dominance and the objective above, we propose an ensemble pruning procedure to select sub-ensembles that improve fairness with limited accuracy degradation. Implementation details are provided in Algorithm~\ref{alg:pareto} of Appendix~\ref{appx:prop}; we also report two simpler and easily-implemented pruning baselines there in Appendix~\ref{appx:alter} for faster execution.

\begin{figure}[b]
\begin{minipage}{.475\textwidth}
\centering 
\subfloat[]{\label{fig:expt4a}
\includegraphics[height=29mm]{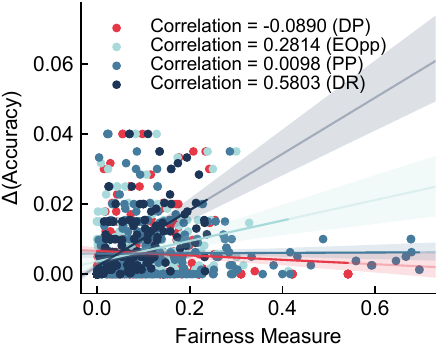}}
\subfloat[]{\label{fig:expt4b}
\includegraphics[height=28mm]{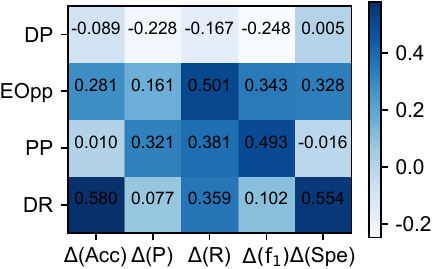}}
\vspace{-4mm}\caption{%
Comparison of the proposed \gls{dr} with three group fairness measures, based on the results from all datasets. 
\protect\subref{fig:expt4a} Scatter diagrams with the degree of correlation, where the $x$- and $y$-axes are different fairness measures and the variation of accuracy between the raw and perturbed data. 
\protect\subref{fig:expt4b} Correlation among multiple criteria. 
}\label{fig:expt4}
\end{minipage}
\begin{minipage}{.475\textwidth}
\vspace{-2mm}
\centering
\subfloat[]{\label{subfig:cf,a}
\includegraphics[height=29mm]{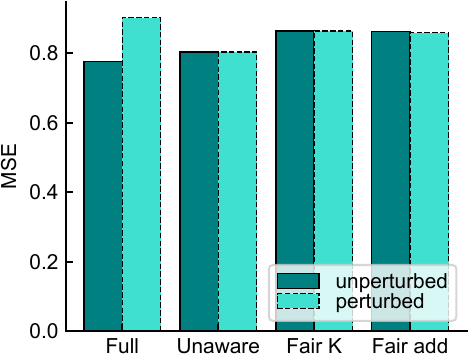}}
\hspace{4mm}
\subfloat[]{\label{subfig:cf,b}
\includegraphics[height=29mm]{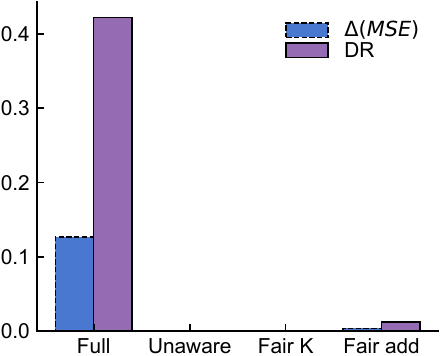}}
\vspace{-4mm}\caption{Example: law school success. 
\protect\subref{subfig:cf,a} Test MSE of different models, where `unperturbed' and `perturbed' denote the results obtained from the original and perturbed data respectively. 
\protect\subref{subfig:cf,b} The comparison between the change in MSE and \emph{\gls{dr}}, which suggests that $\mathrm{DR}\approx 0$ when the corresponding model satisfies or nearly satisfies counterfactual fairness. 
}\label{fig:cf}
\end{minipage}
\end{figure}

\section{Empirical Results}

In this section, we elaborate on experiments\footnote{Our code is available on \url{https://github.com/eustomaqua/FairML}.} to evaluate the effectiveness of the proposed \emph{\gls{dr}} and \poepabbr{} in Algorithm~\ref{alg:pareto}. 
These experiments are conducted to explore the following research questions: 
\textbf{RQ1}. Compared with the baseline fairness measures (that is, three commonly used group fairness measures: \gls{dp}, \gls{eopp}, and \gls{pqp}), does \gls{dr} capture the bias level of classifiers effectively, and can it capture discrimination from both individual and group fairness aspects? 
\textbf{RQ2}. Are the oracle bounds in sections~\ref{method:1}--\ref{method:5} and generalisation bounds in \cref{method:2} valid, and what is the key difference between our bounds in this work and those in previous work \citep{masegosa2020second}? 
\textbf{RQ3}. Compared with the \gls{sota} fairness-aware ensemble-based methods, does \poepabbr{} work well as a pruned sub-ensemble classifier? 
\textbf{RQ4}. As an ensemble pruning method, can \poepabbr{} achieve better results compared with the \gls{sota} ensemble pruning methods? 
Experimental setups and more results are reported in Appendices~\ref{subsec:setup} and \ref{appx:more}, respectively.

\begin{figure*}[tb]%
\begin{minipage}{\textwidth}%
\centering
\subfloat[]{\label{subfig:bnd,a}
\includegraphics[height=31mm]{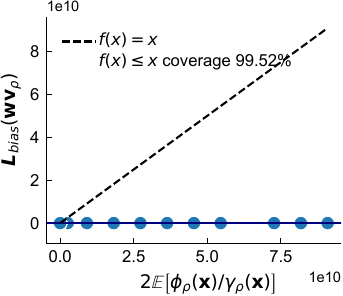}}\hspace{4mm}
\subfloat[]{\label{subfig:bnd,b}
\includegraphics[height=31mm]{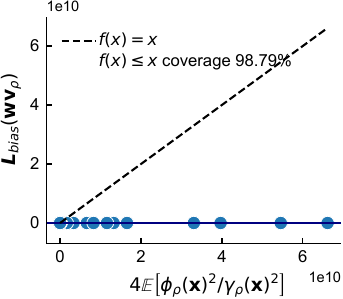}}
\subfloat[]{\label{subfig:bnd,c}
\includegraphics[height=31mm]{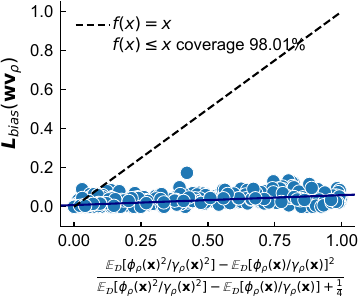}}\hspace{4mm}
\subfloat[]{\label{subfig:bnd,d}
\includegraphics[height=31mm]{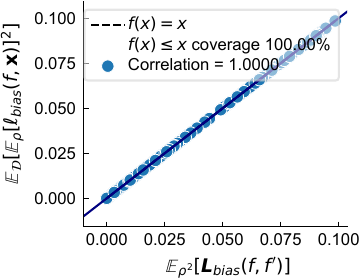}}
\\ \vspace{-4mm}
\subfloat[]{\label{subfig:bnd,e}
\includegraphics[height=31mm]{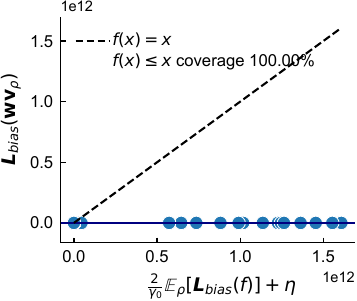}}\hspace{4mm}
\subfloat[]{\label{subfig:bnd,f}
\includegraphics[height=31mm]{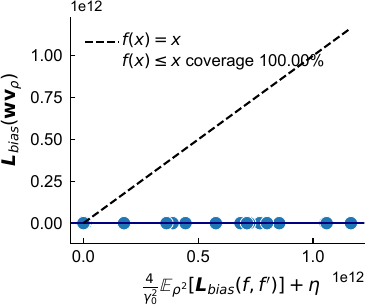}}
\subfloat[]{\label{subfig:bnd,g}
\includegraphics[height=31mm]{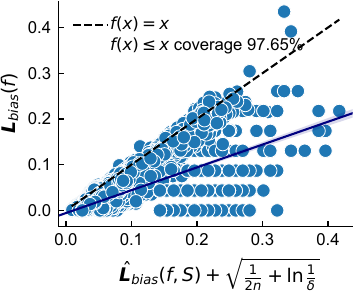}}
\hspace{4mm}
\subfloat[]{\label{subfig:bnd,h}
\includegraphics[height=31mm]{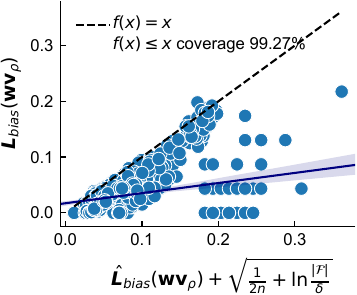}}
\vspace{-4mm}\caption{%
Verification of the proposed bounds in \cref{subsec:bound}, where (a--c) indicate theorems~\ref{thm:1} to \ref{thm:4}, (d) indicates lemma~\ref{lem:2}, (e--f) indicate corollaries~\ref{col:5} and \ref{col:6}, and (g--i) indicate theorems~\ref{thm:n5} to \ref{thm:5}, respectively. Note that the $x$-axes and $y$-axes are the corresponding right-hand sides and left-hand sides of these theorems, respectively; these scatters are supposed to be below the identity line $f(x)=x$, which is consistent with most cases in these empirical results. 
}\label{fig:bounds}
\end{minipage}
\begin{minipage}{\textwidth}%
\vspace{-1mm}%
\centering
\subfloat[]{\includegraphics[height=31mm]{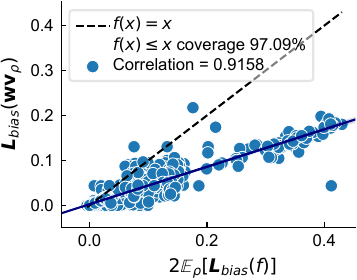}}\hspace{3mm}
\subfloat[]{\includegraphics[height=31mm]{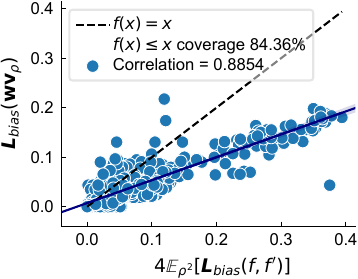}}\hspace{3mm}
\subfloat[]{\includegraphics[height=32mm]{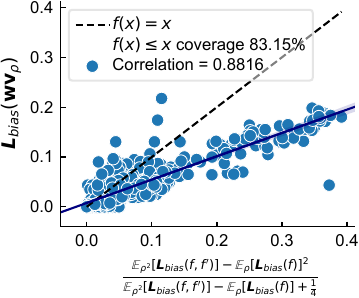}}\hspace{3mm}
\subfloat[]{
\includegraphics[height=32mm]{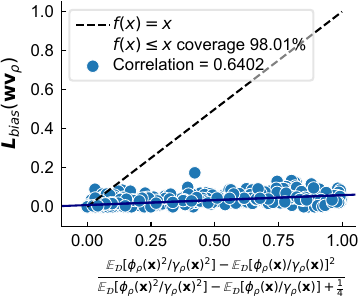}}
\vspace{-4mm}\caption{%
Comparison with the analogous forms to the work of \citet{masegosa2020second}, which are non- margin-dependent bounds. 
(a--c) In comparison with Fig.~\ref{fig:bounds}(a--c), we observe that the margin-dependent bounds of ours have a higher coverage than these analogues. 
(d) Another expression of Fig.~\protect\myref{fig:bounds}{subfig:bnd,c}. 
}\label{fig:prev}
\end{minipage}
\end{figure*}

\subsection{{\bf RQ1}: Validating the proposed fairness quality measure}
\label{expt:quality}

In this subsection, we evaluate the validity of the proposed fairness quality measure (namely \gls{dr}) in \cref{method:4}, compared with three commonly used group fairness measures. 
The empirical results are reported in Figures~\ref{fig:expt4} and \ref{fig:cf}.

As we can see from Fig.~\ref{fig:expt4}\subref{fig:expt4a}, compared with three group fairness measures (\ie{} \gls{dp}, \gls{eopp}, and \gls{pqp}), \gls{dr} has the highest value of correlation (namely the Pearson correlation coefficient) between itself and the variation of accuracy. 
It means that \gls{dr} captures better the characteristic of changed treatment than the three other group fairness measures when \glspl{sa} are perturbed, as the drop in accuracy indicates the existence of underlying discrimination hidden in models. 
Fig.~\ref{fig:expt4}\subref{fig:expt4b} reports the correlation between the variation of other criteria (such as precision, recall/sensitivity, $\text{f}_1$ score, and specificity), including accuracy and multiple fairness measures. 
It shows that \gls{dr} also has a higher correlation with the variation of specificity than the baseline fairness measures. 
Thus, we believe that \gls{dr} captures more changed treatment regarding the similar individuals than other group fairness measures, because \gls{dr} considers both group- and individual-fairness aspects.

\paragraph{Comparison with \gls{cff}: a case study} 
We also compare the proposed \emph{DR} with \gls{cff} using the same example (Law school success) given in the work of Kusner \etal{} \citep{kusner2017counterfactual}. 
The empirical results are reported in Fig.~\ref{fig:cf}, and more details are elaborated on in Appendix~\ref{appx:cf}. 
Kusner \etal{} \citep{kusner2017counterfactual} tested three models to meet \gls{cff}, that is, unaware, fair K, and fair add models. 
As shown in Fig.~\ref{fig:cf}\subref{subfig:cf,b}, \gls{dr} of the aforementioned models equals or is close to zero, which suggests \gls{dr} can somehow reflect \gls{cff} in a quantifiable way.\looseness=-1

\begin{figure*}[t]
\begin{minipage}{\textwidth}
\vspace{-3mm}\centering 
\subfloat[]{\label{subfig:aware,e}
\includegraphics[height=31mm]{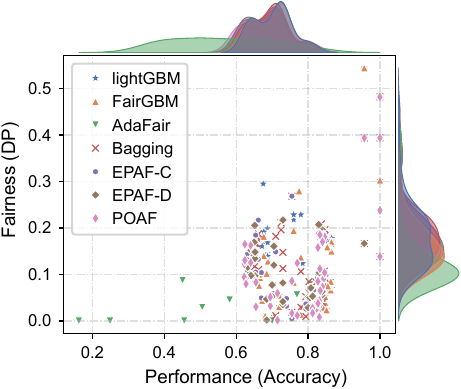}}
\hspace{4mm}
\subfloat[]{\includegraphics[height=31mm]{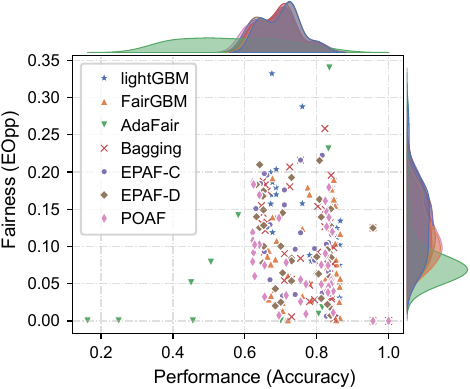}}
\hspace{4mm}
\subfloat[]{\includegraphics[height=31mm]{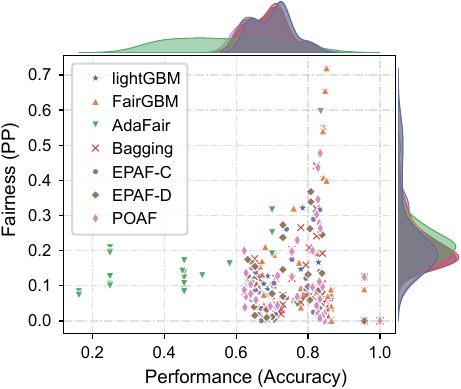}}
\hspace{4mm}
\subfloat[]{\label{subfig:aware,h}
\includegraphics[height=31mm]{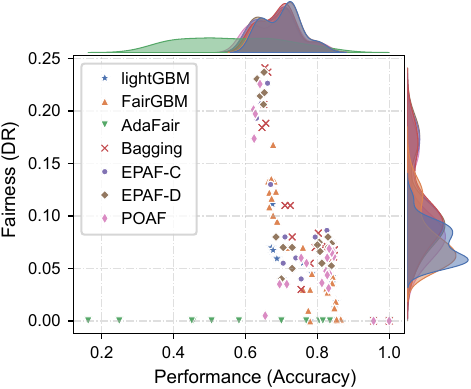}}
\\ \vspace{-4mm}
\subfloat[]{\label{subfig:aware,a}
\includegraphics[height=27mm]{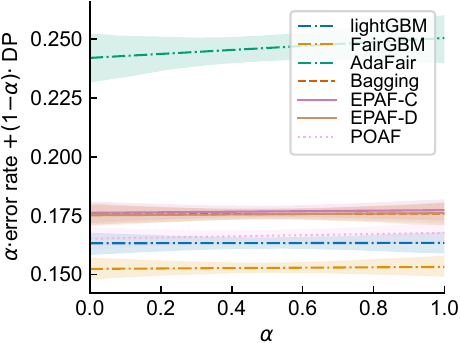}}
\hspace{4mm}
\subfloat[]{\label{subfig:aware,b}
\includegraphics[height=27mm]{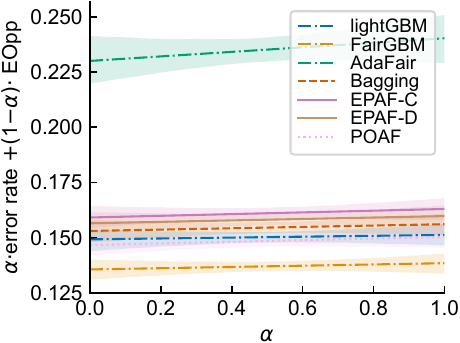}}
\hspace{4mm}
\subfloat[]{\label{subfig:aware,c}
\includegraphics[height=27mm]{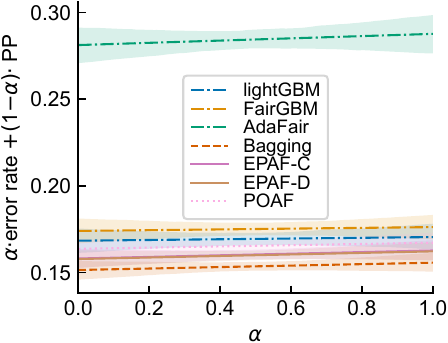}}
\hspace{4mm}
\subfloat[]{\label{subfig:aware,d}
\includegraphics[height=27mm]{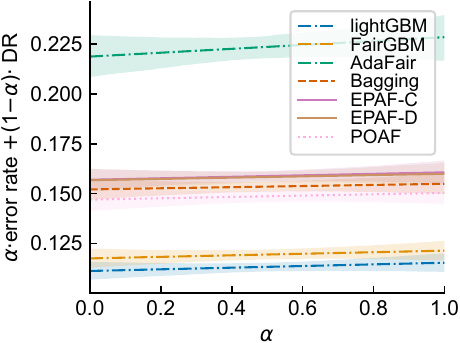}}
\vspace{-4mm}\caption{%
Comparison between \poepabbr{} and fairness-aware ensemble-based methods. 
(a--d) Scatter plots showing fairness and accuracy of each algorithm. 
(e--h) Plots of best test-set fairness-accuracy trade-offs per algorithm (the smaller the better) \citep{cruz2022fairgbm}, where fairness is \gls{dp}, \gls{eopp}, \gls{pqp}, and \emph{DR}, respectively; 
Lines show the mean value, and shades show 95\% confidence intervals. 
}\label{fig:aware}
\end{minipage}
\begin{minipage}{\textwidth}
\vspace{-2mm}\centering
\subfloat[]{\label{fig:aware,a}
\includegraphics[height=27mm]{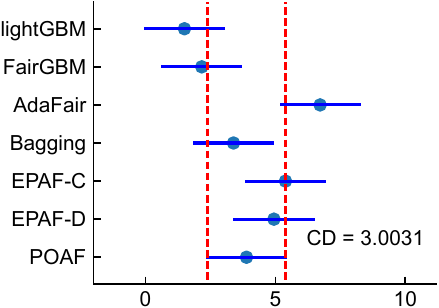}}
\hspace{3mm}
\subfloat[]{\label{fig:aware,b}
\includegraphics[height=28mm]{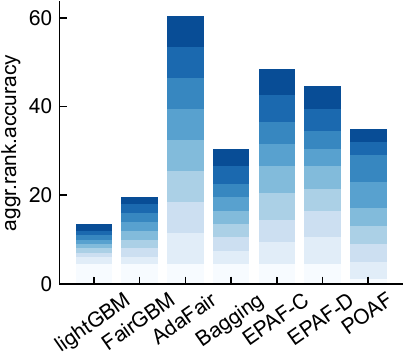}}
\hspace{3mm}
\subfloat[]{\label{fig:aware,c}
\includegraphics[height=27mm]{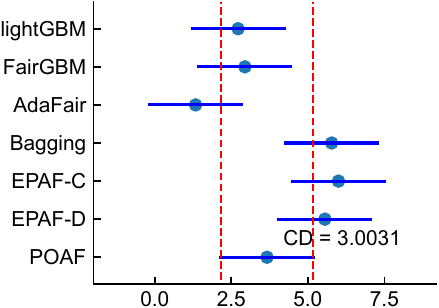}}
\hspace{3mm}
\subfloat[]{\label{fig:aware,d}
\includegraphics[height=28mm]{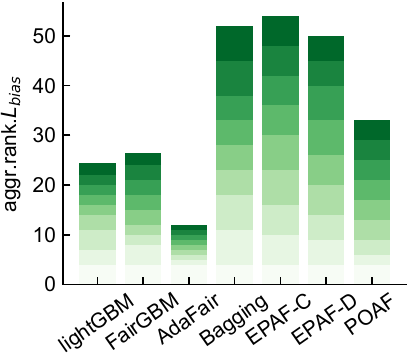}}
\\ \vspace{-1em}
\subfloat[]{\label{fig:aware,e}
\includegraphics[height=28mm]{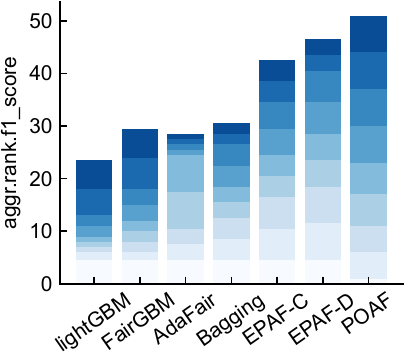}}
\hspace{5mm}
\subfloat[]{\label{fig:aware,f}
\includegraphics[height=28mm]{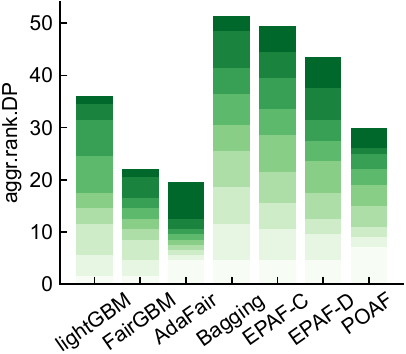}}
\hspace{5mm}
\subfloat[]{\label{fig:aware,g}
\includegraphics[height=28mm]{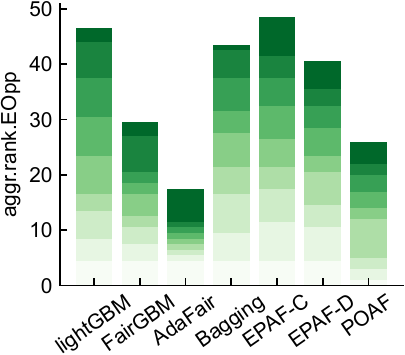}}
\hspace{5mm}
\subfloat[]{\label{fig:aware,h}
\includegraphics[height=28mm]{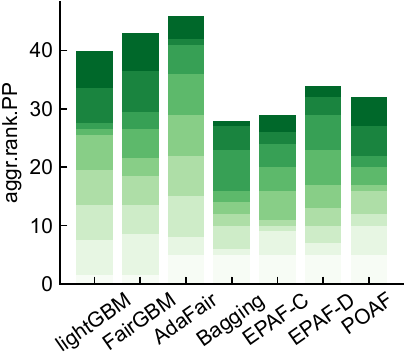}}
\vspace{-4mm}\caption{%
Comparison of the fairness-aware ensemble-based methods with \poepabbr{}. 
(a) Friedman test chart (non-overlapping means significant difference) on the test accuracy, which rejects the null hypothesis that ``all methods have the same evaluation performance'' at 5\% significance level, 
and where CD means the critical difference of average rank difference in Nemenyi post-hoc test \citep{zhou2021machine}; 
(b) The aggregated rank of each method (the smaller the better) \citep{qian2015pareto} on test accuracy; 
(c) Friedman test chart on \gls{dr}; (d) The aggregated rank of each method on \gls{dr}; 
(d--h) The aggregated rank on $\text{f}_1$ score, \gls{dp}, \gls{eopp}, and \gls{pqp}, respectively. 
}\label{appx,fig:aware}
\end{minipage}
\end{figure*}

\subsection{{\bf RQ2}: Validating the oracle bounds and their relaxation, as well as PAC bounds}
\label{expt:bounds}
In this subsection, we verify the proposed oracle bounds in \cref{method:1}, their relaxation in \cref{method:5}, and generalisation bounds in \cref{method:2}. The empirical results are reported in Figure~\ref{fig:bounds} in the form of scatter diagrams.

As we may see from Figures~\myref{fig:bounds}{subfig:bnd,a} to \myref{fig:bounds}{subfig:bnd,c}, all values of $\justLt(\mvrho)$ are smaller than the corresponding right hand sides of theorems~\ref{thm:1} to \ref{thm:4}, indicating their faithfulness, and \cref{thm:4} provides a tighter bound than theorems~\ref{thm:1} and \ref{thm:3}. 
Figure~\ref{fig:prev} presents a more analogous form of the first- and second-order oracle bounds to the work of \citet{masegosa2020second}, and we observe that our margin-dependent bounds have a higher coverage than these analogous forms. It also suggests that ensemble classifiers can normally help improve fairness in most cases. 

As for Figure~\myref{fig:bounds}{subfig:bnd,d}, $\mathbb{E}_\mathcal{D}[ \mathbb{E}_\rho[ \justL(f,\bm{x})]^2]$ remains identical to $\mathbb{E}_{\rho^2}[ \justLt(f,f^\prime)]$ in all cases, and Pearson correlation coefficient between them is one, indicating that \eqref{eq:4} in lemma~\ref{lem:2} is reliable. 
We verify corollaries~\ref{col:5} and \ref{col:6} in Figures~\myref{fig:bounds}{subfig:bnd,e} to \myref{fig:bounds}{subfig:bnd,f} and find that their relaxation forms are faithful as well. 
At last, Figures~\myref{fig:bounds}{subfig:bnd,g} and \myref{fig:bounds}{subfig:bnd,h} show that theorems~\ref{thm:n5} and \ref{thm:5} hold for most of the cases, in which the $x$-coordinates are computed on the training data and the $y$-coordinates are computed on the test data. Note that the $x$- and $y$-coordinates in all others subfigures of Figure~\ref{fig:bounds} and Figures~\ref{fig:prev} are computed on the test data.

\subsection{{\bf RQ3}: Comparison between \poepabbr{} and fairness-aware ensemble-based methods}
\label{expt:aware}

In this subsection, we validate the effectiveness of \poepabbr{} presented in \cref{method:3}, compared with existing fairness-aware ensemble-based methods. 
The experimental results are reported in Figures~\ref{fig:aware} and \ref{appx,fig:aware}, as well as Appendix~\ref{appx,expt:aware}.

As we can see from Figures~\ref{fig:aware}\subref{subfig:aware,a} to \ref{fig:aware}\subref{subfig:aware,d}, \poepabbr{} is significantly better than AdaFair. 
Meanwhile, \poepabbr{} in Figure~\ref{fig:aware}\subref{subfig:aware,b} is also slightly better than lightGBM, although it shows comparable performance yet is not better than FairGBM; \poepabbr{} in Figure~\ref{fig:aware}\subref{subfig:aware,c} is weakly better than lightGBM and FairGBM. 
This indicates that \poepabbr{} could achieve a good balance between accuracy and fairness and usually lead to satisfactory ensemble classifiers. 
Refer to Figures \ref{fig:aware}\subref{subfig:aware,e} to \ref{fig:aware}\subref{subfig:aware,h} for another way to express the results of balance between accuracy and fairness.

\begin{figure*}[tb]
\begin{minipage}{\textwidth}
\centering%
\subfloat[]{\label{fig:prev,rel6p}
\includegraphics[height=27mm]{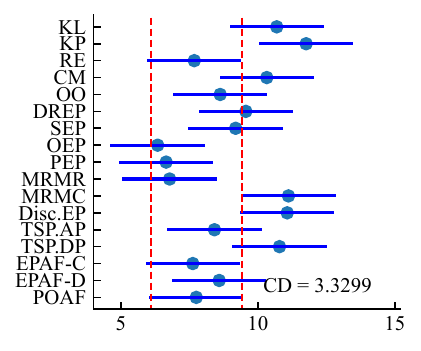}}
\hspace{3mm}
\subfloat[]{\label{fig:prev,rel7p}
\includegraphics[height=27mm]{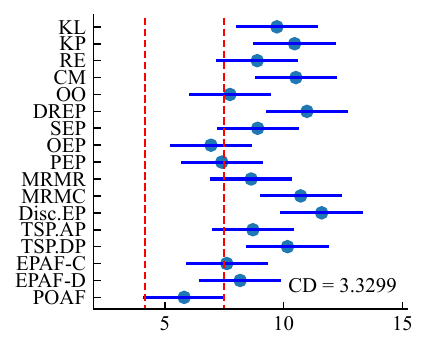}}
\hspace{3mm}
\subfloat[]{\label{fig:prev,rel1p}
\includegraphics[height=27mm]{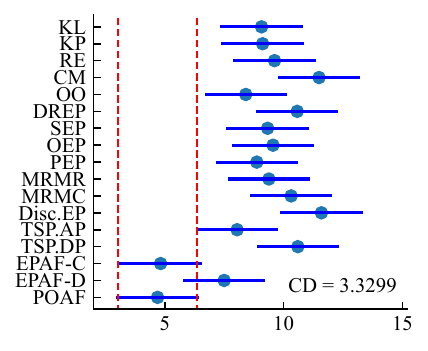}}
\hspace{3mm}
\subfloat[]{\label{fig:prev,rel1}
\includegraphics[height=27mm]{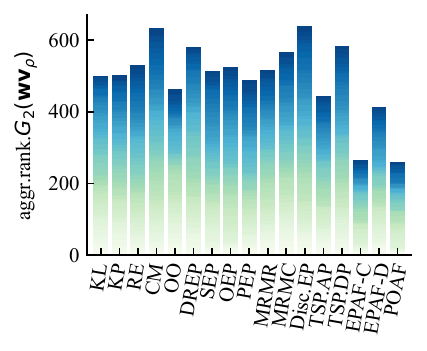}}
\\ \vspace{-4mm}
\subfloat[]{\label{fig:prev,rel6}
\includegraphics[height=27mm]{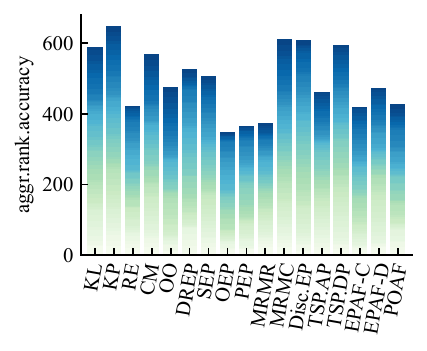}}
\hspace{3mm}
\subfloat[]{\label{fig:prev,rel7}
\includegraphics[height=27mm]{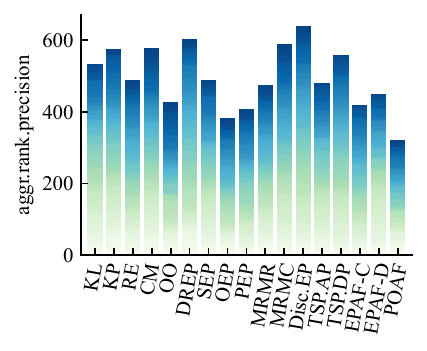}}
\hspace{3mm}
\subfloat[]{\label{fig:prev,rel8}
\includegraphics[height=27mm]{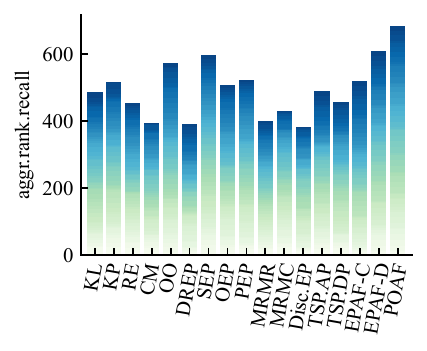}}
\hspace{3mm}
\subfloat[]{\label{fig:prev,rel2}
\includegraphics[height=27mm]{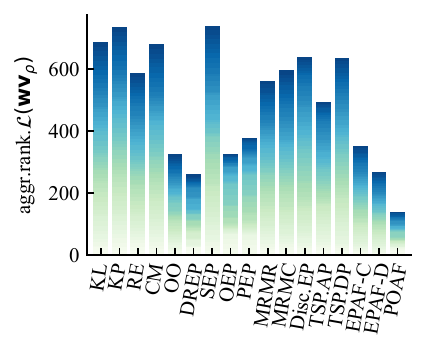}}
\vspace{-4mm}\caption{
Comparison of pruning method with \poepabbr{}, using bagging to conduct homogeneous ensembles. 
(a--c) Friedman test chart on test accuracy, precision, and $\justLt(\mvrho)$, respectively, of which each rejects the null hypothesis at 5\% significance level; 
(d--h) The aggregated rank for each pruning method over $\justLt(\mvrho)$, test accuracy, precision, recall, and $\justObjT(\mvrho)$, respectively.
}\label{fig:past,merge}
\end{minipage}
\end{figure*}

As for Figure~\ref{appx,fig:aware}\subref{fig:aware,a}, we also observe that \poepabbr{} is significantly better than AdaFair, although it shows comparable performance yet is not better than FairGBM. Figures~\ref{appx,fig:aware}\subref{fig:aware,c}--\ref{appx,fig:aware}\subref{fig:aware,d} and \ref{appx,fig:aware}\subref{fig:aware,f}--\ref{appx,fig:aware}\subref{fig:aware,g} indicate that the fairness level of \poepabbr{} is surely better than the corresponding unpruned ensembles (\aka{} bagging); besides, the fairness level of \poepabbr{} is also nearly the best, compared with FairGBM and AdaFair. Therefore, \poepabbr{} could be viewed as a method that improves fairness with acceptable accuracy damage. The same results are also reported in Tables~\ref{tab:aware,acc;part1} and \ref{tab:aware,acc;part2} of Appendix~\ref{appx,expt:aware}.

\subsection{{\bf RQ4}: Comparison between \poepabbr{} and ensemble pruning methods}
\label{expt:prus}

In this subsection, we validate the effectiveness of \poepabbr{} as an ensemble pruning method, in comparison with existing various pruning methods. 
Furthermore, we also validate the ensemble fairness after using \poepabbr{} to check if our primary goal of boosting fairness is fulfilled. 
The experimental results are reported in Figure~\ref{fig:past,merge} and Appendices~\ref{appx,expt:prus} to \ref{expt:lamb}.

Figure~\ref{fig:past,merge}\subref{fig:prev,rel6p} shows that \poepabbr{} could achieve at least the same level of test accuracy compared to baselines, and the same observation is also presented in Figures~\ref{fig:past,merge}\subref{fig:prev,rel7p}, \ref{fig:past,merge}\subref{fig:prev,rel6}, and \ref{fig:past,merge}\subref{fig:prev,rel7}. 
Besides, Figures~\ref{fig:past,merge}\subref{fig:prev,rel1p} and \ref{fig:past,merge}\subref{fig:prev,rel1} indicate that \poepabbr{} achieves the best fairness quality among all opponents, while Figure~\ref{fig:past,merge}\subref{fig:prev,rel2} shows that \poepabbr{} achieves the best performance on $\justObjT(\mvrho)$.

\paragraph{Why we may still need \greedy{} and \ddismi{} sometime, given that we already have \poepabbr{}} 
As we mentioned before in \cref{method:3}, 
\greedy{} and \ddismi{} are easily implemented, and \poepabbr{} is not the only way to achieve a strong sub-ensemble. 
Besides, \poepabbr{} usually takes a long time to reach the final sub-ensemble, shown in Fig.~\ref{fig:sota,us}\subref{subfig:us,2} of Appendix~\ref{expt:lamb}, 
while the time cost of \greedy{} and \ddismi{} is significantly shorter than that of \poepabbr{}. Two major reasons why \poepabbr{} costs longer time are that: one is \poepabbr{} cannot fix the size of the pruned sub-ensembles like \greedy{} and \ddismi{} do, the other is the sub-route of updating $\mathcal{P}$ in \poepabbr{} is more costly than finding one individual classifier for the minimal objective in \greedy{}. 
Also, we can see from Fig.~\ref{fig:past,merge}\subref{fig:prev,rel6p} and \ref{fig:past,merge}\subref{fig:prev,rel1p} that \greedy{} and \ddismi{} achieve good fairness as well meanwhile maintaining comparable accuracy performance. 
Yet \greedy{} and \ddismi{} may not be able to achieve the best trade-off between fairness and accuracy, shown in Fig.~\ref{fig:aware}.

\subsection{Limitations and discussion}
\label{expt:limit}

Discrimination mitigation techniques are meaningful given the wide applications of \gls{ml} models nowadays; therefore, bringing such a technique with learning guarantees matters as it could provide theoretical foundations to boost fairness without potentially vain and repetitive practical attempts. 
In this view, our work throws away a brick in order to get a gem, showing that fairness can indeed be boosted with learning guarantees instead of being dependent on specific (hyper-)parameters. 
The proposed \gls{dr} measure and the proposed oracle bounds are suitable for both binary and multi-class classification, enlarging the applicable fields, which is advantageous. 
However, there are also limitations in the proposed measure and the pruning method \poepabbr{}. 
For instance, the computation of \gls{dr} is relevant to the perturbed \glspl{sa}, which means a randomness factor exists and may affect computational results somehow. 
Besides, although \poepabbr{} could achieve acceptable performance, its time cost is relatively high compared with the baselines, which may restrict its applicable scenarios if the response time is one of the major criteria. 
Therefore, the effect of randomness on \gls{dr} and acceleration of \poepabbr{} are worth exploring in the future.

\section{Conclusion}
We have presented a novel analysis of the expected fairness quality via weighted voting and demonstrated that the ensemble combination may help improve fairness sometimes. 
In other words, increasing the classification margin in boosted ensembles tightens our derived bounds on unfairness (as measured by \gls{dr}), lending support to the observation that ``fairness may be improved via ensemble combination.'' 
We have also shown that \poepabbr{} based on domination and Pareto optimality can fulfil the expectations of achieving fairer ensembles without much accuracy decline, confirmed by extensive empirical results. 
Our work shows that the ensemble combination could boost fairness with theoretical learning guarantees, which is helpful to save some fruitless efforts on (hyper-)parameter tuning.

\section*{Generative AI Disclosure Statement}
ChatGPT (models available for free) was utilised to refine partial texts in this manuscript, to make them more concise.

\begin{acks}
This research is funded by the European Union (MSCA, FairML, project no. 101106768). Views and opinions expressed are those of the author(s) only and do not necessarily reflect those of the European Union or the Research Executive Agency. Neither the European Union nor the granting authority can be held responsible for them.

We thank the anonymous reviewers for helpful comments. 
We also thank Dr. Lei You at the Technical University of Denmark for discussions that helped us revise the theoretical bounds; Prof. Yevgeny Seldin at the University of Copenhagen and Dr. Kun Zhang at the Hefei University of Technology for discussions at early stages of this work; Prof. Anqi Qiu at the National University of Singapore for pointing out the importance of the \gls{dr} metric, which led us to develop this part of the paper further; Prof. Shai Ben-David at the University of Waterloo for comments on earlier versions; Dr. Yi-Shan Wu, Prof. Sadegh Talebi and Prof. Melanie Ganz-Benjaminsen at the University of Copenhagen for advice regarding later stages of the manuscript; the members of CopeNLU at the University of Copenhagen for their comments on this version.
\end{acks}

\bibliographystyle{ACM-Reference-Format}
\bibliography{nus_title_iso4,refsmac}


\begin{thebibliography}{66}


\ifx \showCODEN    \undefined \def \showCODEN     #1{\unskip}     \fi
\ifx \showISBNx    \undefined \def \showISBNx     #1{\unskip}     \fi
\ifx \showISBNxiii \undefined \def \showISBNxiii  #1{\unskip}     \fi
\ifx \showISSN     \undefined \def \showISSN      #1{\unskip}     \fi
\ifx \showLCCN     \undefined \def \showLCCN      #1{\unskip}     \fi
\ifx \shownote     \undefined \def \shownote      #1{#1}          \fi
\ifx \showarticletitle \undefined \def \showarticletitle #1{#1}   \fi
\ifx \showURL      \undefined \def \showURL       {\relax}        \fi
\providecommand\bibfield[2]{#2}
\providecommand\bibinfo[2]{#2}
\providecommand\natexlab[1]{#1}
\providecommand\showeprint[2][]{arXiv:#2}

\bibitem[Agarwal et~al\mbox{.}(2018)]%
        {agarwal2018reductions}
\bibfield{author}{\bibinfo{person}{Alekh Agarwal}, \bibinfo{person}{Alina
  Beygelzimer}, \bibinfo{person}{Miroslav Dud{\'\i}k}, \bibinfo{person}{John
  Langford}, {and} \bibinfo{person}{Hanna Wallach}.}
  \bibinfo{year}{2018}\natexlab{}.
\newblock \showarticletitle{A reductions approach to fair classification}. In
  \bibinfo{booktitle}{\emph{ICML}}, Vol.~\bibinfo{volume}{80}.
  \bibinfo{publisher}{PMLR}, \bibinfo{pages}{60--69}.
\newblock


\bibitem[Agarwal et~al\mbox{.}(2019)]%
        {agarwal2019fair}
\bibfield{author}{\bibinfo{person}{Alekh Agarwal}, \bibinfo{person}{Miroslav
  Dud{\'\i}k}, {and} \bibinfo{person}{Zhiwei~Steven Wu}.}
  \bibinfo{year}{2019}\natexlab{}.
\newblock \showarticletitle{Fair regression: Quantitative definitions and
  reduction-based algorithms}. In \bibinfo{booktitle}{\emph{ICML}},
  Vol.~\bibinfo{volume}{97}. \bibinfo{publisher}{PMLR},
  \bibinfo{pages}{120--129}.
\newblock


\bibitem[Backurs et~al\mbox{.}(2019)]%
        {backurs2019scalable}
\bibfield{author}{\bibinfo{person}{Arturs Backurs}, \bibinfo{person}{Piotr
  Indyk}, \bibinfo{person}{Krzysztof Onak}, \bibinfo{person}{Baruch Schieber},
  \bibinfo{person}{Ali Vakilian}, {and} \bibinfo{person}{Tal Wagner}.}
  \bibinfo{year}{2019}\natexlab{}.
\newblock \showarticletitle{Scalable fair clustering}. In
  \bibinfo{booktitle}{\emph{ICML}}, Vol.~\bibinfo{volume}{97}.
  \bibinfo{publisher}{PMLR}, \bibinfo{pages}{405--413}.
\newblock


\bibitem[Barocas et~al\mbox{.}(2023)]%
        {barocas2023fairness}
\bibfield{author}{\bibinfo{person}{Solon Barocas}, \bibinfo{person}{Moritz
  Hardt}, {and} \bibinfo{person}{Arvind Narayanan}.}
  \bibinfo{year}{2023}\natexlab{}.
\newblock \bibinfo{booktitle}{\emph{Fairness and machine learning: Limitations
  and opportunities}}.
\newblock \bibinfo{publisher}{MIT Press}, \bibinfo{address}{Cambridge, MA,
  USA}.
\newblock


\bibitem[Bechavod and Ligett(2017)]%
        {bechavod2017penalizing}
\bibfield{author}{\bibinfo{person}{Yahav Bechavod} {and}
  \bibinfo{person}{Katrina Ligett}.} \bibinfo{year}{2017}\natexlab{}.
\newblock \showarticletitle{Penalizing unfairness in binary classification}.
\newblock \bibinfo{journal}{\emph{arXiv preprint arXiv:1707.00044}}
  (\bibinfo{year}{2017}).
\newblock


\bibitem[Berk et~al\mbox{.}(2021)]%
        {berk2021fairness}
\bibfield{author}{\bibinfo{person}{Richard Berk}, \bibinfo{person}{Hoda
  Heidari}, \bibinfo{person}{Shahin Jabbari}, \bibinfo{person}{Michael Kearns},
  {and} \bibinfo{person}{Aaron Roth}.} \bibinfo{year}{2021}\natexlab{}.
\newblock \showarticletitle{Fairness in criminal justice risk assessments: The
  state of the art}.
\newblock \bibinfo{journal}{\emph{Sociol Methods Res}} \bibinfo{volume}{50},
  \bibinfo{number}{1} (\bibinfo{year}{2021}), \bibinfo{pages}{3--44}.
\newblock


\bibitem[Bian et~al\mbox{.}(2020)]%
        {bian2019ensemble}
\bibfield{author}{\bibinfo{person}{Yijun Bian}, \bibinfo{person}{Yijun Wang},
  \bibinfo{person}{Yaqiang Yao}, {and} \bibinfo{person}{Huanhuan Chen}.}
  \bibinfo{year}{2020}\natexlab{}.
\newblock \showarticletitle{Ensemble pruning based on objection maximization
  with a general distributed framework}.
\newblock \bibinfo{journal}{\emph{IEEE Trans Neural Netw Learn Syst}}
  \bibinfo{volume}{31}, \bibinfo{number}{9} (\bibinfo{year}{2020}),
  \bibinfo{pages}{3766--3774}.
\newblock
\href{https://doi.org/10.1109/TNNLS.2019.2945116}{doi:\nolinkurl{10.1109/TNNLS.2019.2945116}}


\bibitem[Breiman(1996)]%
        {breiman1996bagging}
\bibfield{author}{\bibinfo{person}{Leo Breiman}.}
  \bibinfo{year}{1996}\natexlab{}.
\newblock \showarticletitle{Bagging predictors}.
\newblock \bibinfo{journal}{\emph{Mach Learn}} \bibinfo{volume}{24},
  \bibinfo{number}{2} (\bibinfo{year}{1996}), \bibinfo{pages}{123--140}.
\newblock


\bibitem[Calmon et~al\mbox{.}(2017)]%
        {calmon2017optimized}
\bibfield{author}{\bibinfo{person}{Flavio Calmon}, \bibinfo{person}{Dennis
  Wei}, \bibinfo{person}{Bhanukiran Vinzamuri}, \bibinfo{person}{Karthikeyan
  Natesan~Ramamurthy}, {and} \bibinfo{person}{Kush~R Varshney}.}
  \bibinfo{year}{2017}\natexlab{}.
\newblock \showarticletitle{Optimized pre-processing for discrimination
  prevention}. In \bibinfo{booktitle}{\emph{NIPS}}, Vol.~\bibinfo{volume}{30}.
  \bibinfo{publisher}{Curran Associates, Inc.}
\newblock


\bibitem[Cao et~al\mbox{.}(2018)]%
        {cao2018optimizing}
\bibfield{author}{\bibinfo{person}{Jingjing Cao}, \bibinfo{person}{Wenfeng Li},
  \bibinfo{person}{Congcong Ma}, {and} \bibinfo{person}{Zhiwen Tao}.}
  \bibinfo{year}{2018}\natexlab{}.
\newblock \showarticletitle{Optimizing multi-sensor deployment via ensemble
  pruning for wearable activity recognition}.
\newblock \bibinfo{journal}{\emph{Inf Fusion}}  \bibinfo{volume}{41}
  (\bibinfo{year}{2018}), \bibinfo{pages}{68--79}.
\newblock


\bibitem[Chierichetti et~al\mbox{.}(2017)]%
        {chierichetti2017fair}
\bibfield{author}{\bibinfo{person}{Flavio Chierichetti}, \bibinfo{person}{Ravi
  Kumar}, \bibinfo{person}{Silvio Lattanzi}, {and} \bibinfo{person}{Sergei
  Vassilvitskii}.} \bibinfo{year}{2017}\natexlab{}.
\newblock \showarticletitle{Fair clustering through fairlets}. In
  \bibinfo{booktitle}{\emph{NIPS}}, Vol.~\bibinfo{volume}{30}.
  \bibinfo{publisher}{Curran Associates, Inc.}
\newblock


\bibitem[Chouldechova(2017)]%
        {chouldechova2017fair}
\bibfield{author}{\bibinfo{person}{Alexandra Chouldechova}.}
  \bibinfo{year}{2017}\natexlab{}.
\newblock \showarticletitle{Fair prediction with disparate impact: A study of
  bias in recidivism prediction instruments}.
\newblock \bibinfo{journal}{\emph{Big Data}} \bibinfo{volume}{5},
  \bibinfo{number}{2} (\bibinfo{year}{2017}), \bibinfo{pages}{153--163}.
\newblock


\bibitem[Claucich et~al\mbox{.}(2025)]%
        {claucich2025fairness}
\bibfield{author}{\bibinfo{person}{Estanislao Claucich}, \bibinfo{person}{Sara
  Hooker}, \bibinfo{person}{Diego~H Milone}, \bibinfo{person}{Enzo Ferrante},
  {and} \bibinfo{person}{Rodrigo Echeveste}.} \bibinfo{year}{2025}\natexlab{}.
\newblock \showarticletitle{Fairness of Deep Ensembles: On the interplay
  between per-group task difficulty and under-representation}. In
  \bibinfo{booktitle}{\emph{FAccT}}. \bibinfo{pages}{3138--3147}.
\newblock


\bibitem[Corbett-Davies et~al\mbox{.}(2017)]%
        {corbett2017algorithmic}
\bibfield{author}{\bibinfo{person}{Sam Corbett-Davies}, \bibinfo{person}{Emma
  Pierson}, \bibinfo{person}{Avi Feller}, \bibinfo{person}{Sharad Goel}, {and}
  \bibinfo{person}{Aziz Huq}.} \bibinfo{year}{2017}\natexlab{}.
\newblock \showarticletitle{Algorithmic decision making and the cost of
  fairness}. In \bibinfo{booktitle}{\emph{SIGKDD}}. \bibinfo{publisher}{ACM},
  \bibinfo{address}{New York, NY, USA}, \bibinfo{pages}{797--806}.
\newblock
\showISBNx{9781450348874}


\bibitem[Cruz et~al\mbox{.}(2023)]%
        {cruz2022fairgbm}
\bibfield{author}{\bibinfo{person}{Andr{\'e}~F Cruz}, \bibinfo{person}{Catarina
  Bel{\'e}m}, \bibinfo{person}{Jo{\~a}o Bravo}, \bibinfo{person}{Pedro
  Saleiro}, {and} \bibinfo{person}{Pedro Bizarro}.}
  \bibinfo{year}{2023}\natexlab{}.
\newblock \showarticletitle{FairGBM: Gradient Boosting with Fairness
  Constraints}. In \bibinfo{booktitle}{\emph{ICLR}}.
\newblock


\bibitem[Dem{\v{s}}ar(2006)]%
        {demvsar2006statistical}
\bibfield{author}{\bibinfo{person}{Janez Dem{\v{s}}ar}.}
  \bibinfo{year}{2006}\natexlab{}.
\newblock \showarticletitle{Statistical comparisons of classifiers over
  multiple data sets}.
\newblock \bibinfo{journal}{\emph{J Mach Learn Res}}  \bibinfo{volume}{7}
  (\bibinfo{year}{2006}), \bibinfo{pages}{1--30}.
\newblock


\bibitem[Dwork et~al\mbox{.}(2012)]%
        {dwork2012fairness}
\bibfield{author}{\bibinfo{person}{Cynthia Dwork}, \bibinfo{person}{Moritz
  Hardt}, \bibinfo{person}{Toniann Pitassi}, \bibinfo{person}{Omer Reingold},
  {and} \bibinfo{person}{Richard Zemel}.} \bibinfo{year}{2012}\natexlab{}.
\newblock \showarticletitle{Fairness through awareness}. In
  \bibinfo{booktitle}{\emph{ITCS}} (Cambridge, Massachusetts)
  \emph{(\bibinfo{series}{ITCS '12})}. \bibinfo{publisher}{ACM},
  \bibinfo{address}{New York, NY, USA}, \bibinfo{pages}{214--226}.
\newblock
\showISBNx{9781450311151}


\bibitem[Dwork and Ilvento(2018)]%
        {dwork2018fairness}
\bibfield{author}{\bibinfo{person}{Cynthia Dwork} {and}
  \bibinfo{person}{Christina Ilvento}.} \bibinfo{year}{2018}\natexlab{}.
\newblock \showarticletitle{Fairness under composition}. In
  \bibinfo{booktitle}{\emph{ITCS}}. Schloss Dagstuhl-Leibniz-Zentrum fuer
  Informatik.
\newblock


\bibitem[Dwork et~al\mbox{.}(2018)]%
        {dwork2018decoupled}
\bibfield{author}{\bibinfo{person}{Cynthia Dwork}, \bibinfo{person}{Nicole
  Immorlica}, \bibinfo{person}{Adam~Tauman Kalai}, {and} \bibinfo{person}{Max
  Leiserson}.} \bibinfo{year}{2018}\natexlab{}.
\newblock \showarticletitle{Decoupled classifiers for group-fair and efficient
  machine learning}. In \bibinfo{booktitle}{\emph{FAT}},
  Vol.~\bibinfo{volume}{81}. \bibinfo{publisher}{PMLR},
  \bibinfo{pages}{119--133}.
\newblock


\bibitem[Feldman et~al\mbox{.}(2015)]%
        {feldman2015certifying}
\bibfield{author}{\bibinfo{person}{Michael Feldman}, \bibinfo{person}{Sorelle~A
  Friedler}, \bibinfo{person}{John Moeller}, \bibinfo{person}{Carlos
  Scheidegger}, {and} \bibinfo{person}{Suresh Venkatasubramanian}.}
  \bibinfo{year}{2015}\natexlab{}.
\newblock \showarticletitle{Certifying and removing disparate impact}. In
  \bibinfo{booktitle}{\emph{SIGKDD}}. \bibinfo{publisher}{Association for
  Computing Machinery}, \bibinfo{address}{New York, NY, USA},
  \bibinfo{pages}{259--268}.
\newblock


\bibitem[Freund(1995)]%
        {freund1995boosting}
\bibfield{author}{\bibinfo{person}{Yoav Freund}.}
  \bibinfo{year}{1995}\natexlab{}.
\newblock \showarticletitle{Boosting a weak learning algorithm by majority}.
\newblock \bibinfo{journal}{\emph{Inf Comput}} \bibinfo{volume}{121},
  \bibinfo{number}{2} (\bibinfo{year}{1995}), \bibinfo{pages}{256--285}.
\newblock


\bibitem[Freund et~al\mbox{.}(1996)]%
        {freund1996experiments}
\bibfield{author}{\bibinfo{person}{Yoav Freund}, \bibinfo{person}{Robert~E
  Schapire}, {et~al\mbox{.}}} \bibinfo{year}{1996}\natexlab{}.
\newblock \showarticletitle{Experiments with a new boosting algorithm}. In
  \bibinfo{booktitle}{\emph{ICML}}, Vol.~\bibinfo{volume}{96}. Citeseer,
  \bibinfo{pages}{148--156}.
\newblock


\bibitem[Friedler et~al\mbox{.}(2019)]%
        {friedler2019comparative}
\bibfield{author}{\bibinfo{person}{Sorelle~A Friedler}, \bibinfo{person}{Carlos
  Scheidegger}, \bibinfo{person}{Suresh Venkatasubramanian},
  \bibinfo{person}{Sonam Choudhary}, \bibinfo{person}{Evan~P Hamilton}, {and}
  \bibinfo{person}{Derek Roth}.} \bibinfo{year}{2019}\natexlab{}.
\newblock \showarticletitle{A comparative study of fairness-enhancing
  interventions in machine learning}. In \bibinfo{booktitle}{\emph{FAT}}.
  \bibinfo{publisher}{Association for Computing Machinery},
  \bibinfo{address}{New York, NY, USA}, \bibinfo{pages}{329--338}.
\newblock


\bibitem[Gajane and Pechenizkiy(2018)]%
        {gajane2017formalizing}
\bibfield{author}{\bibinfo{person}{Pratik Gajane} {and} \bibinfo{person}{Mykola
  Pechenizkiy}.} \bibinfo{year}{2018}\natexlab{}.
\newblock \showarticletitle{On formalizing fairness in prediction with machine
  learning}. In \bibinfo{booktitle}{\emph{FAT/ML}}.
\newblock


\bibitem[Grgi{\'c}-Hla{\v{c}}a et~al\mbox{.}(2017)]%
        {grgic2017fairness}
\bibfield{author}{\bibinfo{person}{Nina Grgi{\'c}-Hla{\v{c}}a},
  \bibinfo{person}{Muhammad~Bilal Zafar}, \bibinfo{person}{Krishna~P Gummadi},
  {and} \bibinfo{person}{Adrian Weller}.} \bibinfo{year}{2017}\natexlab{}.
\newblock \showarticletitle{On fairness, diversity and randomness in
  algorithmic decision making}. In \bibinfo{booktitle}{\emph{FAT/ML}}.
\newblock


\bibitem[Grgi{\'c}-Hla{\v{c}}a et~al\mbox{.}(2018)]%
        {grgic2018beyond}
\bibfield{author}{\bibinfo{person}{Nina Grgi{\'c}-Hla{\v{c}}a},
  \bibinfo{person}{Muhammad~Bilal Zafar}, \bibinfo{person}{Krishna~P Gummadi},
  {and} \bibinfo{person}{Adrian Weller}.} \bibinfo{year}{2018}\natexlab{}.
\newblock \showarticletitle{Beyond distributive fairness in algorithmic
  decision making: Feature selection for procedurally fair learning}. In
  \bibinfo{booktitle}{\emph{AAAI}}, Vol.~\bibinfo{volume}{32}.
\newblock


\bibitem[Haas(2019)]%
        {haas2019price}
\bibfield{author}{\bibinfo{person}{Christian Haas}.}
  \bibinfo{year}{2019}\natexlab{}.
\newblock \showarticletitle{The price of fairness - A framework to explore
  trade-offs in algorithmic fairness}. In \bibinfo{booktitle}{\emph{ICIS}}.
  Association for Information Systems.
\newblock


\bibitem[Hardt et~al\mbox{.}(2016)]%
        {hardt2016equality}
\bibfield{author}{\bibinfo{person}{Moritz Hardt}, \bibinfo{person}{Eric Price},
  {and} \bibinfo{person}{Nathan Srebro}.} \bibinfo{year}{2016}\natexlab{}.
\newblock \showarticletitle{Equality of opportunity in supervised learning}. In
  \bibinfo{booktitle}{\emph{NIPS}}, Vol.~\bibinfo{volume}{29}.
  \bibinfo{publisher}{Curran Associates Inc.}, \bibinfo{address}{Red Hook, NY,
  USA}, \bibinfo{pages}{3323--3331}.
\newblock


\bibitem[Iosifidis and Ntoutsi(2019)]%
        {iosifidis2019adafair}
\bibfield{author}{\bibinfo{person}{Vasileios Iosifidis} {and}
  \bibinfo{person}{Eirini Ntoutsi}.} \bibinfo{year}{2019}\natexlab{}.
\newblock \showarticletitle{AdaFair: Cumulative fairness adaptive boosting}. In
  \bibinfo{booktitle}{\emph{CIKM}}. \bibinfo{publisher}{ACM},
  \bibinfo{address}{New York, NY, USA}, \bibinfo{pages}{781--790}.
\newblock


\bibitem[Jiang et~al\mbox{.}(2020)]%
        {jiang2020wasserstein}
\bibfield{author}{\bibinfo{person}{Ray Jiang}, \bibinfo{person}{Aldo
  Pacchiano}, \bibinfo{person}{Tom Stepleton}, \bibinfo{person}{Heinrich
  Jiang}, {and} \bibinfo{person}{Silvia Chiappa}.}
  \bibinfo{year}{2020}\natexlab{}.
\newblock \showarticletitle{Wasserstein fair classification}. In
  \bibinfo{booktitle}{\emph{UAI}}. PMLR, \bibinfo{pages}{862--872}.
\newblock


\bibitem[Joseph et~al\mbox{.}(2016)]%
        {joseph2016fairness}
\bibfield{author}{\bibinfo{person}{Matthew Joseph}, \bibinfo{person}{Michael
  Kearns}, \bibinfo{person}{Jamie~H Morgenstern}, {and} \bibinfo{person}{Aaron
  Roth}.} \bibinfo{year}{2016}\natexlab{}.
\newblock \showarticletitle{Fairness in learning: Classic and contextual
  bandits}. In \bibinfo{booktitle}{\emph{NIPS}}, Vol.~\bibinfo{volume}{29}.
  \bibinfo{publisher}{Curran Associates, Inc.}
\newblock


\bibitem[Kamiran and Calders(2012)]%
        {kamiran2012data}
\bibfield{author}{\bibinfo{person}{Faisal Kamiran} {and} \bibinfo{person}{Toon
  Calders}.} \bibinfo{year}{2012}\natexlab{}.
\newblock \showarticletitle{Data preprocessing techniques for classification
  without discrimination}.
\newblock \bibinfo{journal}{\emph{Knowl Inf Syst}} \bibinfo{volume}{33},
  \bibinfo{number}{1} (\bibinfo{year}{2012}), \bibinfo{pages}{1--33}.
\newblock


\bibitem[Kamiran et~al\mbox{.}(2010)]%
        {kamiran2010discrimination}
\bibfield{author}{\bibinfo{person}{Faisal Kamiran}, \bibinfo{person}{Toon
  Calders}, {and} \bibinfo{person}{Mykola Pechenizkiy}.}
  \bibinfo{year}{2010}\natexlab{}.
\newblock \showarticletitle{Discrimination aware decision tree learning}. In
  \bibinfo{booktitle}{\emph{ICDM}}. \bibinfo{publisher}{IEEE},
  \bibinfo{pages}{869--874}.
\newblock


\bibitem[Kamishima et~al\mbox{.}(2012)]%
        {kamishima2012fairness}
\bibfield{author}{\bibinfo{person}{Toshihiro Kamishima},
  \bibinfo{person}{Shotaro Akaho}, \bibinfo{person}{Hideki Asoh}, {and}
  \bibinfo{person}{Jun Sakuma}.} \bibinfo{year}{2012}\natexlab{}.
\newblock \showarticletitle{Fairness-aware classifier with prejudice remover
  regularizer}. In \bibinfo{booktitle}{\emph{ECML-PKDD}}.
  \bibinfo{publisher}{Springer Berlin Heidelberg}, \bibinfo{address}{Berlin,
  Heidelberg}, \bibinfo{pages}{35--50}.
\newblock


\bibitem[Ke et~al\mbox{.}(2017)]%
        {ke2017lightgbm}
\bibfield{author}{\bibinfo{person}{Guolin Ke}, \bibinfo{person}{Qi Meng},
  \bibinfo{person}{Thomas Finley}, \bibinfo{person}{Taifeng Wang},
  \bibinfo{person}{Wei Chen}, \bibinfo{person}{Weidong Ma},
  \bibinfo{person}{Qiwei Ye}, {and} \bibinfo{person}{Tie-Yan Liu}.}
  \bibinfo{year}{2017}\natexlab{}.
\newblock \showarticletitle{LightGBM: A highly efficient gradient boosting
  decision tree}. In \bibinfo{booktitle}{\emph{NIPS}},
  Vol.~\bibinfo{volume}{30}. \bibinfo{pages}{3146--3154}.
\newblock


\bibitem[Kilbertus et~al\mbox{.}(2017)]%
        {kilbertus2017avoiding}
\bibfield{author}{\bibinfo{person}{Niki Kilbertus}, \bibinfo{person}{Mateo
  Rojas~Carulla}, \bibinfo{person}{Giambattista Parascandolo},
  \bibinfo{person}{Moritz Hardt}, \bibinfo{person}{Dominik Janzing}, {and}
  \bibinfo{person}{Bernhard Sch{\"o}lkopf}.} \bibinfo{year}{2017}\natexlab{}.
\newblock \showarticletitle{Avoiding discrimination through causal reasoning}.
  In \bibinfo{booktitle}{\emph{NIPS}}, Vol.~\bibinfo{volume}{30}.
\newblock


\bibitem[Ko et~al\mbox{.}(2023)]%
        {ko2023fair}
\bibfield{author}{\bibinfo{person}{Wei-Yin Ko}, \bibinfo{person}{Daniel
  D'souza}, \bibinfo{person}{Karina Nguyen}, \bibinfo{person}{Randall
  Balestriero}, {and} \bibinfo{person}{Sara Hooker}.}
  \bibinfo{year}{2023}\natexlab{}.
\newblock \showarticletitle{Fair-ensemble: When fairness naturally emerges from
  deep ensembling}.
\newblock \bibinfo{journal}{\emph{arXiv preprint arXiv:2303.00586}}
  (\bibinfo{year}{2023}).
\newblock


\bibitem[Kusner et~al\mbox{.}(2017)]%
        {kusner2017counterfactual}
\bibfield{author}{\bibinfo{person}{Matt~J Kusner}, \bibinfo{person}{Joshua
  Loftus}, \bibinfo{person}{Chris Russell}, {and} \bibinfo{person}{Ricardo
  Silva}.} \bibinfo{year}{2017}\natexlab{}.
\newblock \showarticletitle{Counterfactual fairness}. In
  \bibinfo{booktitle}{\emph{NIPS}}, Vol.~\bibinfo{volume}{30}. NIPS
  Proceedings, \bibinfo{pages}{4069--4079}.
\newblock


\bibitem[Li et~al\mbox{.}(2012)]%
        {li2012diversity}
\bibfield{author}{\bibinfo{person}{N Li}, \bibinfo{person}{Y Yu}, {and}
  \bibinfo{person}{Z-H Zhou}.} \bibinfo{year}{2012}\natexlab{}.
\newblock \showarticletitle{Diversity Regularized Ensemble Pruning}. In
  \bibinfo{booktitle}{\emph{ECML-PKDD}}. \bibinfo{pages}{330--345}.
\newblock
\showISBNx{978-3-642-33460-3}


\bibitem[Louizos et~al\mbox{.}(2016)]%
        {louizos2016variational}
\bibfield{author}{\bibinfo{person}{Christos Louizos}, \bibinfo{person}{Kevin
  Swersky}, \bibinfo{person}{Yujia Li}, \bibinfo{person}{Max Welling}, {and}
  \bibinfo{person}{Richard~S Zemel}.} \bibinfo{year}{2016}\natexlab{}.
\newblock \showarticletitle{The variational fair autoencoder}. In
  \bibinfo{booktitle}{\emph{ICLR}}.
\newblock


\bibitem[Margineantu and Dietterich(1997)]%
        {margineantu1997pruning}
\bibfield{author}{\bibinfo{person}{DD Margineantu} {and} \bibinfo{person}{TG
  Dietterich}.} \bibinfo{year}{1997}\natexlab{}.
\newblock \showarticletitle{Pruning Adaptive Boosting}. In
  \bibinfo{booktitle}{\emph{ICML}}, Vol.~\bibinfo{volume}{97}.
  \bibinfo{pages}{211--218}.
\newblock


\bibitem[Mart{\'\i}nez-Mu{\~n}oz and Su{\'a}rez(2006)]%
        {martinez2006pruning}
\bibfield{author}{\bibinfo{person}{Mart{\'\i}nez-Mu{\~n}oz} {and}
  \bibinfo{person}{Su{\'a}rez}.} \bibinfo{year}{2006}\natexlab{}.
\newblock \showarticletitle{Pruning in Ordered Bagging Ensembles}. In
  \bibinfo{booktitle}{\emph{ICML}} (Pittsburgh, Pennsylvania, USA).
  \bibinfo{pages}{609--616}.
\newblock
\showISBNx{1-59593-383-2}
\href{https://doi.org/10.1145/1143844.1143921}{doi:\nolinkurl{10.1145/1143844.1143921}}


\bibitem[Mart{\'\i}nez-Mu{\~n}oz et~al\mbox{.}(2009)]%
        {martinez2009analysis}
\bibfield{author}{\bibinfo{person}{G Mart{\'\i}nez-Mu{\~n}oz},
  \bibinfo{person}{D Hern{\'a}ndez-Lobato}, {and} \bibinfo{person}{A
  Su{\'a}rez}.} \bibinfo{year}{2009}\natexlab{}.
\newblock \showarticletitle{An Analysis of Ensemble Pruning Techniques Based on
  Ordered Aggregation}.
\newblock \bibinfo{journal}{\emph{IEEE Trans Pattern Anal Mach Intell}}
  \bibinfo{volume}{31}, \bibinfo{number}{2} (\bibinfo{year}{2009}),
  \bibinfo{pages}{245--259}.
\newblock
\showISSN{0162-8828}
\href{https://doi.org/10.1109/TPAMI.2008.78}{doi:\nolinkurl{10.1109/TPAMI.2008.78}}


\bibitem[Mart{\i}nez-Munoz and Su{\'a}rez(2004)]%
        {martinez2004aggregation}
\bibfield{author}{\bibinfo{person}{Gonzalo Mart{\i}nez-Munoz} {and}
  \bibinfo{person}{Alberto Su{\'a}rez}.} \bibinfo{year}{2004}\natexlab{}.
\newblock \showarticletitle{Aggregation ordering in bagging}. In
  \bibinfo{booktitle}{\emph{AIA}}. Citeseer, \bibinfo{pages}{258--263}.
\newblock


\bibitem[Masegosa et~al\mbox{.}(2020)]%
        {masegosa2020second}
\bibfield{author}{\bibinfo{person}{Andr{\'e}s~R Masegosa},
  \bibinfo{person}{Stephan~Sloth Lorenzen}, \bibinfo{person}{Christian Igel},
  {and} \bibinfo{person}{Yevgeny Seldin}.} \bibinfo{year}{2020}\natexlab{}.
\newblock \showarticletitle{Second order PAC-Bayesian bounds for the weighted
  majority vote}. In \bibinfo{booktitle}{\emph{NeurIPS}},
  Vol.~\bibinfo{volume}{33}. \bibinfo{publisher}{Curran Associates, Inc.},
  \bibinfo{pages}{5263--5273}.
\newblock


\bibitem[Menon and Williamson(2018)]%
        {menon2018cost}
\bibfield{author}{\bibinfo{person}{Aditya~Krishna Menon} {and}
  \bibinfo{person}{Robert~C Williamson}.} \bibinfo{year}{2018}\natexlab{}.
\newblock \showarticletitle{The cost of fairness in binary classification}. In
  \bibinfo{booktitle}{\emph{FAT}}, Vol.~\bibinfo{volume}{81}.
  \bibinfo{publisher}{PMLR}, \bibinfo{pages}{107--118}.
\newblock


\bibitem[Nilforoshan et~al\mbox{.}(2022)]%
        {nilforoshan2022causal}
\bibfield{author}{\bibinfo{person}{Hamed Nilforoshan},
  \bibinfo{person}{Johann~D Gaebler}, \bibinfo{person}{Ravi Shroff}, {and}
  \bibinfo{person}{Sharad Goel}.} \bibinfo{year}{2022}\natexlab{}.
\newblock \showarticletitle{Causal conceptions of fairness and their
  consequences}. In \bibinfo{booktitle}{\emph{ICML}},
  Vol.~\bibinfo{volume}{162}. \bibinfo{publisher}{PMLR},
  \bibinfo{pages}{16848--16887}.
\newblock


\bibitem[Pearl et~al\mbox{.}(2009)]%
        {pearl2000models}
\bibfield{author}{\bibinfo{person}{Judea Pearl} {et~al\mbox{.}}}
  \bibinfo{year}{2009}\natexlab{}.
\newblock \bibinfo{booktitle}{\emph{Models, reasoning and inference (second
  edition)}}.
\newblock \bibinfo{publisher}{Cambridge University Press}. 204--205 pages.
\newblock


\bibitem[Pessach and Shmueli(2021)]%
        {pessach2021improving}
\bibfield{author}{\bibinfo{person}{Dana Pessach} {and} \bibinfo{person}{Erez
  Shmueli}.} \bibinfo{year}{2021}\natexlab{}.
\newblock \showarticletitle{Improving fairness of artificial intelligence
  algorithms in privileged-group selection bias data settings}.
\newblock \bibinfo{journal}{\emph{Expert Syst Appl}}  \bibinfo{volume}{185}
  (\bibinfo{year}{2021}), \bibinfo{pages}{115667}.
\newblock


\bibitem[Pleiss et~al\mbox{.}(2017)]%
        {pleiss2017fairness}
\bibfield{author}{\bibinfo{person}{Geoff Pleiss}, \bibinfo{person}{Manish
  Raghavan}, \bibinfo{person}{Felix Wu}, \bibinfo{person}{Jon Kleinberg}, {and}
  \bibinfo{person}{Kilian~Q Weinberger}.} \bibinfo{year}{2017}\natexlab{}.
\newblock \showarticletitle{On fairness and calibration}. In
  \bibinfo{booktitle}{\emph{NIPS}}, Vol.~\bibinfo{volume}{30}.
  \bibinfo{publisher}{Curran Associates, Inc.}
\newblock


\bibitem[Qian et~al\mbox{.}(2015)]%
        {qian2015pareto}
\bibfield{author}{\bibinfo{person}{Chao Qian}, \bibinfo{person}{Yang Yu}, {and}
  \bibinfo{person}{Zhi-Hua Zhou}.} \bibinfo{year}{2015}\natexlab{}.
\newblock \showarticletitle{Pareto ensemble pruning}. In
  \bibinfo{booktitle}{\emph{AAAI}}, Vol.~\bibinfo{volume}{29}.
  \bibinfo{pages}{2935--2941}.
\newblock


\bibitem[Quadrianto and Sharmanska(2017)]%
        {quadrianto2017recycling}
\bibfield{author}{\bibinfo{person}{Novi Quadrianto} {and}
  \bibinfo{person}{Viktoriia Sharmanska}.} \bibinfo{year}{2017}\natexlab{}.
\newblock \showarticletitle{Recycling privileged learning and distribution
  matching for fairness}. In \bibinfo{booktitle}{\emph{NIPS}},
  Vol.~\bibinfo{volume}{30}. \bibinfo{publisher}{Curran Associates, Inc.}
\newblock


\bibitem[Samadi et~al\mbox{.}(2018)]%
        {samadi2018price}
\bibfield{author}{\bibinfo{person}{Samira Samadi}, \bibinfo{person}{Uthaipon
  Tantipongpipat}, \bibinfo{person}{Jamie~H Morgenstern},
  \bibinfo{person}{Mohit Singh}, {and} \bibinfo{person}{Santosh Vempala}.}
  \bibinfo{year}{2018}\natexlab{}.
\newblock \showarticletitle{The price of fair pca: One extra dimension}. In
  \bibinfo{booktitle}{\emph{NeurIPS}}, Vol.~\bibinfo{volume}{31}.
  \bibinfo{publisher}{Curran Associates, Inc.}
\newblock


\bibitem[Schweighofer et~al\mbox{.}(2025)]%
        {schweighofer2025the}
\bibfield{author}{\bibinfo{person}{Kajetan Schweighofer},
  \bibinfo{person}{Adrian Arnaiz-Rodriguez}, \bibinfo{person}{Sepp Hochreiter},
  {and} \bibinfo{person}{Nuria~M Oliver}.} \bibinfo{year}{2025}\natexlab{}.
\newblock \showarticletitle{The disparate benefits of deep ensembles}. In
  \bibinfo{booktitle}{\emph{ICML}}.
\newblock
\urldef\tempurl%
\url{https://openreview.net/forum?id=tjPxZiqeHB}
\showURL{%
\tempurl}


\bibitem[Verma and Rubin(2018)]%
        {verma2018fairness}
\bibfield{author}{\bibinfo{person}{Sahil Verma} {and} \bibinfo{person}{Julia
  Rubin}.} \bibinfo{year}{2018}\natexlab{}.
\newblock \showarticletitle{Fairness definitions explained}. In
  \bibinfo{booktitle}{\emph{FairWare}}. \bibinfo{publisher}{ACM},
  \bibinfo{address}{New York, NY, USA}, \bibinfo{pages}{1--7}.
\newblock


\bibitem[Wick et~al\mbox{.}(2019)]%
        {wick2019unlocking}
\bibfield{author}{\bibinfo{person}{Michael Wick}, \bibinfo{person}{Swetasudha
  Panda}, {and} \bibinfo{person}{Jean-Baptiste Tristan}.}
  \bibinfo{year}{2019}\natexlab{}.
\newblock \showarticletitle{Unlocking fairness: a trade-off revisited}. In
  \bibinfo{booktitle}{\emph{NeurIPS}}, Vol.~\bibinfo{volume}{32}.
  \bibinfo{publisher}{Curran Associates, Inc.}, \bibinfo{pages}{8783--8792}.
\newblock


\bibitem[Wightman(1998)]%
        {wightman1998lsac}
\bibfield{author}{\bibinfo{person}{Linda~F Wightman}.}
  \bibinfo{year}{1998}\natexlab{}.
\newblock \showarticletitle{LSAC National Longitudinal Bar Passage Study. LSAC
  Research Report Series.}
\newblock  (\bibinfo{year}{1998}).
\newblock


\bibitem[Woodworth et~al\mbox{.}(2017)]%
        {woodworth2017learning}
\bibfield{author}{\bibinfo{person}{Blake Woodworth}, \bibinfo{person}{Suriya
  Gunasekar}, \bibinfo{person}{Mesrob~I Ohannessian}, {and}
  \bibinfo{person}{Nathan Srebro}.} \bibinfo{year}{2017}\natexlab{}.
\newblock \showarticletitle{Learning non-discriminatory predictors}. In
  \bibinfo{booktitle}{\emph{COLT}}, Vol.~\bibinfo{volume}{65}.
  \bibinfo{publisher}{PMLR}, \bibinfo{pages}{1920--1953}.
\newblock


\bibitem[Xia et~al\mbox{.}(2018)]%
        {xia2018maximum}
\bibfield{author}{\bibinfo{person}{Xin Xia}, \bibinfo{person}{Tao Lin}, {and}
  \bibinfo{person}{Zhi Chen}.} \bibinfo{year}{2018}\natexlab{}.
\newblock \showarticletitle{Maximum relevancy maximum complementary based
  ordered aggregation for ensemble pruning}.
\newblock \bibinfo{journal}{\emph{Appl Intell}} \bibinfo{volume}{48},
  \bibinfo{number}{9} (\bibinfo{year}{2018}), \bibinfo{pages}{2568--2579}.
\newblock


\bibitem[Zadeh et~al\mbox{.}(2017)]%
        {zadeh2017scalable}
\bibfield{author}{\bibinfo{person}{Sepehr~Abbasi Zadeh},
  \bibinfo{person}{Mehrdad Ghadiri}, \bibinfo{person}{Vahab Mirrokni}, {and}
  \bibinfo{person}{Morteza Zadimoghaddam}.} \bibinfo{year}{2017}\natexlab{}.
\newblock \showarticletitle{Scalable feature selection via distributed
  diversity maximization}. In \bibinfo{booktitle}{\emph{AAAI}},
  Vol.~\bibinfo{volume}{31}. \bibinfo{pages}{2876--2883}.
\newblock


\bibitem[Zafar et~al\mbox{.}(2017a)]%
        {zafar2017fairness2}
\bibfield{author}{\bibinfo{person}{Muhammad~Bilal Zafar},
  \bibinfo{person}{Isabel Valera}, \bibinfo{person}{Manuel Gomez~Rodriguez},
  {and} \bibinfo{person}{Krishna~P Gummadi}.} \bibinfo{year}{2017}\natexlab{a}.
\newblock \showarticletitle{Fairness beyond disparate treatment \& disparate
  impact: Learning classification without disparate mistreatment}. In
  \bibinfo{booktitle}{\emph{WWW}}. \bibinfo{publisher}{International World Wide
  Web Conferences Steering Committee}, \bibinfo{address}{Republic and Canton of
  Geneva, CHE}, \bibinfo{pages}{1171--1180}.
\newblock


\bibitem[Zafar et~al\mbox{.}(2017b)]%
        {zafar2017fairness1}
\bibfield{author}{\bibinfo{person}{Muhammad~Bilal Zafar},
  \bibinfo{person}{Isabel Valera}, \bibinfo{person}{Manuel~Gomez Rogriguez},
  {and} \bibinfo{person}{Krishna~P Gummadi}.} \bibinfo{year}{2017}\natexlab{b}.
\newblock \showarticletitle{Fairness constraints: Mechanisms for fair
  classification}. In \bibinfo{booktitle}{\emph{AISTATS}},
  Vol.~\bibinfo{volume}{54}. \bibinfo{publisher}{PMLR},
  \bibinfo{pages}{962--970}.
\newblock


\bibitem[Zemel et~al\mbox{.}(2013)]%
        {zemel2013learning}
\bibfield{author}{\bibinfo{person}{Rich Zemel}, \bibinfo{person}{Yu Wu},
  \bibinfo{person}{Kevin Swersky}, \bibinfo{person}{Toni Pitassi}, {and}
  \bibinfo{person}{Cynthia Dwork}.} \bibinfo{year}{2013}\natexlab{}.
\newblock \showarticletitle{Learning fair representations}. In
  \bibinfo{booktitle}{\emph{ICML}}. \bibinfo{publisher}{PMLR},
  \bibinfo{address}{Atlanta, Georgia, USA}, \bibinfo{pages}{325--333}.
\newblock


\bibitem[Zhang et~al\mbox{.}(2019)]%
        {zhang2019two}
\bibfield{author}{\bibinfo{person}{Hua Zhang}, \bibinfo{person}{Yujie Song},
  \bibinfo{person}{Bo Jiang}, \bibinfo{person}{Bi Chen}, {and}
  \bibinfo{person}{Guogen Shan}.} \bibinfo{year}{2019}\natexlab{}.
\newblock \showarticletitle{Two-stage bagging pruning for reducing the ensemble
  size and improving the classification performance}.
\newblock \bibinfo{journal}{\emph{Math Probl Eng}}  \bibinfo{volume}{2019}
  (\bibinfo{year}{2019}).
\newblock


\bibitem[Zhang et~al\mbox{.}(2021)]%
        {zhang2021farf}
\bibfield{author}{\bibinfo{person}{Wenbin Zhang}, \bibinfo{person}{Albert
  Bifet}, \bibinfo{person}{Xiangliang Zhang}, \bibinfo{person}{Jeremy~C Weiss},
  {and} \bibinfo{person}{Wolfgang Nejdl}.} \bibinfo{year}{2021}\natexlab{}.
\newblock \showarticletitle{Farf: A fair and adaptive random forests
  classifier}. In \bibinfo{booktitle}{\emph{PAKDD}}.
  \bibinfo{publisher}{Springer International Publishing},
  \bibinfo{address}{Cham}, \bibinfo{pages}{245--256}.
\newblock


\bibitem[Zhou(2021)]%
        {zhou2021machine}
\bibfield{author}{\bibinfo{person}{Zhi-Hua Zhou}.}
  \bibinfo{year}{2021}\natexlab{}.
\newblock \bibinfo{booktitle}{\emph{Machine learning}}.
\newblock \bibinfo{publisher}{Springer Nature}.
\newblock


\end{thebibliography}

\appendix

\clearpage
\section{Supplementary Methodology}

\subsection{Ensemble pruning methods to raise accuracy and fairness concurrently}
\label{appx:prop}

Based on the domination concept and this objective $\justObjT(f,f')$, we propose an ensemble pruning method, named as ``\emph{\poepfull{} (}\poepabbr{}\emph{)},'' presented in Algorithm~\ref{alg:pareto}, aiming to construct stronger ensemble classifiers with less accuracy damage. 
Note that a selector vector $\mathbf{r}\in\{0,1\}^m$ in line 1 is used to indicate potential sub-ensembles, wherein $|\mathbf{r}|=k$ and the picked sub-ensemble can be denoted by $F_\mathbf{r} \defineq \mathbf{r}$ as well. 
The notations $\mathcal{N}_-(\mathbf{r}')$ and $\mathcal{N}_+(\mathbf{r}')$ in line~9 signify the sets of neighbour solutions of $\mathbf{r}'$, wherein each element in $\mathcal{N}_-(\mathbf{r}')$ has one less individual member than $\mathbf{r}'$ and each element in $\mathcal{N}_+(\mathbf{r}')$ has one more individual member than $\mathbf{r}'$, respectively. 
The used bi-objective here is denoted by $\mathcal{G}(\mathbf{r})= (\justAccF(\mathbf{r},S),\justLf(\mathbf{r},S))$, wherein $\mathbf{r}$ indicates the individual members that are selected to join the pruned sub-ensemble. 
Also note that a bi-objective optimisation problem may have multiple solutions that are Pareto optimal, rather than one single Pareto optimal solution only. A solution $\mvrho(\cdot)$ is Pareto optimal if there is no other solution in $\mathcal{F}$ that dominates $\mvrho(\cdot)$. 
It is worth mentioning that \poepabbr{} is not the only way to achieve a strong sub-ensemble, and two other easily-implemented pruning methods are also able to achieve this goal as well as accelerate the execution time (presented in Fig.~\ref{fig:sota,us}\subref{subfig:us,2} and Appendix~\ref{appx:alter}).

\begin{algorithm}[b]
\caption{\poepfull{} (\poepabbr)}
\label{alg:pareto}
\begin{algorithmic}[1]
   \REQUIRE training set $S\!=\{(\bm{x}_i,y_i)\}_{i=1}^n$, original ensemble $F\!=\{f_j(\cdot)\}_{j=1}^m$ via weighted vote, and threshold $k$ as maximum size of the sub-ensemble after pruning 
   \ENSURE Pruned sub-ensemble $H$ ($H\subset F$ and $|H|\leqslant k$)
   \STATE Randomly pick $k$ individual members from $F$, indicated by $\mathbf{r}$\label{alg:line,2} 
   \STATE Initialise a candidate set for pruned sub-ensembles $\mathcal{P}\!=\! \{\mathbf{r}\}$\label{alg:line,4}
   \FOR{$i=1$ to $k$}
      \STATE Randomly choose $\mathbf{r}$ from $\mathcal{P}$ with equal probability \label{alg:line,6}
      \STATE Generate $\mathbf{r}'$ by flipping each bit of $\mathbf{r}$ with probability $\sfrac{1}{m}$\label{alg:line,7}
      \IF{$\exists \,\mathbf{z}\in \mathcal{P}$ such that $\mathbf{z} \succ_\mathcal{G} \mathbf{r}'$}
         \STATE \textbf{continue} \label{alg:line,9}
      \ENDIF
      \STATE $\mathcal{P}= (\mathcal{P}\setminus\{ \mathbf{z}\in \mathcal{P}\mid \mathbf{r}'\succeq_\mathcal{G}\mathbf{z}\})\bigcup\{\mathbf{r}'\}$ \label{alg:line,10}
      \STATE Let $\mathcal{V}= \mathcal{N}_-(\mathbf{r}') \bigcup \mathcal{N}_+(\mathbf{r}')$ \label{alg:line,11}
      \STATE Sort $\mathcal{V}$ by $\argmin_{\mathbf{v}\in\mathcal{V}} \justObjF(\mathbf{v},S)$ in ascending order \label{alg:line,12}
      \FOR{$\mathbf{v}\in\mathcal{V}$}
        \IF{$\exists \,\mathbf{z}\in \mathcal{P}$ such that $\mathbf{z}\succ_{\mathcal{G}} \mathbf{v}$}
           \STATE \textbf{continue} \label{alg:line,15}
        \ENDIF
        \STATE $\mathcal{P}=(\mathcal{P}\setminus\{ \mathbf{z}\in \mathcal{P}\mid \mathbf{v}\succeq_\mathcal{G} \mathbf{z} \})\bigcup\{\mathbf{v}\}$ \label{alg:line,16}
      \ENDFOR
   \ENDFOR
   \STATE $H= \argmin_{\mathbf{r}\in \mathcal{P}} \justObjF(\mathbf{r},S)$ \label{alg:line,17}
\end{algorithmic}
\end{algorithm}

\begin{algorithm}[t]
\caption{Centralised version of \propfull{} (\greedy{})}
\label{alg:centralised}
\begin{algorithmic}[1]
   \REQUIRE training set $S=\{(\bm{x}_i,y_i)\}_{i=1}^n$, original ensemble $F=\{f_j(\cdot)\}_{j=1}^m$, and threshold $k$ as maximum size after pruning 
   \ENSURE Pruned sub-ensemble $H$ ($H\subset F$ and $|H|\leqslant k$) 
   \STATE $H\gets$ an arbitrary individual member $f_i\in F$
   \FOR{$i=2$ {\bf to} $k$}
      \STATE $f^*\gets \argmin_{f_i\in F\setminus H} \sum_{f_j\in H} \justObjF(f_i,f_j,S)$
      \STATE Move $f^*$ from $F$ to $H$
   \ENDFOR
\end{algorithmic}
\end{algorithm}
\begin{algorithm}[t]
\caption{Distributed version of \propfull{} (\ddismi{})}
\label{alg:distributed}
\begin{algorithmic}[1]
   \REQUIRE training set $S=\{(\bm{x}_i,y_i)\}_{i=1}^n$, original ensemble $F=\{f_j(\cdot)\}_{j=1}^m$, threshold $k$ as maximum size after pruning, and number of machines $n_m$ 
   \ENSURE Pruned sub-ensemble $H$ ($H\subset F$ and $|H|\leqslant k$) 
   \STATE Partition $F$ randomly into $n_m$ groups as equally as possible, \ie{} $F_1,...,F_{n_m}$
   \FOR{$i=1$ {\bf to} $n_m$}
      \STATE $H_i \gets$ \propabbr\emph{-C}($F_i,k$)
   \ENDFOR
   \STATE $H'\gets$ \propabbr\emph{-C}($\bigcup_{i=1}^{n_m}H_i ,k$)
   \STATE $H\gets \argmin_{T\in\{ H_1,...,H_{n_m}, H' \}} \justObjF(T,S)$
\end{algorithmic}
\end{algorithm}

\subsection{Two extra easily-implemented pruning methods}
\label{appx:alter}
There are many methods to solve the aforementioned bi-objective optimisation problem, and \poepabbr{} using the concept of Pareto optimal is just one of them. 
Besides, \poepabbr{} usually comes with a drawback of long time cost. 
Therefore, inspired by the work \citep{zadeh2017scalable,bian2019ensemble}, 
we also come up with two extra pruning methods that could be easily implemented, named as ``\emph{\propfull{} (}\propabbr{}\emph{)},'' presented in Algorithms~\ref{alg:centralised} and \ref{alg:distributed}. 
They utilise the balanced objective function in \eqref{eq:10} as well yet with one key difference from \poepabbr{}, that is, they do not have the concepts of domination nor Pareto optimality. 
The objective used in \propabbr{} adopts the weighted sum method in bi-objective optimisation problems, which is simple and easy to implement. 
However, it also comes along with the disadvantage of the difficulty of setting an appropriate weight for each factor. 
Besides, it cannot find certain Pareto-optimal solutions in
the case of a non-convex objective space as well. 
\propabbr{} has two versions of it, that is, a centralised version (\greedy{}) and a distributed one (\ddismi{}). 
The key difference between them is that a sub-procedure (line 3) in \ddismi{} could be parallelized and therefore greatly contribute to a speedup for the whole pruning procedure, which is also verified in Fig.~\ref{appx:fig:parl} of Section~\ref{appx:expt,ut}, 
providing a comparison between them over time cost. 
Note that the sub-procedure could be replaced with any other existing pruning method as well.

\section{Experimental Setups}
\label{subsec:setup}
In this section, we present experimental setups including datasets, evaluation metrics, baseline fairness measures, baseline fairness-aware ensemble-based methods, baseline ensemble pruning methods, and implementation details.

\begin{table}[t]
\centering\caption{Dataset statistics. 
}\label{tab:dataset}
\vspace{-3.5mm}%
\scalebox{.731}{%
\begin{threeparttable}[b]
\begin{tabular}{c|r|r|r|r| rr|r}
\toprule
\multirow{2}{*}{\bf Dataset}& \multirow{2}{*}{\bf \#inst} & \multicolumn{2}{c|}{\bf \#feature} & 
\multirow{2}{*}{\tabincell{l}{\bf \#sen-att}} & \multicolumn{3}{c}{\bf \#privileged group for sen-att}\\
\cline{3-4} \cline{6-8}
& & raw & binarized & & 1st sen-att & 2nd sen-att & joint \\
\midrule
ricci  &   118 &  5 &  6  & 1 &    68 in race & --- & --- \\
credit &  1000 & 21 & 59  & 2 &   690 in  sex &   851 in age & 625 \\
income & 30162 & 14 & 99  & 2 & 25933 in race & 20380 in sex & 18038 \\
ppr    &  6167 & 11 & 402 & 2 &  4994 in  sex & 2100 in race & 1620 \\
ppvr   &  4010 & 11 & 328 & 2 &  3173 in  sex & 1452 in race & 1119 \\
\bottomrule
\end{tabular}
\begin{tablenotes}
\item[1] Note that the columns named `\#inst' and `\#sen-att' represent the number of instances and the number of sensitive attributes in the dataset, respectively. 
\item[2] The joint sensitive attribute of one instance represents that both two of the sensitive attributes of this instance belong to the corresponding privileged group. 
\end{tablenotes}
\end{threeparttable}
}%
\end{table}

\begin{figure*}[t]
\begin{minipage}{\textwidth}
\centering%
\subfloat[]{
\includegraphics[height=21.5mm]{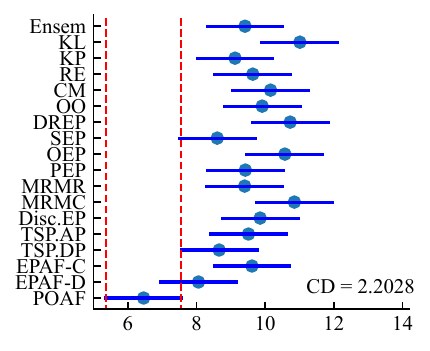}}
\subfloat[]{
\includegraphics[height=21.5mm]{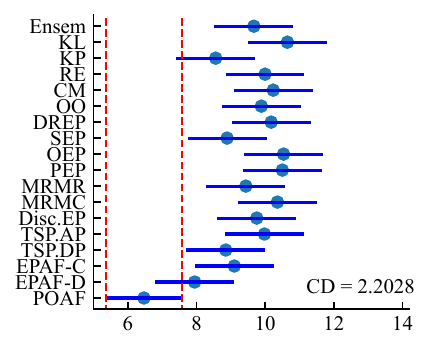}}
\subfloat[]{
\includegraphics[height=21.5mm]{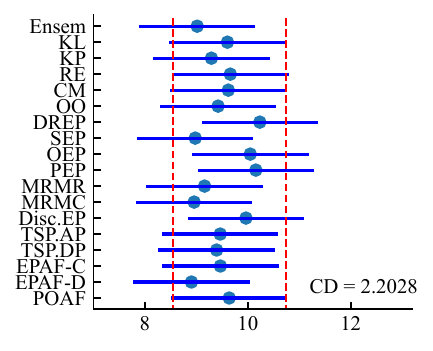}}
\subfloat[]{
\includegraphics[height=21.5mm]{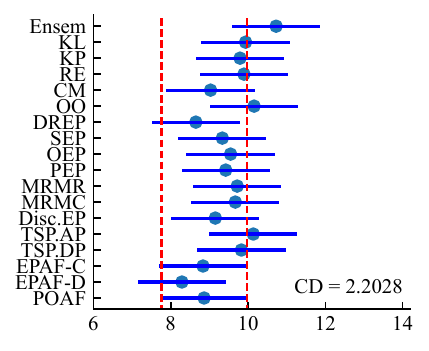}}
\\ \vspace{-1em}%
\subfloat[]{
\includegraphics[height=22.5mm]{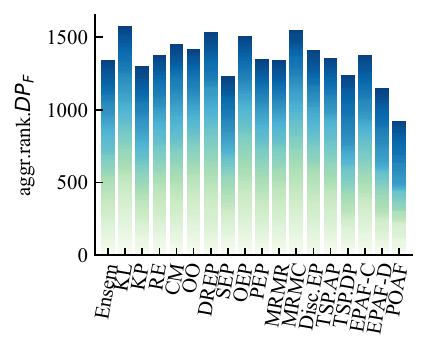}}
\subfloat[]{
\includegraphics[height=22.5mm]{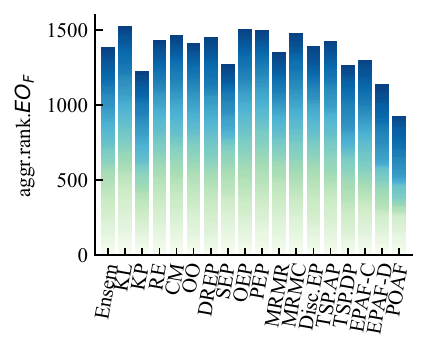}}
\subfloat[]{
\includegraphics[height=22.5mm]{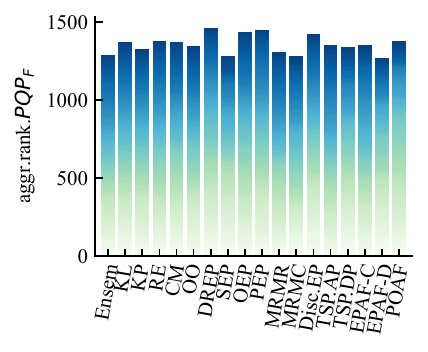}}
\subfloat[]{
\includegraphics[height=22.5mm]{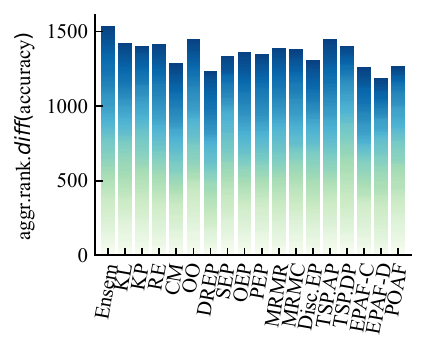}}
\vspace{-4mm}\caption{%
Comparison of the pruning methods with \poepabbr{}, using AdaBoostM1 to conduct homogeneous ensembles. (a--d) Friedman test chart on three group fairness measures (\ie{} \gls{dp}, \gls{eopp}, and \gls{pqp}) and the discrepancy of test accuracy between the privileged group and marginalised groups, rejecting the null hypothesis at 5\% significance level. (e--h) The aggregated rank. 
}\label{fig:sota,fair,am1}
\end{minipage}
\begin{minipage}{\textwidth}
\vspace{-2mm}\centering%
\subfloat[]{
\includegraphics[height=21.5mm]{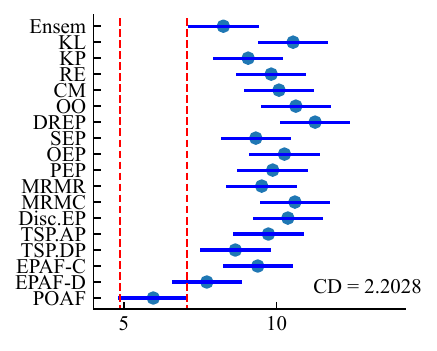}}
\subfloat[]{
\includegraphics[height=21.5mm]{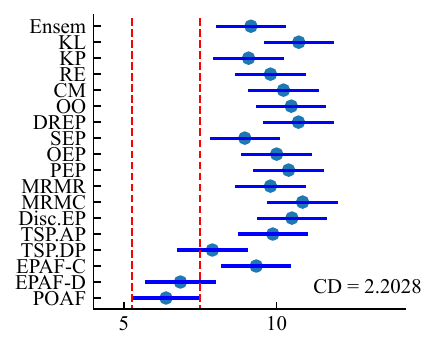}}
\subfloat[]{
\includegraphics[height=21.5mm]{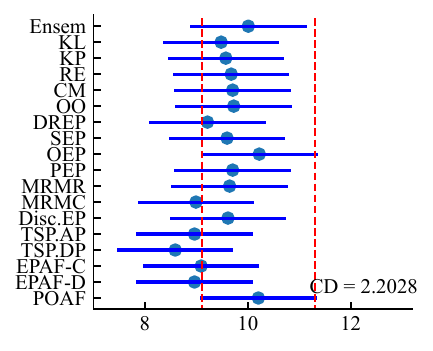}}
\subfloat[]{
\includegraphics[height=21.5mm]{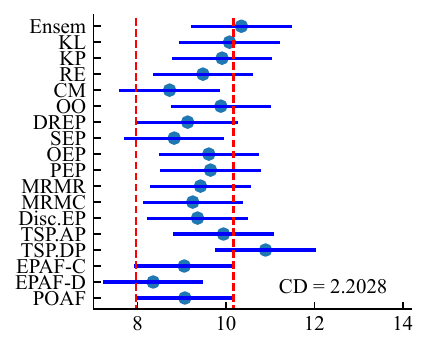}}
\\ \vspace{-1em}%
\subfloat[]{
\includegraphics[height=22.5mm]{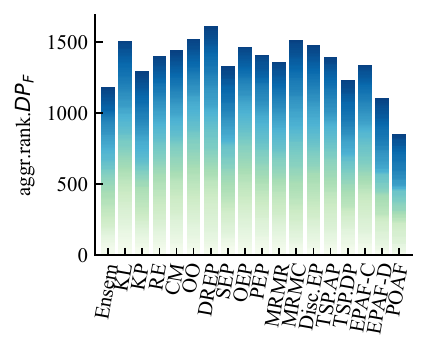}}
\subfloat[]{
\includegraphics[height=22.5mm]{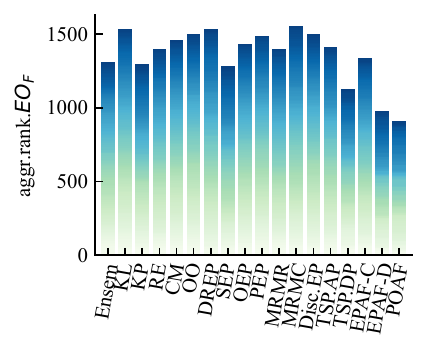}}
\subfloat[]{
\includegraphics[height=22.5mm]{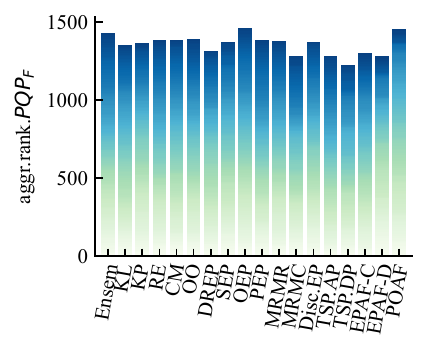}}
\subfloat[]{
\includegraphics[height=22.5mm]{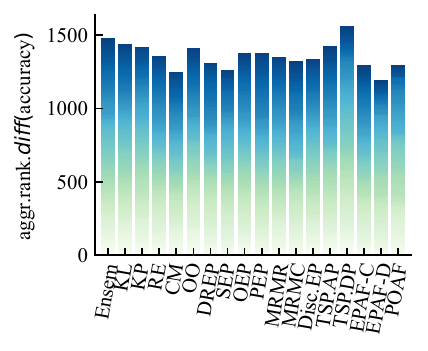}}
\vspace{-4mm}\caption{%
Comparison of the pruning methods with \poepabbr{}, using SAMME to conduct homogeneous ensembles. 
(a--d) Friedman test chart on three group fairness measures (\ie{} \gls{dp}, \gls{eopp}, and \gls{pqp}) and the discrepancy of test accuracy between the privileged group and marginalised groups, respectively, rejecting the null hypothesis at 5\% significance level. 
(e--h) The aggregated rank. 
}\label{fig:sota,fair,sam}%
\end{minipage}
\end{figure*}
\begin{figure*}[t]
\begin{minipage}{\textwidth}
\centering%
\subfloat[]{
\includegraphics[height=22mm]{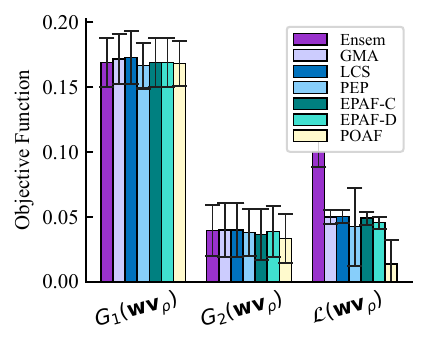}}
\subfloat[]{
\includegraphics[height=22mm]{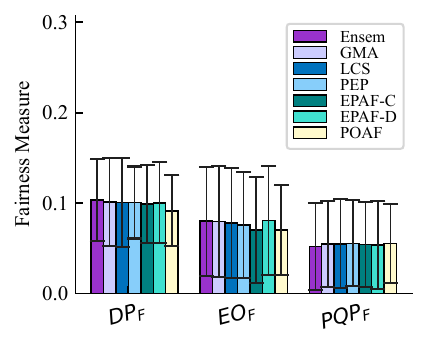}}
\subfloat[]{
\includegraphics[height=22mm]{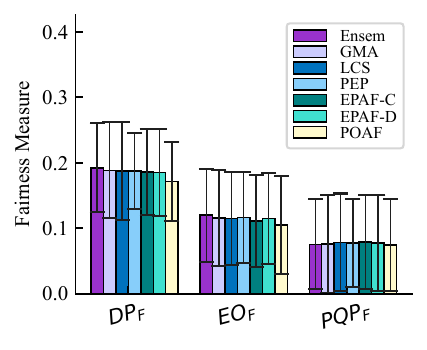}}
\subfloat[]{
\includegraphics[height=22mm]{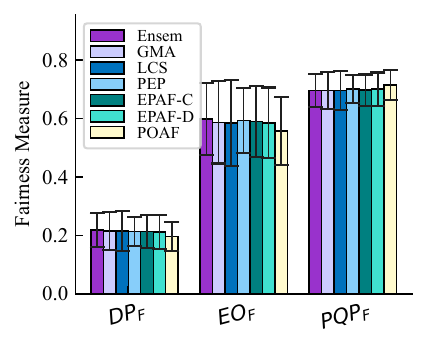}}
\\ \vspace{-1em}
\subfloat[]{
\includegraphics[height=22mm]{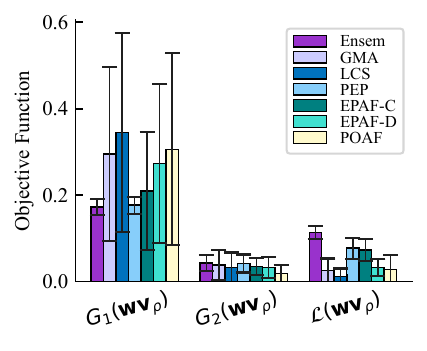}}
\subfloat[]{
\includegraphics[height=22mm]{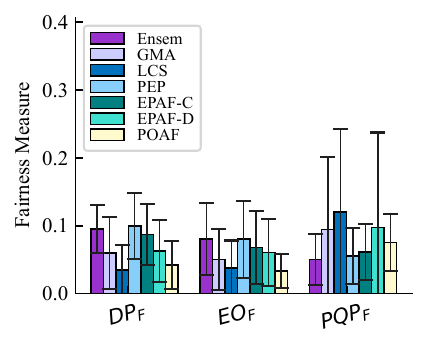}}
\subfloat[]{
\includegraphics[height=22mm]{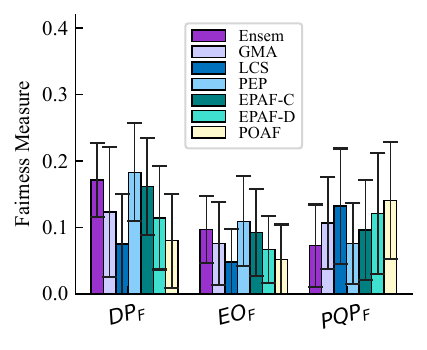}}
\subfloat[]{
\includegraphics[height=22mm]{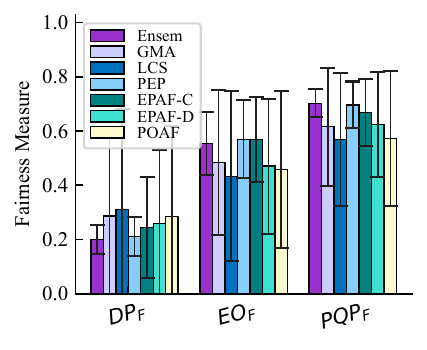}}
\\ \vspace{-1em}
\subfloat[]{
\includegraphics[height=22mm]{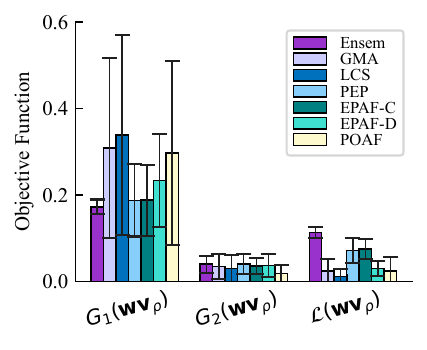}}
\subfloat[]{
\includegraphics[height=22mm]{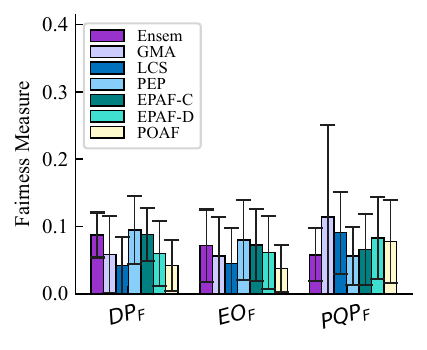}}
\subfloat[]{
\includegraphics[height=22mm]{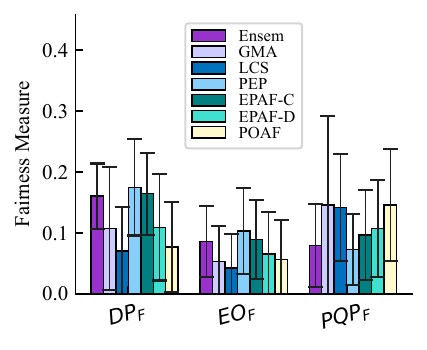}}
\subfloat[]{
\includegraphics[height=22mm]{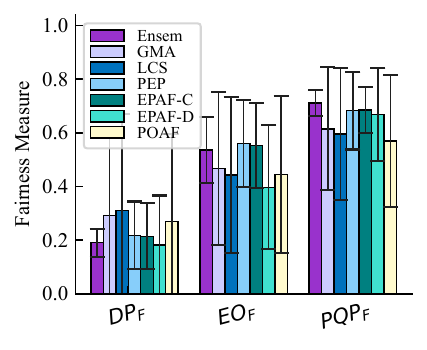}}
\vspace{-4mm}\caption{%
Comparison of objective functions (\ie{} $\justAccT(\mvrho)$, $\justLt(\mvrho)$, and $\justObjT(\mvrho)$) and three group fairness measures (\ie{} \gls{dp}, \gls{eopp}, and \gls{pqp}) on the Income dataset. 
(a--d), (e--h), and (i--l) Using bagging, AdaBoostM1, and SAMME to conduct homogeneous ensemble classifiers, respectively. 
Note that (b), (f), and (j) indicate the fairness discrepancy on the Race attribute, (c), (g), and (k) indicate that on the Sex attribute, and (d), (h), and (l) indicate that on the joint sensitive attribute, respectively, wherein the closer to zero, the better. 
Note that the smaller the better in (a), (e), and (i). 
}\label{fig:sens,adult}
\end{minipage}
\end{figure*}

\begin{table*}[t]
\centering\caption{%
Comparison of fairness-aware ensemble methods with \poepabbr{}. (a--b) Comparison on the test accuracy (\%) and $\text{f}_1$ score. 
}\label{tab:aware,acc;part1}
\vspace{-1.2em}%
\subfloat[]{\label{tab:aware,acc,a}
\scalebox{.54}{%
\begin{threeparttable}[b]
\begin{tabular}{rr rrrrrrr}
\toprule
Dataset & $\text{Att}_\text{sen}$ & \multicolumn{1}{c}{lightGBM} & \multicolumn{1}{c}{FairGBM} & \multicolumn{1}{c}{AdaFair} & \multicolumn{1}{c}{Bagging} & \multicolumn{1}{c}{\emph{EPAF-C}} & \multicolumn{1}{c}{\emph{EPAF-D}} & \multicolumn{1}{c}{\emph{POAF}} \\
\midrule
ricci &race & 98.26\topelement{$\pm$2.13} & 98.26\topelement{$\pm$2.13} & 98.26\topelement{$\pm$2.13} & 98.26\topelement{$\pm$2.13} & 98.26\topelement{$\pm$2.13} & 98.26\topelement{$\pm$2.13} & \textbf{99.13\topelement{$\pm$1.74}} \\
credit&sex  & \textbf{75.00\topelement{$\pm$3.21}} & \textbf{75.00\topelement{$\pm$3.21}} & 70.00\topelement{$\pm$0.00}\dag & 74.00\topelement{$\pm$2.47} & 71.60\topelement{$\pm$2.96} & 71.00\topelement{$\pm$1.76} & 71.80\topelement{$\pm$4.14} \\
      &age  & \textbf{75.00\topelement{$\pm$3.21}} & 74.80\topelement{$\pm$3.01} & 70.00\topelement{$\pm$0.00}\dag & 74.00\topelement{$\pm$2.47} & 71.60\topelement{$\pm$2.96} & 71.00\topelement{$\pm$1.76} & 71.80\topelement{$\pm$4.14} \\
income&race & \textbf{86.10\topelement{$\pm$0.42}}\dag & 86.02\topelement{$\pm$0.45}\dag & 24.89\topelement{$\pm$0.00}\ddag & 84.15\topelement{$\pm$0.44} & 83.53\topelement{$\pm$0.53} & 83.55\topelement{$\pm$0.42} & 83.80\topelement{$\pm$0.68} \\
      &sex  & \textbf{86.10\topelement{$\pm$0.42}}\dag & 86.08\topelement{$\pm$0.51}\dag & 24.89\topelement{$\pm$0.00}\ddag & 84.15\topelement{$\pm$0.44} & 83.53\topelement{$\pm$0.53} & 83.55\topelement{$\pm$0.42} & 83.80\topelement{$\pm$0.68} \\
ppr   &sex  & \textbf{67.97\topelement{$\pm$0.62}}\dag & 67.50\topelement{$\pm$0.86}\dag & 45.54\topelement{$\pm$0.00}\ddag & 65.41\topelement{$\pm$0.54} & 64.45\topelement{$\pm$1.20} & 64.56\topelement{$\pm$0.78} & 62.73\topelement{$\pm$0.69} \\
      &race & \textbf{67.97\topelement{$\pm$0.62}}\dag & 67.48\topelement{$\pm$0.72}\dag & 45.54\topelement{$\pm$0.00}\ddag & 65.41\topelement{$\pm$0.54} & 64.45\topelement{$\pm$1.20} & 64.56\topelement{$\pm$0.78} & 62.73\topelement{$\pm$0.69} \\
ppvr  &sex  & \textbf{84.24\topelement{$\pm$0.76}}\dag & 84.17\topelement{$\pm$0.90}\dag & 49.39\topelement{$\pm$19.77}\ddag & 81.67\topelement{$\pm$1.64} & 81.10\topelement{$\pm$1.28}\ddag & 81.30\topelement{$\pm$1.36} & 82.60\topelement{$\pm$0.74} \\
      &race & \textbf{84.24\topelement{$\pm$0.76}}\dag & 84.24\topelement{$\pm$0.76}\dag & 69.01\topelement{$\pm$26.42} & 81.67\topelement{$\pm$1.64} & 81.10\topelement{$\pm$1.28}\ddag & 81.30\topelement{$\pm$1.36} & 82.60\topelement{$\pm$0.74} \\
\midrule
\multicolumn{2}{c}{W/T/L}    & 0/3/6 & 0/3/6 & 5/2/2 & 0/7/2 & 2/5/2 & 0/7/2 & --- \\
\multicolumn{2}{c}{avg.rank} & \textbf{1.50} & 2.17 & 6.72 & 3.39 & 5.39 & 4.94 & 3.89 \\
\bottomrule
\end{tabular}
\begin{tablenotes}%
\item[1] The reported results are the average values of each method and the corresponding standard deviation under 5-fold cross-validation on each dataset. 
\item[2] By two-tailed paired $t$-test at 5\% significance level, $\ddagger$ and $\dagger$ denote that the performance of \poepabbr{} is superior to and inferior to that of the comparative baseline method,  respectively. 
\item[3] The last two rows show the results of $t$-test and average rank, respectively. 
The `W/T/L' in $t$-test indicates the numbers that \poepabbr{} is superior to, not significantly different from, or inferior to the corresponding comparative pruning methods. The average rank is calculated according to the Friedman test~\cite{demvsar2006statistical}. 
\end{tablenotes}
\end{threeparttable}
}}
\\
\vspace{-1em}
\subfloat[]{\label{tab:aware,acc,b}
\scalebox{.54}{%
\begin{tabular}{rr rrrrrrr}
\toprule
Dataset & $\text{Att}_\text{sen}$ & \multicolumn{1}{c}{lightGBM} & \multicolumn{1}{c}{FairGBM} & \multicolumn{1}{c}{AdaFair} & \multicolumn{1}{c}{Bagging} & \multicolumn{1}{c}{\emph{EPAF-C}} & \multicolumn{1}{c}{\emph{EPAF-D}} & \multicolumn{1}{c}{\emph{POAF}} \\
\midrule
ricci&race & 0.9818\topelement{$\pm$0.0224} & 0.9818\topelement{$\pm$0.0224} & 0.9818\topelement{$\pm$0.0224} & 0.9818\topelement{$\pm$0.0224} & 0.9818\topelement{$\pm$0.0224} & 0.9818\topelement{$\pm$0.0224} & \textbf{0.9913\topelement{$\pm$0.0174}} \\
credit&sex & \textbf{0.8347\topelement{$\pm$0.0190}} & \textbf{0.8347\topelement{$\pm$0.0190}} & 0.8235\topelement{$\pm$0.0000}\dag & 0.8232\topelement{$\pm$0.0164} & 0.8048\topelement{$\pm$0.0189} & 0.8017\topelement{$\pm$0.0101} & 0.8093\topelement{$\pm$0.0311} \\
&age & \textbf{0.8347\topelement{$\pm$0.0190}} & 0.8326\topelement{$\pm$0.0178} & 0.8235\topelement{$\pm$0.0000}\dag & 0.8232\topelement{$\pm$0.0164} & 0.8048\topelement{$\pm$0.0189} & 0.8017\topelement{$\pm$0.0101} & 0.8093\topelement{$\pm$0.0311} \\
income&race & \textbf{0.6766\topelement{$\pm$0.0095}}\dag & 0.6740\topelement{$\pm$0.0102}\dag & 0.3986\topelement{$\pm$0.0000}\ddag & 0.6615\topelement{$\pm$0.0073} & 0.6530\topelement{$\pm$0.0084} & 0.6523\topelement{$\pm$0.0052} & 0.6412\topelement{$\pm$0.0267} \\
&sex & \textbf{0.6766\topelement{$\pm$0.0095}}\dag & 0.6742\topelement{$\pm$0.0114}\dag & 0.3986\topelement{$\pm$0.0000}\ddag & 0.6615\topelement{$\pm$0.0073} & 0.6530\topelement{$\pm$0.0084} & 0.6523\topelement{$\pm$0.0052} & 0.6412\topelement{$\pm$0.0267} \\
ppr&sex & 0.6202\topelement{$\pm$0.0117}\dag & 0.6118\topelement{$\pm$0.0144}\dag & \textbf{0.6258\topelement{$\pm$0.0000}}\dag & 0.6066\topelement{$\pm$0.0112}\dag & 0.5970\topelement{$\pm$0.0197}\dag & 0.5968\topelement{$\pm$0.0144}\dag & 0.5346\topelement{$\pm$0.0341} \\
&race & 0.6202\topelement{$\pm$0.0117}\dag & 0.6177\topelement{$\pm$0.0185}\dag & \textbf{0.6258\topelement{$\pm$0.0000}}\dag & 0.6066\topelement{$\pm$0.0112}\dag & 0.5970\topelement{$\pm$0.0197}\dag & 0.5968\topelement{$\pm$0.0144}\dag & 0.5346\topelement{$\pm$0.0341} \\
ppvr&sex & 0.2388\topelement{$\pm$0.0061}\dag & 0.2380\topelement{$\pm$0.0068}\dag & \textbf{0.3341\topelement{$\pm$0.0308}}\dag & 0.3007\topelement{$\pm$0.0283}\dag & 0.2894\topelement{$\pm$0.0325}\dag & 0.2969\topelement{$\pm$0.0383}\dag & 0.2356\topelement{$\pm$0.0325} \\
&race & 0.2388\topelement{$\pm$0.0061}\dag & 0.2388\topelement{$\pm$0.0061}\dag & \textbf{0.3374\topelement{$\pm$0.0570}}\dag & 0.3007\topelement{$\pm$0.0283}\dag & 0.2894\topelement{$\pm$0.0325}\dag & 0.2969\topelement{$\pm$0.0383}\dag & 0.2356\topelement{$\pm$0.0325} \\
\midrule
\multicolumn{2}{c}{W/T/L} & 0/3/6 & 0/3/6 & 2/1/6 & 0/5/4 & 0/5/4 & 0/5/4 & --- \\
\multicolumn{2}{c}{avg.rank} & \textbf{2.61} & 3.28 & 3.17 & 3.39 & 4.72 & 5.17 & 5.67 \\
\bottomrule
\end{tabular}
}}
\vspace{-5mm}
\end{table*}
\begin{table*}[t]
\centering\caption{%
Comparison of the fairness-aware ensemble methods with \poepabbr{}. (a) Comparison over the proposed \gls{dr} measure; (b--d) Comparison over three group fairness measures (that is, \gls{dp}, \gls{eopp}, and \gls{pqp}), respectively.  
}\label{tab:aware,acc;part2}
\vspace{-1.2em}%
\subfloat[]{\label{tab:aware,acc,c}
\scalebox{.54}{
\begin{tabular}{rr rrrrrrr}
\toprule
Dataset & $\text{Att}_\text{sen}$ & \multicolumn{1}{c}{lightGBM} & \multicolumn{1}{c}{FairGBM} & \multicolumn{1}{c}{AdaFair} & \multicolumn{1}{c}{Bagging} & \multicolumn{1}{c}{\emph{EPAF-C}} & \multicolumn{1}{c}{\emph{EPAF-D}} & \multicolumn{1}{c}{\emph{POAF}} \\
\midrule
ricci & race
 & \textbf{0.0000\topelement{$\pm$0.0000}} & \textbf{0.0000\topelement{$\pm$0.0000}} & \textbf{0.0000\topelement{$\pm$0.0000}} & \textbf{0.0000\topelement{$\pm$0.0000}} & \textbf{0.0000\topelement{$\pm$0.0000}} & \textbf{0.0000\topelement{$\pm$0.0000}} & \textbf{0.0000\topelement{$\pm$0.0000}} \\
credit & sex
 & 0.0420\topelement{$\pm$0.0234} & 0.0450\topelement{$\pm$0.0257} & \textbf{0.0000\topelement{$\pm$0.0000}}\dag & 0.0770\topelement{$\pm$0.0312} & 0.0730\topelement{$\pm$0.0312} & 0.0620\topelement{$\pm$0.0147} & 0.0380\topelement{$\pm$0.0194} \\
& age
 & 0.0420\topelement{$\pm$0.0234} & 0.0360\topelement{$\pm$0.0196} & \textbf{0.0000\topelement{$\pm$0.0000}}\dag & 0.0770\topelement{$\pm$0.0312} & 0.0730\topelement{$\pm$0.0312} & 0.0620\topelement{$\pm$0.0147} & 0.0380\topelement{$\pm$0.0194} \\
income & race
 & 0.0011\topelement{$\pm$0.0006}\dag & 0.0006\topelement{$\pm$0.0002}\dag & \textbf{0.0000\topelement{$\pm$0.0000}}\dag & 0.0671\topelement{$\pm$0.0038} & 0.0738\topelement{$\pm$0.0073} & 0.0733\topelement{$\pm$0.0042}\ddag & 0.0644\topelement{$\pm$0.0032} \\
& sex
 & 0.0011\topelement{$\pm$0.0006}\dag & 0.0013\topelement{$\pm$0.0005}\dag & \textbf{0.0000\topelement{$\pm$0.0000}}\dag & 0.0671\topelement{$\pm$0.0038} & 0.0738\topelement{$\pm$0.0073} & 0.0733\topelement{$\pm$0.0042}\ddag & 0.0644\topelement{$\pm$0.0032} \\
ppr & sex
 & 0.0735\topelement{$\pm$0.0193}\dag & 0.1153\topelement{$\pm$0.0129}\dag & \textbf{0.0000\topelement{$\pm$0.0000}}\dag & 0.2114\topelement{$\pm$0.0238} & 0.2148\topelement{$\pm$0.0135} & 0.2211\topelement{$\pm$0.0112} & 0.2000\topelement{$\pm$0.0165} \\
& race
 & 0.0735\topelement{$\pm$0.0193}\dag & 0.1321\topelement{$\pm$0.0218}\dag & \textbf{0.0000\topelement{$\pm$0.0000}}\dag & 0.2114\topelement{$\pm$0.0238} & 0.2148\topelement{$\pm$0.0135} & 0.2211\topelement{$\pm$0.0112} & 0.2000\topelement{$\pm$0.0165} \\
ppvr & sex
 & 0.0275\topelement{$\pm$0.0110} & 0.0282\topelement{$\pm$0.0111} & \textbf{0.0000\topelement{$\pm$0.0000}}\dag & 0.0699\topelement{$\pm$0.0092}\ddag & 0.0659\topelement{$\pm$0.0105}\ddag & 0.0642\topelement{$\pm$0.0091}\ddag & 0.0407\topelement{$\pm$0.0060} \\
& race
 & 0.0275\topelement{$\pm$0.0110} & 0.0275\topelement{$\pm$0.0110} & \textbf{0.0000\topelement{$\pm$0.0000}}\dag & 0.0699\topelement{$\pm$0.0092}\ddag & 0.0659\topelement{$\pm$0.0105}\ddag & 0.0642\topelement{$\pm$0.0091}\ddag & 0.0407\topelement{$\pm$0.0060} \\
\midrule
\multicolumn{2}{c}{W/T/L} & 0/5/4 & 0/5/4 & 0/1/8 & 2/7/0 & 2/7/0 & 4/5/0 & --- \\
\multicolumn{2}{c}{avg.rank} & 2.72 & 2.94 & \textbf{1.33} & 5.78 & 6.00 & 5.56 & 3.67 \\
\bottomrule
\end{tabular}
}}
\\ \vspace{-1em}
\subfloat[]{\label{tab:aware,fair,a}
\scalebox{.54}{
\begin{tabular}{rr rrrrrrr}
\toprule
Dataset & $\text{Att}_\text{sen}$ & \multicolumn{1}{c}{lightGBM} & \multicolumn{1}{c}{FairGBM} & \multicolumn{1}{c}{AdaFair} & \multicolumn{1}{c}{Bagging} & \multicolumn{1}{c}{\emph{EPAF-C}} & \multicolumn{1}{c}{\emph{EPAF-D}} & \multicolumn{1}{c}{\emph{POAF}} \\
\midrule
ricci & race
 & \textbf{0.3093\topelement{$\pm$0.1497}} & \textbf{0.3093\topelement{$\pm$0.1497}} & 0.3150\topelement{$\pm$0.1368} & 0.3150\topelement{$\pm$0.1368} & 0.3150\topelement{$\pm$0.1368} & 0.3150\topelement{$\pm$0.1368} & 0.3293\topelement{$\pm$0.1237} \\
credit & sex
 & 0.0602\topelement{$\pm$0.0794} & 0.0556\topelement{$\pm$0.0702} & \textbf{0.0000\topelement{$\pm$0.0000}}\dag & 0.0807\topelement{$\pm$0.0762} & 0.0739\topelement{$\pm$0.0519} & 0.0664\topelement{$\pm$0.0538} & 0.0406\topelement{$\pm$0.0270} \\
& age
 & 0.1291\topelement{$\pm$0.0890} & 0.1130\topelement{$\pm$0.0932} & \textbf{0.0000\topelement{$\pm$0.0000}}\dag & 0.1762\topelement{$\pm$0.0612}\ddag & 0.1148\topelement{$\pm$0.0856} & 0.1006\topelement{$\pm$0.0646} & 0.0588\topelement{$\pm$0.0428} \\
income & race
 & 0.0738\topelement{$\pm$0.0078}\dag & 0.0730\topelement{$\pm$0.0068}\dag & \textbf{0.0000\topelement{$\pm$0.0000}}\dag & 0.0962\topelement{$\pm$0.0086} & 0.0958\topelement{$\pm$0.0140} & 0.0956\topelement{$\pm$0.0159} & 0.0932\topelement{$\pm$0.0137} \\
& sex
 & 0.1689\topelement{$\pm$0.0066} & 0.1594\topelement{$\pm$0.0067}\dag & \textbf{0.0000\topelement{$\pm$0.0000}}\dag & 0.1899\topelement{$\pm$0.0112} & 0.1924\topelement{$\pm$0.0101} & 0.1910\topelement{$\pm$0.0099} & 0.1757\topelement{$\pm$0.0104} \\
ppr & sex
 & 0.1646\topelement{$\pm$0.0202}\ddag & 0.0853\topelement{$\pm$0.0391} & \textbf{0.0000\topelement{$\pm$0.0000}}\dag & 0.1506\topelement{$\pm$0.0332} & 0.1490\topelement{$\pm$0.0374} & 0.1450\topelement{$\pm$0.0355} & 0.1186\topelement{$\pm$0.0136} \\
& race
 & 0.2029\topelement{$\pm$0.0482}\ddag & 0.1091\topelement{$\pm$0.0481} & \textbf{0.0000\topelement{$\pm$0.0000}}\dag & 0.1359\topelement{$\pm$0.0324}\ddag & 0.1419\topelement{$\pm$0.0302}\ddag & 0.1331\topelement{$\pm$0.0249} & 0.1189\topelement{$\pm$0.0370} \\
ppvr & sex
 & 0.0440\topelement{$\pm$0.0147} & 0.0450\topelement{$\pm$0.0133} & 0.0436\topelement{$\pm$0.0289} & 0.0683\topelement{$\pm$0.0151}\ddag & 0.0521\topelement{$\pm$0.0161}\ddag & 0.0600\topelement{$\pm$0.0088}\ddag & \textbf{0.0342\topelement{$\pm$0.0190}} \\
& race
 & \textbf{0.0357\topelement{$\pm$0.0125}} & \textbf{0.0357\topelement{$\pm$0.0125}} & 0.0494\topelement{$\pm$0.0324} & 0.0361\topelement{$\pm$0.0175} & 0.0428\topelement{$\pm$0.0146} & 0.0448\topelement{$\pm$0.0136} & 0.0361\topelement{$\pm$0.0099} \\
\midrule
\multicolumn{2}{c}{W/T/L} & 2/6/1 & 0/7/2 & 0/3/6 & 3/6/0 & 2/7/0 & 1/8/0 & --- \\
\multicolumn{2}{c}{avg.rank} & 4.00 & 2.44 & \textbf{2.17} & 5.72 & 5.50 & 4.83 & 3.33 \\
\bottomrule
\end{tabular}
}}
\\ \vspace{-1em}
\subfloat[]{\label{tab:aware,fair,b}
\scalebox{.54}{
\begin{tabular}{rr rrrrrrr}
\toprule
Dataset & $\text{Att}_\text{sen}$ & \multicolumn{1}{c}{lightGBM} & \multicolumn{1}{c}{FairGBM} & \multicolumn{1}{c}{AdaFair} & \multicolumn{1}{c}{Bagging} & \multicolumn{1}{c}{\emph{EPAF-C}} & \multicolumn{1}{c}{\emph{EPAF-D}} & \multicolumn{1}{c}{\emph{POAF}} \\
\midrule
ricci & race
 & 0.0250\topelement{$\pm$0.0500} & 0.0250\topelement{$\pm$0.0500} & 0.0250\topelement{$\pm$0.0500} & 0.0250\topelement{$\pm$0.0500} & 0.0250\topelement{$\pm$0.0500} & 0.0250\topelement{$\pm$0.0500} & \textbf{0.0000\topelement{$\pm$0.0000}} \\
credit & sex
 & 0.0646\topelement{$\pm$0.0486} & 0.0579\topelement{$\pm$0.0359} & \textbf{0.0000\topelement{$\pm$0.0000}}\dag & 0.0798\topelement{$\pm$0.0683} & 0.0983\topelement{$\pm$0.0561} & 0.0808\topelement{$\pm$0.0422} & 0.0363\topelement{$\pm$0.0261} \\
& age
 & 0.0945\topelement{$\pm$0.1027} & 0.0798\topelement{$\pm$0.0784} & \textbf{0.0000\topelement{$\pm$0.0000}}\dag & 0.1138\topelement{$\pm$0.0728} & 0.1115\topelement{$\pm$0.0669} & 0.0813\topelement{$\pm$0.0588} & 0.0649\topelement{$\pm$0.0409} \\
income & race
 & 0.0324\topelement{$\pm$0.0280} & 0.0298\topelement{$\pm$0.0252} & \textbf{0.0000\topelement{$\pm$0.0000}} & 0.0369\topelement{$\pm$0.0335} & 0.0333\topelement{$\pm$0.0375}\dag & 0.0443\topelement{$\pm$0.0423} & 0.0563\topelement{$\pm$0.0415} \\
& sex
 & 0.1097\topelement{$\pm$0.0205}\ddag & 0.0760\topelement{$\pm$0.0244} & \textbf{0.0000\topelement{$\pm$0.0000}}\dag & 0.0804\topelement{$\pm$0.0334} & 0.0788\topelement{$\pm$0.0257} & 0.0743\topelement{$\pm$0.0163} & 0.0688\topelement{$\pm$0.0307} \\
ppr & sex
 & 0.1507\topelement{$\pm$0.0197} & 0.0616\topelement{$\pm$0.0330} & \textbf{0.0000\topelement{$\pm$0.0000}}\dag & 0.1376\topelement{$\pm$0.0283} & 0.1413\topelement{$\pm$0.0335} & 0.1405\topelement{$\pm$0.0480} & 0.1120\topelement{$\pm$0.0420} \\
& race
 & 0.2202\topelement{$\pm$0.0568}\ddag & 0.1157\topelement{$\pm$0.0444} & \textbf{0.0000\topelement{$\pm$0.0000}}\dag & 0.1614\topelement{$\pm$0.0244} & 0.1585\topelement{$\pm$0.0382} & 0.1508\topelement{$\pm$0.0223} & 0.1285\topelement{$\pm$0.0404} \\
ppvr & sex
 & 0.1466\topelement{$\pm$0.0455} & 0.1466\topelement{$\pm$0.0455} & \textbf{0.0633\topelement{$\pm$0.0466}} & 0.1458\topelement{$\pm$0.0785} & 0.1260\topelement{$\pm$0.0620} & 0.1244\topelement{$\pm$0.0678} & 0.1060\topelement{$\pm$0.0506} \\
& race
 & 0.1096\topelement{$\pm$0.0457} & 0.1096\topelement{$\pm$0.0457} & 0.1197\topelement{$\pm$0.1399} & \textbf{0.0886\topelement{$\pm$0.0462}} & 0.1225\topelement{$\pm$0.0617} & 0.1159\topelement{$\pm$0.0364} & 0.1102\topelement{$\pm$0.0417} \\
\midrule
\multicolumn{2}{c}{W/T/L} & 2/7/0 & 0/9/0 & 0/4/5 & 0/9/0 & 0/8/1 & 0/9/0 & --- \\
\multicolumn{2}{c}{avg.rank} & 5.17 & 3.28 & \textbf{1.94} & 4.83 & 5.39 & 4.50 & 2.89 \\
\bottomrule
\end{tabular}
}}
\\ \vspace{-1em}
\subfloat[]{\label{tab:aware,fair,c}
\scalebox{.54}{
\begin{tabular}{rr rrrrrrr}
\toprule
Dataset & $\text{Att}_\text{sen}$ & \multicolumn{1}{c}{lightGBM} & \multicolumn{1}{c}{FairGBM} & \multicolumn{1}{c}{AdaFair} & \multicolumn{1}{c}{Bagging} & \multicolumn{1}{c}{\emph{EPAF-C}} & \multicolumn{1}{c}{\emph{EPAF-D}} & \multicolumn{1}{c}{\emph{POAF}} \\
\midrule
ricci & race
 & \textbf{0.0182\topelement{$\pm$0.0364}} & \textbf{0.0182\topelement{$\pm$0.0364}} & 0.0250\topelement{$\pm$0.0500} & 0.0250\topelement{$\pm$0.0500} & 0.0250\topelement{$\pm$0.0500} & 0.0250\topelement{$\pm$0.0500} & 0.0250\topelement{$\pm$0.0500} \\
credit & sex
 & 0.0882\topelement{$\pm$0.0545} & 0.0893\topelement{$\pm$0.0538} & 0.0754\topelement{$\pm$0.0586} & \textbf{0.0566\topelement{$\pm$0.0384}}\dag & 0.0845\topelement{$\pm$0.0521} & 0.0660\topelement{$\pm$0.0479} & 0.0875\topelement{$\pm$0.0331} \\
& age
 & 0.1354\topelement{$\pm$0.0995} & 0.1351\topelement{$\pm$0.1036} & 0.1385\topelement{$\pm$0.1243} & 0.1340\topelement{$\pm$0.0896} & \textbf{0.1010\topelement{$\pm$0.1090}} & 0.1154\topelement{$\pm$0.1154} & 0.1146\topelement{$\pm$0.0817} \\
income & race
 & 0.0615\topelement{$\pm$0.0301} & 0.0591\topelement{$\pm$0.0299} & 0.1052\topelement{$\pm$0.0109}\ddag & 0.0390\topelement{$\pm$0.0156} & \textbf{0.0356\topelement{$\pm$0.0343}} & 0.0419\topelement{$\pm$0.0364} & 0.0454\topelement{$\pm$0.0302} \\
& sex
 & 0.0367\topelement{$\pm$0.0271}\ddag & 0.0274\topelement{$\pm$0.0161} & 0.2003\topelement{$\pm$0.0052}\ddag & 0.0189\topelement{$\pm$0.0140} & 0.0319\topelement{$\pm$0.0108}\ddag & 0.0282\topelement{$\pm$0.0156} & \textbf{0.0095\topelement{$\pm$0.0082}} \\
ppr & sex
 & \textbf{0.0681\topelement{$\pm$0.0449}} & 0.1122\topelement{$\pm$0.0717} & 0.1284\topelement{$\pm$0.0263} & 0.0955\topelement{$\pm$0.0502} & 0.1097\topelement{$\pm$0.0474} & 0.1172\topelement{$\pm$0.0541} & 0.1093\topelement{$\pm$0.0471} \\
& race
 & \textbf{0.0426\topelement{$\pm$0.0379}}\dag & 0.0930\topelement{$\pm$0.0422} & 0.0977\topelement{$\pm$0.0099} & 0.1057\topelement{$\pm$0.0465} & 0.0947\topelement{$\pm$0.0410} & 0.0991\topelement{$\pm$0.0461} & 0.0866\topelement{$\pm$0.0313} \\
ppvr & sex
 & 0.5270\topelement{$\pm$0.1920} & 0.5324\topelement{$\pm$0.1819} & \textbf{0.1128\topelement{$\pm$0.0419}} & 0.2159\topelement{$\pm$0.1211} & 0.2122\topelement{$\pm$0.1322} & 0.2138\topelement{$\pm$0.1067} & 0.2870\topelement{$\pm$0.1582} \\
& race
 & 0.3218\topelement{$\pm$0.2237} & 0.3218\topelement{$\pm$0.2237} & 0.2134\topelement{$\pm$0.2030} & \textbf{0.1618\topelement{$\pm$0.1008}} & 0.1899\topelement{$\pm$0.1060} & 0.1652\topelement{$\pm$0.0976} & 0.2238\topelement{$\pm$0.1016} \\
\midrule
\multicolumn{2}{c}{W/T/L} & 1/7/1 & 0/9/0 & 2/7/0 & 0/8/1 & 1/8/0 & 0/9/0 & --- \\
\multicolumn{2}{c}{avg.rank} & 4.44 & 4.78 & 5.11 & \textbf{3.11} & 3.22 & 3.78 & 3.56 \\
\bottomrule
\end{tabular}
}}
\end{table*}

\paragraph{Datasets} 
Five public datasets that we use\footnote{%
Ricci \url{https://rdrr.io/cran/Stat2Data/man/Ricci.html}, 
Credit \url{https://archive.ics.uci.edu/dataset/144/statlog+german+credit+data}, 
Income \url{https://archive.ics.uci.edu/ml/datasets/adult}, 
PPR and PPVR (\ie{} Propublica-Recidivism, Propublica-Violent-Recidivism) \url{https://github.com/propublica/compas-analysis/}
} include Ricci, Credit, Income, PPR, and PPVR, with more details provided in Table~\ref{tab:dataset}.

\paragraph{Evaluation metrics, and baseline fairness measures}

As data imbalance usually exists within unfair datasets, we consider multiple criteria to evaluate the prediction performance from different perspectives, including accuracy, precision, recall (\aka{} sensitivity), $\mathrm{f}_1$ score, and specificity. 
Moreover, to measure the discrimination degree within classifiers as well as to evaluate the validity of \emph{\gls{dr}} in capturing the discriminative risk of classifiers, we consider three commonly-used group fairness measures (that is, \gls{dp} \citep{feldman2015certifying,gajane2017formalizing}, \gls{eopp} \citep{hardt2016equality}, and \gls{pqp} \citep{chouldechova2017fair,verma2018fairness}) and compare \gls{dr} with these three baseline fairness measures. 
Additionally, for efficiency metrics in comparison of various ensemble pruning methods, we directly compare the time cost and space cost of different methods.\looseness=-1 

\paragraph{Baseline fairness-aware ensemble-based methods, and ensemble pruning methods}
To evaluate the validity of \emph{\gls{dr}} in \cref{expt:quality} and the effectiveness of \poepabbr{} as a pruned sub-ensemble in \cref{expt:aware}, ensemble classifiers are conducted using several ensemble methods including bagging \citep{breiman1996bagging}, AdaBoost \citep{freund1996experiments,freund1995boosting}, lightGBM \citep{ke2017lightgbm}, as well as two fairness-aware ensemble-based methods (that is, FairGBM \citep{cruz2022fairgbm} and AdaFair \citep{iosifidis2019adafair}). 
Besides, to verify the proposed oracle bounds and generalisation bounds in section~\ref{method:1,2}, 
bagging, AdaBoostM1, and SAMME are used in \cref{expt:bounds} to constitute an ensemble classifier on various kinds of individual classifiers including \glspl*{dt}, naive Bayesian (NB) classifiers, $k$-nearest neighbours (KNN) classifiers, Logistic Regression (LR), support vector machines (SVM), linear SVMs (linSVM), and multilayer perceptrons (MLP). 
Furthermore, to evaluate the effectiveness of \poepabbr{} as an ensemble pruning method in \cref{expt:prus}, various ensemble pruning methods are considered as baselines, including a variety of ranking-based methods as well as optimisation-based methods. 
The used ranking-based methods include 
KL-divergence pruning (KL), Kappa pruning (KP) \citep{margineantu1997pruning}, reduce error pruning (RE) \citep{martinez2009analysis}, complementarity measure pruning (CM) \citep{martinez2004aggregation, cao2018optimizing}, orientation ordering pruning (OO) \citep{martinez2006pruning}, diversity regularised ensemble pruning (DREP) \citep{li2012diversity}, ordering-based ensemble pruning (OEP) \citep{qian2015pareto},  accuracy-based ensemble pruning (TSP.AP), and diversity-based ensemble pruning (TSP.DP) \citep{zhang2019two}. 
The used optimisation-based methods include 
single-objective ensemble pruning (SEP), Pareto ensemble pruning (PEP) \citep{qian2015pareto}, the maximal relevant minimal redundant (MRMR) method, the maximum relevancy maximum complementary (MRMC) method \citep{xia2018maximum}, and the discriminative ensemble pruning (DiscEP) \citep{cao2018optimizing}.

\begin{table*}[t]
\begin{minipage}{\textwidth}
\centering\caption{%
Comparison of pruning methods with \poepabbr{} on \gls{dr}. The column named `Ensem' is the corresponding result for the entire ensemble without pruning; The best \gls{dr} with a lower standard deviation is indicated with bold fonts for each dataset (row). (a--c) Using bagging, AdaBoostM1, and SAMME respectively to produce the original unpruned ensemble classifiers. 
}\label{tab:dt,prus,rel1}
\vspace{-1.2em}%
\toptabcol
\subfloat[]{
\scalebox{.54}{%
\begin{threeparttable}[b]
\begin{tabular}{rr rrrrrrrrr rrr}
\toprule
Dataset & \multicolumn{1}{c}{Ensem} & \multicolumn{1}{c}{KP} & \multicolumn{1}{c}{OO} & \multicolumn{1}{c}{DREP} & \multicolumn{1}{c}{SEP} & \multicolumn{1}{c}{OEP} & \multicolumn{1}{c}{PEP} & \multicolumn{1}{c}{MRMC} & \multicolumn{1}{c}{Disc.EP} & \multicolumn{1}{c}{TSP.AP} & \multicolumn{1}{c}{\greedy} & \multicolumn{1}{c}{\ddismi} & \multicolumn{1}{c}{\poepabbr} \\
\midrule
ricci &
\textbf{0.0000\topelement{$\pm$0.0000}} & \textbf{0.0000\topelement{$\pm$0.0000}} & \textbf{0.0000\topelement{$\pm$0.0000}} & \textbf{0.0000\topelement{$\pm$0.0000}} & \textbf{0.0000\topelement{$\pm$0.0000}} & \textbf{0.0000\topelement{$\pm$0.0000}} & \textbf{0.0000\topelement{$\pm$0.0000}} & \textbf{0.0000\topelement{$\pm$0.0000}} & \textbf{0.0000\topelement{$\pm$0.0000}} & \textbf{0.0000\topelement{$\pm$0.0000}} & \textbf{0.0000\topelement{$\pm$0.0000}} & \textbf{0.0000\topelement{$\pm$0.0000}} & \textbf{0.0000\topelement{$\pm$0.0000}} \\
credit &
0.0450\topelement{$\pm$0.0215} & 0.0610\topelement{$\pm$0.0243}\ddag & 0.0550\topelement{$\pm$0.0215} & 0.0390\topelement{$\pm$0.0185} & 0.0430\topelement{$\pm$0.0217} & 0.0410\topelement{$\pm$0.0156} & 0.0350\topelement{$\pm$0.0235} & 0.0500\topelement{$\pm$0.0276} & 0.0590\topelement{$\pm$0.0311} & 0.0400\topelement{$\pm$0.0187} & 0.0460\topelement{$\pm$0.0195} & 0.0450\topelement{$\pm$0.0200} & \textbf{0.0330\topelement{$\pm$0.0130}} \\
income &
0.0453\topelement{$\pm$0.0053}\ddag & 0.0507\topelement{$\pm$0.0054}\ddag & 0.0460\topelement{$\pm$0.0056}\ddag & 0.0535\topelement{$\pm$0.0030}\ddag & 0.0484\topelement{$\pm$0.0050}\ddag & 0.0453\topelement{$\pm$0.0053}\ddag & 0.0453\topelement{$\pm$0.0053}\ddag & 0.0502\topelement{$\pm$0.0094}\ddag & 0.0498\topelement{$\pm$0.0041}\ddag & 0.0444\topelement{$\pm$0.0054}\ddag & 0.0481\topelement{$\pm$0.0042}\ddag & 0.0471\topelement{$\pm$0.0041}\ddag & \textbf{0.0358\topelement{$\pm$0.0058}} \\
ppr &
0.1367\topelement{$\pm$0.0132} & 0.1539\topelement{$\pm$0.0143}\ddag & 0.1399\topelement{$\pm$0.0111}\ddag & 0.1427\topelement{$\pm$0.0127}\ddag & 0.1489\topelement{$\pm$0.0167}\ddag & 0.1367\topelement{$\pm$0.0132} & 0.1367\topelement{$\pm$0.0132} & 0.1406\topelement{$\pm$0.0070}\ddag & 0.1451\topelement{$\pm$0.0100}\ddag & 0.1380\topelement{$\pm$0.0121}\ddag & 0.1370\topelement{$\pm$0.0222} & 0.1391\topelement{$\pm$0.0205} & \textbf{0.1209\topelement{$\pm$0.0111}} \\
ppvr &
0.0507\topelement{$\pm$0.0073}\ddag & 0.0522\topelement{$\pm$0.0066}\ddag & 0.0492\topelement{$\pm$0.0117}\ddag & 0.0639\topelement{$\pm$0.0111}\ddag & 0.0534\topelement{$\pm$0.0121}\ddag & 0.0507\topelement{$\pm$0.0073}\ddag & 0.0514\topelement{$\pm$0.0046}\ddag & 0.0532\topelement{$\pm$0.0097}\ddag & 0.0549\topelement{$\pm$0.0119}\ddag & 0.0427\topelement{$\pm$0.0090}\ddag & 0.0494\topelement{$\pm$0.0084}\ddag & 0.0427\topelement{$\pm$0.0082}\ddag & \textbf{0.0275\topelement{$\pm$0.0056}} \\
\midrule
W/T/L & 2/3/0 & 4/1/0 & 3/2/0 & 3/2/0 & 3/2/0 & 2/3/0 & 2/3/0 & 3/2/0 & 3/2/0 & 3/2/0 & 2/3/0 & 2/3/0 & --- \\
avg.rank & 5.6 & 10.8 & 7.2 & 9.2 & 9.0 & 5.1 & 4.8 & 9.4 & 10.4 & 4.4 & 6.8 & 6.1 & \textbf{2.2} \\
\bottomrule
\end{tabular}
\begin{tablenotes}\large
\item[1] The reported results are the average values of each method and the corresponding standard deviation under 5-fold cross-validation on each dataset. 
\item[2] By two-tailed paired $t$-test at 5\% significance level, $\ddagger$ and $\dagger$ denote that the performance of \poepabbr{} is superior to and inferior to that of the comparative baseline method,  respectively. 
\item[3] The last two rows show the results of $t$-test and average rank, respectively. 
The `W/T/L' in $t$-test indicates the numbers that \poepabbr{} is superior to, not significantly different from, or inferior to the corresponding comparative pruning methods. The average rank is calculated according to the Friedman test~\cite{demvsar2006statistical}. 
\end{tablenotes}
\end{threeparttable}
}}
\\ \vspace{-1em}
\subfloat[]{
\scalebox{.54}{
\begin{tabular}{rr rrrrrrrrr rrr}
\toprule
Dataset & \multicolumn{1}{c}{Ensem} & \multicolumn{1}{c}{KP} & \multicolumn{1}{c}{OO} & \multicolumn{1}{c}{DREP} & \multicolumn{1}{c}{SEP} & \multicolumn{1}{c}{OEP} & \multicolumn{1}{c}{PEP} & \multicolumn{1}{c}{MRMC} & \multicolumn{1}{c}{Disc.EP} & \multicolumn{1}{c}{TSP.AP} & \multicolumn{1}{c}{\greedy} & \multicolumn{1}{c}{\ddismi} & \multicolumn{1}{c}{\poepabbr} \\
\midrule
ricci &
\textbf{0.0000\topelement{$\pm$0.0000}} & \textbf{0.0000\topelement{$\pm$0.0000}} & \textbf{0.0000\topelement{$\pm$0.0000}} & \textbf{0.0000\topelement{$\pm$0.0000}} & \textbf{0.0000\topelement{$\pm$0.0000}} & \textbf{0.0000\topelement{$\pm$0.0000}} & \textbf{0.0000\topelement{$\pm$0.0000}} & \textbf{0.0000\topelement{$\pm$0.0000}} & \textbf{0.0000\topelement{$\pm$0.0000}} & \textbf{0.0000\topelement{$\pm$0.0000}} & \textbf{0.0000\topelement{$\pm$0.0000}} & \textbf{0.0000\topelement{$\pm$0.0000}} & \textbf{0.0000\topelement{$\pm$0.0000}} \\
credit &
0.0610\topelement{$\pm$0.0263} & 0.0650\topelement{$\pm$0.0374} & 0.0650\topelement{$\pm$0.0166} & 0.0820\topelement{$\pm$0.0451} & 0.0680\topelement{$\pm$0.0091} & 0.0600\topelement{$\pm$0.0221} & 0.0710\topelement{$\pm$0.0331}\ddag & 0.0540\topelement{$\pm$0.0383} & 0.0830\topelement{$\pm$0.0076}\ddag & 0.0820\topelement{$\pm$0.0303}\ddag & 0.0580\topelement{$\pm$0.0325} & 0.0440\topelement{$\pm$0.0167} & \textbf{0.0380\topelement{$\pm$0.0305}} \\
income &
0.0592\topelement{$\pm$0.0026} & 0.0608\topelement{$\pm$0.0041} & 0.0596\topelement{$\pm$0.0025} & 0.0645\topelement{$\pm$0.0133} & 0.0683\topelement{$\pm$0.0107} & 0.0609\topelement{$\pm$0.0048} & 0.0608\topelement{$\pm$0.0041} & 0.0608\topelement{$\pm$0.0045} & 0.0608\topelement{$\pm$0.0041} & 0.0598\topelement{$\pm$0.0022} & 0.0602\topelement{$\pm$0.0022} & 0.0660\topelement{$\pm$0.0121} & \textbf{0.0523\topelement{$\pm$0.0060}} \\
ppr &
0.1607\topelement{$\pm$0.0188} & 0.1836\topelement{$\pm$0.0065}\ddag & 0.1685\topelement{$\pm$0.0175} & 0.1636\topelement{$\pm$0.0216} & 0.1805\topelement{$\pm$0.0267} & 0.1703\topelement{$\pm$0.0169} & 0.1703\topelement{$\pm$0.0169} & 0.1812\topelement{$\pm$0.0134}\ddag & 0.1849\topelement{$\pm$0.0107}\ddag & 0.1682\topelement{$\pm$0.0171} & 0.1708\topelement{$\pm$0.0128}\ddag & 0.1843\topelement{$\pm$0.0185}\ddag & \textbf{0.1490\topelement{$\pm$0.0094}} \\
ppvr &
0.0474\topelement{$\pm$0.0095} & 0.0494\topelement{$\pm$0.0104} & 0.0494\topelement{$\pm$0.0104} & 0.0519\topelement{$\pm$0.0075}\ddag & 0.0592\topelement{$\pm$0.0175} & 0.0497\topelement{$\pm$0.0103} & 0.0494\topelement{$\pm$0.0104} & 0.0459\topelement{$\pm$0.0067} & 0.0494\topelement{$\pm$0.0104} & 0.0494\topelement{$\pm$0.0053}\ddag & 0.0472\topelement{$\pm$0.0051}\ddag & 0.0582\topelement{$\pm$0.0098}\ddag & \textbf{0.0432\topelement{$\pm$0.0049}} \\
\midrule
W/T/L & 0/5/0 & 1/4/0 & 0/5/0 & 1/4/0 & 0/5/0 & 0/5/0 & 1/4/0 & 1/4/0 & 2/3/0 & 2/3/0 & 2/3/0 & 2/3/0 & --- \\
avg.rank & 4.2 & 8.1 & 5.9 & 8.7 & 10.2 & 7.7 & 7.7 & 5.6 & 9.6 & 6.7 & 5.4 & 9.0 & \textbf{2.2} \\
\bottomrule
\end{tabular}
}}
\\ \vspace{-1em}
\subfloat[]{
\scalebox{.54}{
\begin{tabular}{rr rrrrrrrrr rrr}
\toprule
Dataset & \multicolumn{1}{c}{Ensem} & \multicolumn{1}{c}{KP} & \multicolumn{1}{c}{OO} & \multicolumn{1}{c}{DREP} & \multicolumn{1}{c}{SEP} & \multicolumn{1}{c}{OEP} & \multicolumn{1}{c}{PEP} & \multicolumn{1}{c}{MRMC} & \multicolumn{1}{c}{Disc.EP} & \multicolumn{1}{c}{TSP.AP} & \multicolumn{1}{c}{\greedy} & \multicolumn{1}{c}{\ddismi} & \multicolumn{1}{c}{\poepabbr} \\
\midrule
ricci &
\textbf{0.0000\topelement{$\pm$0.0000}} & \textbf{0.0000\topelement{$\pm$0.0000}} & \textbf{0.0000\topelement{$\pm$0.0000}} & \textbf{0.0000\topelement{$\pm$0.0000}} & \textbf{0.0000\topelement{$\pm$0.0000}} & \textbf{0.0000\topelement{$\pm$0.0000}} & \textbf{0.0000\topelement{$\pm$0.0000}} & \textbf{0.0000\topelement{$\pm$0.0000}} & \textbf{0.0000\topelement{$\pm$0.0000}} & \textbf{0.0000\topelement{$\pm$0.0000}} & \textbf{0.0000\topelement{$\pm$0.0000}} & \textbf{0.0000\topelement{$\pm$0.0000}} & \textbf{0.0000\topelement{$\pm$0.0000}} \\
credit &
0.0600\topelement{$\pm$0.0247} & 0.0770\topelement{$\pm$0.0211}\ddag & 0.0610\topelement{$\pm$0.0233} & 0.0720\topelement{$\pm$0.0284} & 0.0750\topelement{$\pm$0.0240} & 0.0580\topelement{$\pm$0.0275} & 0.0790\topelement{$\pm$0.0352} & 0.0880\topelement{$\pm$0.0442}\ddag & 0.0900\topelement{$\pm$0.0203}\ddag & 0.0800\topelement{$\pm$0.0348} & 0.0530\topelement{$\pm$0.0164} & 0.0450\topelement{$\pm$0.0122} & \textbf{0.0300\topelement{$\pm$0.0209}} \\
income &
0.0572\topelement{$\pm$0.0070} & 0.0560\topelement{$\pm$0.0059} & 0.0585\topelement{$\pm$0.0083} & 0.0562\topelement{$\pm$0.0073} & 0.0630\topelement{$\pm$0.0096}\ddag & 0.0593\topelement{$\pm$0.0087} & 0.0565\topelement{$\pm$0.0059} & 0.0593\topelement{$\pm$0.0087} & 0.0560\topelement{$\pm$0.0059} & 0.0580\topelement{$\pm$0.0084} & 0.0576\topelement{$\pm$0.0089} & 0.0664\topelement{$\pm$0.0138}\ddag & \textbf{0.0524\topelement{$\pm$0.0039}} \\
ppr &
0.1547\topelement{$\pm$0.0112} & 0.1797\topelement{$\pm$0.0108}\ddag & 0.1591\topelement{$\pm$0.0078} & 0.1529\topelement{$\pm$0.0270} & 0.1773\topelement{$\pm$0.0147}\ddag & 0.1606\topelement{$\pm$0.0092} & 0.1578\topelement{$\pm$0.0094} & 0.1761\topelement{$\pm$0.0144}\ddag & 0.1838\topelement{$\pm$0.0128} & 0.1584\topelement{$\pm$0.0062} & 0.1630\topelement{$\pm$0.0149} & 0.1750\topelement{$\pm$0.0232}\ddag & \textbf{0.1477\topelement{$\pm$0.0312}} \\
ppvr &
0.0487\topelement{$\pm$0.0089} & 0.0512\topelement{$\pm$0.0090} & 0.0512\topelement{$\pm$0.0090} & 0.0707\topelement{$\pm$0.0229} & 0.0549\topelement{$\pm$0.0201} & 0.0512\topelement{$\pm$0.0090} & 0.0519\topelement{$\pm$0.0096} & 0.0512\topelement{$\pm$0.0140} & 0.0512\topelement{$\pm$0.0090} & 0.0594\topelement{$\pm$0.0270} & 0.0574\topelement{$\pm$0.0170} & 0.0499\topelement{$\pm$0.0176} & \textbf{0.0424\topelement{$\pm$0.0188}} \\
\midrule
W/T/L & 0/5/0 & 2/3/0 & 0/5/0 & 0/5/0 & 2/3/0 & 0/5/0 & 0/5/0 & 2/3/0 & 1/4/0 & 0/5/0 & 0/5/0 & 2/3/0 & --- \\
avg.rank & 4.6 & 7.3 & 6.8 & 6.6 & 9.6 & 6.9 & 7.0 & 9.1 & 8.3 & 8.6 & 7.2 & 6.8 & \textbf{2.2} \\
\bottomrule
\end{tabular}
}}
\end{minipage}
\begin{minipage}{\textwidth}
\centering\caption{%
Comparison of pruning methods with \poepabbr{} on the test accuracy (\%). The column named `Ensem' is the corresponding result for the entire ensemble without pruning; The best accuracy with a lower standard deviation is indicated with bold fonts for each dataset (row). (a--c) Using bagging, AdaBoostM1, and SAMME to produce homogeneous ensemble classifiers.  
}\label{tab:dt,prus,rel6}
\vspace{-1.2em}%
\subfloat[]{
\scalebox{.52}{%
\begin{tabular}{rr rrrrrrrrr rrr}
\toprule
Dataset & \multicolumn{1}{c}{Ensem} & \multicolumn{1}{c}{KP} & \multicolumn{1}{c}{OO} & \multicolumn{1}{c}{DREP} & \multicolumn{1}{c}{SEP} & \multicolumn{1}{c}{OEP} & \multicolumn{1}{c}{PEP} & \multicolumn{1}{c}{MRMC} & \multicolumn{1}{c}{Disc.EP} & \multicolumn{1}{c}{TSP.AP} & \multicolumn{1}{c}{\greedy} & \multicolumn{1}{c}{\ddismi} & \multicolumn{1}{c}{\poepabbr} \\
\midrule
ricci &
\textbf{99.13\topelement{$\pm$1.94}} & \textbf{99.13\topelement{$\pm$1.94}} & \textbf{99.13\topelement{$\pm$1.94}} & \textbf{99.13\topelement{$\pm$1.94}} & \textbf{99.13\topelement{$\pm$1.94}} & \textbf{99.13\topelement{$\pm$1.94}} & \textbf{99.13\topelement{$\pm$1.94}} & \textbf{99.13\topelement{$\pm$1.94}} & \textbf{99.13\topelement{$\pm$1.94}} & \textbf{99.13\topelement{$\pm$1.94}} & \textbf{99.13\topelement{$\pm$1.94}} & \textbf{99.13\topelement{$\pm$1.94}} & \textbf{99.13\topelement{$\pm$1.94}} \\
credit &
73.90\topelement{$\pm$1.19} & 73.10\topelement{$\pm$3.23} & 72.70\topelement{$\pm$2.46} & 74.10\topelement{$\pm$2.72} & 73.70\topelement{$\pm$2.02} & 73.90\topelement{$\pm$1.19} & 74.10\topelement{$\pm$1.14} & 73.20\topelement{$\pm$3.38} & 73.10\topelement{$\pm$2.90} & \textbf{74.30\topelement{$\pm$3.42}} & 71.50\topelement{$\pm$0.94} & 72.70\topelement{$\pm$2.36} & 74.10\topelement{$\pm$3.07} \\
income &
\textbf{83.82\topelement{$\pm$0.44}}\dag & 83.24\topelement{$\pm$0.57} & 83.41\topelement{$\pm$0.63} & 82.64\topelement{$\pm$0.95} & 83.61\topelement{$\pm$0.34}\dag & \textbf{83.82\topelement{$\pm$0.44}}\dag & \textbf{83.82\topelement{$\pm$0.44}}\dag & 83.28\topelement{$\pm$0.40} & 83.14\topelement{$\pm$0.49} & 83.79\topelement{$\pm$0.48}\dag & 83.20\topelement{$\pm$0.52} & 83.17\topelement{$\pm$0.44} & 83.04\topelement{$\pm$0.48} \\
ppr &
\textbf{64.68\topelement{$\pm$0.36}}\dag & 63.96\topelement{$\pm$0.83}\dag & 64.42\topelement{$\pm$0.87}\dag & 62.47\topelement{$\pm$1.46} & 64.24\topelement{$\pm$1.37}\dag & \textbf{64.68\topelement{$\pm$0.36}}\dag & \textbf{64.68\topelement{$\pm$0.36}}\dag & 64.04\topelement{$\pm$1.06}\dag & 63.62\topelement{$\pm$0.87} & 64.51\topelement{$\pm$1.14}\dag & 63.77\topelement{$\pm$1.05} & 63.75\topelement{$\pm$0.99} & 62.74\topelement{$\pm$1.13} \\
ppvr &
81.17\topelement{$\pm$0.88} & 80.95\topelement{$\pm$0.94} & 81.37\topelement{$\pm$1.75} & 79.40\topelement{$\pm$2.19} & 80.65\topelement{$\pm$1.40}\ddag & 81.17\topelement{$\pm$0.88} & 81.37\topelement{$\pm$0.92} & 80.35\topelement{$\pm$0.97}\ddag & 80.45\topelement{$\pm$1.57} & 81.65\topelement{$\pm$0.83} & 81.15\topelement{$\pm$0.80} & 80.97\topelement{$\pm$0.96} & \textbf{82.15\topelement{$\pm$1.03}} \\
\midrule
W/T/L & 0/3/2 & 0/4/1 & 0/4/1 & 0/5/0 & 1/2/2 & 0/3/2 & 0/3/2 & 1/3/1 & 0/5/0 & 0/3/2 & 0/5/0 & 0/5/0 & --- \\
avg.rank & 4.4 & 8.3 & 6.6 & 9.8 & 7.0 & 4.4 & \textbf{3.6} & 8.2 & 9.9 & \textbf{3.6} & 9.0 & 9.2 & 7.0 \\
\bottomrule
\end{tabular}
}}
\\ \vspace{-1em}
\subfloat[]{
\scalebox{.52}{
\begin{tabular}{rr rrrrrrrrr rrr}
\toprule
Dataset & \multicolumn{1}{c}{Ensem} & \multicolumn{1}{c}{KP} & \multicolumn{1}{c}{OO} & \multicolumn{1}{c}{DREP} & \multicolumn{1}{c}{SEP} & \multicolumn{1}{c}{OEP} & \multicolumn{1}{c}{PEP} & \multicolumn{1}{c}{MRMC} & \multicolumn{1}{c}{Disc.EP} & \multicolumn{1}{c}{TSP.AP} & \multicolumn{1}{c}{\greedy} & \multicolumn{1}{c}{\ddismi} & \multicolumn{1}{c}{\poepabbr} \\
\midrule
ricci &
\textbf{100.00\topelement{$\pm$0.00}} & \textbf{100.00\topelement{$\pm$0.00}} & \textbf{100.00\topelement{$\pm$0.00}} & \textbf{100.00\topelement{$\pm$0.00}} & \textbf{100.00\topelement{$\pm$0.00}} & \textbf{100.00\topelement{$\pm$0.00}} & \textbf{100.00\topelement{$\pm$0.00}} & \textbf{100.00\topelement{$\pm$0.00}} & \textbf{100.00\topelement{$\pm$0.00}} & \textbf{100.00\topelement{$\pm$0.00}} & \textbf{100.00\topelement{$\pm$0.00}} & \textbf{100.00\topelement{$\pm$0.00}} & \textbf{100.00\topelement{$\pm$0.00}} \\
credit &
\textbf{71.60\topelement{$\pm$1.78}}\dag & 69.50\topelement{$\pm$2.62} & 71.00\topelement{$\pm$1.94} & 70.00\topelement{$\pm$3.26} & 68.50\topelement{$\pm$1.58} & 71.50\topelement{$\pm$2.47}\dag & 70.40\topelement{$\pm$2.53} & 69.80\topelement{$\pm$2.33} & 69.60\topelement{$\pm$1.08}\dag & 70.30\topelement{$\pm$3.55} & 71.40\topelement{$\pm$2.01}\dag & 70.10\topelement{$\pm$1.08}\dag & 67.50\topelement{$\pm$3.89} \\
income &
\textbf{83.32\topelement{$\pm$0.28}}\dag & 83.29\topelement{$\pm$0.28}\dag & 83.14\topelement{$\pm$0.27}\dag & 80.96\topelement{$\pm$0.47} & 82.07\topelement{$\pm$1.57} & 83.05\topelement{$\pm$0.40}\dag & 83.29\topelement{$\pm$0.28}\dag & 82.91\topelement{$\pm$0.44}\dag & 83.29\topelement{$\pm$0.28}\dag & 83.11\topelement{$\pm$0.23}\dag & 83.11\topelement{$\pm$0.23}\dag & 80.65\topelement{$\pm$0.82} & 80.51\topelement{$\pm$2.05} \\
ppr &
\textbf{63.30\topelement{$\pm$1.54}} & 62.79\topelement{$\pm$1.77} & 62.86\topelement{$\pm$0.97} & 61.59\topelement{$\pm$2.40} & 60.96\topelement{$\pm$1.44} & 62.74\topelement{$\pm$1.37} & 62.74\topelement{$\pm$1.37} & 62.87\topelement{$\pm$1.37}\dag & 62.01\topelement{$\pm$2.04} & 62.86\topelement{$\pm$1.23} & 62.86\topelement{$\pm$1.60}\dag & 60.71\topelement{$\pm$2.00} & 58.49\topelement{$\pm$3.41} \\
ppvr &
\textbf{81.57\topelement{$\pm$0.46}} & 81.40\topelement{$\pm$0.51} & 81.40\topelement{$\pm$0.51} & 78.55\topelement{$\pm$0.91}\ddag & 79.58\topelement{$\pm$1.75} & 81.52\topelement{$\pm$0.59} & 81.55\topelement{$\pm$0.72} & 81.25\topelement{$\pm$0.74} & 81.40\topelement{$\pm$0.51} & 80.25\topelement{$\pm$1.77} & 79.75\topelement{$\pm$2.13} & 78.68\topelement{$\pm$1.97} & 80.85\topelement{$\pm$1.64} \\
\midrule
W/T/L & 0/3/2 & 0/4/1 & 0/4/1 & 1/4/0 & 0/5/0 & 0/3/2 & 0/4/1 & 0/3/2 & 0/3/2 & 0/4/1 & 0/2/3 & 0/4/1 & --- \\
avg.rank & \textbf{2.2} & 6.4 & 5.1 & 9.8 & 10.2 & 5.5 & 4.9 & 6.8 & 6.8 & 6.6 & 5.9 & 10.0 & 10.8 \\
\bottomrule
\end{tabular}
}}
\\ \vspace{-1em}
\subfloat[]{
\scalebox{.52}{
\begin{tabular}{rr rrrrrrrrr rrr}
\toprule
Dataset & \multicolumn{1}{c}{Ensem} & \multicolumn{1}{c}{KP} & \multicolumn{1}{c}{OO} & \multicolumn{1}{c}{DREP} & \multicolumn{1}{c}{SEP} & \multicolumn{1}{c}{OEP} & \multicolumn{1}{c}{PEP} & \multicolumn{1}{c}{MRMC} & \multicolumn{1}{c}{Disc.EP} & \multicolumn{1}{c}{TSP.AP} & \multicolumn{1}{c}{\greedy} & \multicolumn{1}{c}{\ddismi} & \multicolumn{1}{c}{\poepabbr} \\
\midrule
ricci &
98.26\topelement{$\pm$2.38} & 98.26\topelement{$\pm$2.38} & 98.26\topelement{$\pm$2.38} & \textbf{99.13\topelement{$\pm$1.94}} & 98.26\topelement{$\pm$2.38} & 98.26\topelement{$\pm$2.38} & 98.26\topelement{$\pm$2.38} & 98.26\topelement{$\pm$2.38} & 98.26\topelement{$\pm$2.38} & 98.26\topelement{$\pm$2.38} & \textbf{99.13\topelement{$\pm$1.94}} & 98.26\topelement{$\pm$2.38} & 98.26\topelement{$\pm$2.38} \\
credit &
\textbf{74.40\topelement{$\pm$2.33}} & 70.70\topelement{$\pm$2.17} & 70.00\topelement{$\pm$3.94} & 70.70\topelement{$\pm$2.41} & 70.50\topelement{$\pm$2.09} & 73.30\topelement{$\pm$3.19} & 73.50\topelement{$\pm$3.94} & 71.00\topelement{$\pm$1.06} & 70.30\topelement{$\pm$3.13} & 72.30\topelement{$\pm$2.17} & 70.80\topelement{$\pm$2.28} & 70.40\topelement{$\pm$1.95} & 68.70\topelement{$\pm$3.51} \\
income &
\textbf{83.04\topelement{$\pm$0.43}}\dag & 82.98\topelement{$\pm$0.43}\dag & 82.73\topelement{$\pm$0.57}\dag & 81.06\topelement{$\pm$0.91} & 81.48\topelement{$\pm$1.17} & 82.50\topelement{$\pm$0.59}\dag & 82.85\topelement{$\pm$0.52}\dag & 82.50\topelement{$\pm$0.59}\dag & 82.98\topelement{$\pm$0.43}\dag & 82.68\topelement{$\pm$0.49}\dag & 82.10\topelement{$\pm$1.31} & 79.84\topelement{$\pm$1.60} & 80.64\topelement{$\pm$1.63} \\
ppr &
\textbf{63.65\topelement{$\pm$0.85}} & 62.63\topelement{$\pm$1.09} & 63.07\topelement{$\pm$0.88} & 62.35\topelement{$\pm$0.80} & 62.89\topelement{$\pm$0.98} & 63.41\topelement{$\pm$1.15} & 63.64\topelement{$\pm$1.03} & 62.56\topelement{$\pm$1.33} & 62.65\topelement{$\pm$0.40} & 63.25\topelement{$\pm$0.77} & 62.68\topelement{$\pm$0.53} & 61.79\topelement{$\pm$0.92} & 61.23\topelement{$\pm$1.93} \\
ppvr &
\textbf{81.17\topelement{$\pm$1.03}} & 80.90\topelement{$\pm$1.09} & 80.90\topelement{$\pm$1.09} & 77.73\topelement{$\pm$1.50} & 80.72\topelement{$\pm$2.09} & 80.90\topelement{$\pm$1.09} & 81.00\topelement{$\pm$1.11} & 80.55\topelement{$\pm$1.03} & 80.90\topelement{$\pm$1.09} & 79.48\topelement{$\pm$2.55} & 79.13\topelement{$\pm$2.14}\ddag & 77.58\topelement{$\pm$1.51} & 80.22\topelement{$\pm$1.96} \\
\midrule
W/T/L & 0/4/1 & 0/4/1 & 0/4/1 & 0/5/0 & 0/5/0 & 0/4/1 & 0/4/1 & 0/4/1 & 0/4/1 & 0/4/1 & 0/5/0 & 1/4/0 & --- \\
avg.rank & \textbf{2.4} & 6.3 & 6.9 & 8.6 & 8.0 & 5.2 & 3.6 & 7.7 & 6.8 & 6.4 & 6.9 & 11.2 & 11.0 \\
\bottomrule
\end{tabular}
}}
\end{minipage}
\end{table*}

\begin{table*}[t]
\centering\caption{%
Comparison of pruning methods with \poepabbr{} on three group fairness measures, using \gls{dt} as base learners and bagging to construct homogeneous ensemble classifiers. The column `$\text{Att}_\text{sen}$' reports results for different sensitive attributes; Bold indicates the best fairness performance with lower variance, measured by the smallest disparity between privileged and marginalised groups. (a–c) Results on the test set under \gls{dp}, \gls{eopp}, and \gls{pqp}, respectively.
}\label{tab:fair,bag}
\vspace{-1.4em}%
\toptabcol
\subfloat[]{\label{tab:fair,bag,a}
\scalebox{.53}{%
\begin{tabular}{rr rrrrrrrrrr rrr}
\toprule
Dataset & $\text{Att}_\text{sen}$ & \multicolumn{1}{c}{Ensem} & \multicolumn{1}{c}{KP} & \multicolumn{1}{c}{OO} & \multicolumn{1}{c}{DREP} & \multicolumn{1}{c}{SEP} & \multicolumn{1}{c}{OEP} & \multicolumn{1}{c}{PEP} & \multicolumn{1}{c}{MRMC} & \multicolumn{1}{c}{Disc.EP} & \multicolumn{1}{c}{TSP.AP} & \multicolumn{1}{c}{\greedy} & \multicolumn{1}{c}{\ddismi} & \multicolumn{1}{c}{\poepabbr} \\
\midrule
ricci & race & 
\textbf{0.3097\topelement{$\pm$0.1530}} & \textbf{0.3097\topelement{$\pm$0.1530}} & \textbf{0.3097\topelement{$\pm$0.1530}} & \textbf{0.3097\topelement{$\pm$0.1530}} & \textbf{0.3097\topelement{$\pm$0.1530}} & \textbf{0.3097\topelement{$\pm$0.1530}} & \textbf{0.3097\topelement{$\pm$0.1530}} & \textbf{0.3097\topelement{$\pm$0.1530}} & \textbf{0.3097\topelement{$\pm$0.1530}} & \textbf{0.3097\topelement{$\pm$0.1530}} & \textbf{0.3097\topelement{$\pm$0.1530}} & \textbf{0.3097\topelement{$\pm$0.1530}} & \textbf{0.3097\topelement{$\pm$0.1530}} \\
credit & sex & 
0.0722\topelement{$\pm$0.0665} & 0.0584\topelement{$\pm$0.0426} & 0.1159\topelement{$\pm$0.0890} & \textbf{0.0414\topelement{$\pm$0.0379}} & 0.0648\topelement{$\pm$0.0427} & 0.0833\topelement{$\pm$0.0743} & 0.0815\topelement{$\pm$0.0711} & 0.0610\topelement{$\pm$0.0453} & 0.1012\topelement{$\pm$0.0497} & 0.0586\topelement{$\pm$0.0299} & 0.0970\topelement{$\pm$0.0605} & 0.0962\topelement{$\pm$0.0563} & 0.0741\topelement{$\pm$0.0581} \\
& age & 
0.0874\topelement{$\pm$0.0754} & 0.1157\topelement{$\pm$0.0785} & 0.1218\topelement{$\pm$0.0431}\ddag & 0.1075\topelement{$\pm$0.1168} & 0.0873\topelement{$\pm$0.0423} & 0.0938\topelement{$\pm$0.0931}\ddag & 0.0883\topelement{$\pm$0.0767} & 0.1266\topelement{$\pm$0.0737} & 0.1149\topelement{$\pm$0.1066} & 0.0766\topelement{$\pm$0.0814} & 0.0699\topelement{$\pm$0.0295} & \textbf{0.0489\topelement{$\pm$0.0419}} & 0.0691\topelement{$\pm$0.0226} \\
\bottomrule
\end{tabular}
}}
\\ \vspace{-1em}
\subfloat[]{\label{tab:fair,bag,b}
\scalebox{.53}{
\begin{tabular}{rr rrrrrrrrrr rrr}
\toprule
Dataset & $\text{Att}_\text{sen}$ & \multicolumn{1}{c}{Ensem} & \multicolumn{1}{c}{KP} & \multicolumn{1}{c}{OO} & \multicolumn{1}{c}{DREP} & \multicolumn{1}{c}{SEP} & \multicolumn{1}{c}{OEP} & \multicolumn{1}{c}{PEP} & \multicolumn{1}{c}{MRMC} & \multicolumn{1}{c}{Disc.EP} & \multicolumn{1}{c}{TSP.AP} & \multicolumn{1}{c}{\greedy} & \multicolumn{1}{c}{\ddismi} & \multicolumn{1}{c}{\poepabbr} \\
\midrule
ricci & race &
\textbf{0.0000\topelement{$\pm$0.0000}} & \textbf{0.0000\topelement{$\pm$0.0000}} & \textbf{0.0000\topelement{$\pm$0.0000}} & \textbf{0.0000\topelement{$\pm$0.0000}} & \textbf{0.0000\topelement{$\pm$0.0000}} & \textbf{0.0000\topelement{$\pm$0.0000}} & \textbf{0.0000\topelement{$\pm$0.0000}} & \textbf{0.0000\topelement{$\pm$0.0000}} & \textbf{0.0000\topelement{$\pm$0.0000}} & \textbf{0.0000\topelement{$\pm$0.0000}} & \textbf{0.0000\topelement{$\pm$0.0000}} & \textbf{0.0000\topelement{$\pm$0.0000}} & \textbf{0.0000\topelement{$\pm$0.0000}} \\
credit & sex &
0.0986\topelement{$\pm$0.0550} & 0.0646\topelement{$\pm$0.0413} & 0.1174\topelement{$\pm$0.0895} & \textbf{0.0201\topelement{$\pm$0.0061}} & 0.0802\topelement{$\pm$0.0401} & 0.1000\topelement{$\pm$0.0592} & 0.1006\topelement{$\pm$0.0599} & 0.0903\topelement{$\pm$0.0296} & 0.0740\topelement{$\pm$0.0404} & 0.0424\topelement{$\pm$0.0332} & 0.0711\topelement{$\pm$0.0468} & 0.0913\topelement{$\pm$0.0429} & 0.0748\topelement{$\pm$0.0603} \\
& age &
0.1217\topelement{$\pm$0.0619} & 0.1636\topelement{$\pm$0.0659} & 0.1221\topelement{$\pm$0.0439}\ddag & 0.1026\topelement{$\pm$0.0894} & 0.0802\topelement{$\pm$0.0852} & 0.1392\topelement{$\pm$0.0835} & 0.1275\topelement{$\pm$0.0757} & 0.1213\topelement{$\pm$0.0744}\ddag & 0.1428\topelement{$\pm$0.1531} & 0.1302\topelement{$\pm$0.0796} & 0.0764\topelement{$\pm$0.0303}\ddag & 0.0652\topelement{$\pm$0.0273} & \textbf{0.0506\topelement{$\pm$0.0350}} \\
\bottomrule
\end{tabular}
}}
\\ \vspace{-1em}
\subfloat[]{\label{tab:fair,bag,c}
\scalebox{.53}{
\begin{tabular}{rr rrrrrrrrrr rrr}
\toprule
Dataset & $\text{Att}_\text{sen}$ & \multicolumn{1}{c}{Ensem} & \multicolumn{1}{c}{KP} & \multicolumn{1}{c}{OO} & \multicolumn{1}{c}{DREP} & \multicolumn{1}{c}{SEP} & \multicolumn{1}{c}{OEP} & \multicolumn{1}{c}{PEP} & \multicolumn{1}{c}{MRMC} & \multicolumn{1}{c}{Disc.EP} & \multicolumn{1}{c}{TSP.AP} & \multicolumn{1}{c}{\greedy} & \multicolumn{1}{c}{\ddismi} & \multicolumn{1}{c}{\poepabbr} \\
\midrule
ricci & race &
\textbf{0.0250\topelement{$\pm$0.0559}} & \textbf{0.0250\topelement{$\pm$0.0559}} & \textbf{0.0250\topelement{$\pm$0.0559}} & \textbf{0.0250\topelement{$\pm$0.0559}} & \textbf{0.0250\topelement{$\pm$0.0559}} & \textbf{0.0250\topelement{$\pm$0.0559}} & \textbf{0.0250\topelement{$\pm$0.0559}} & \textbf{0.0250\topelement{$\pm$0.0559}} & \textbf{0.0250\topelement{$\pm$0.0559}} & \textbf{0.0250\topelement{$\pm$0.0559}} & \textbf{0.0250\topelement{$\pm$0.0559}} & \textbf{0.0250\topelement{$\pm$0.0559}} & \textbf{0.0250\topelement{$\pm$0.0559}} \\
credit & sex &
0.0727\topelement{$\pm$0.0309} & 0.0784\topelement{$\pm$0.0668} & 0.0636\topelement{$\pm$0.0609} & 0.0578\topelement{$\pm$0.0531} & 0.0872\topelement{$\pm$0.0454} & 0.0638\topelement{$\pm$0.0382} & 0.0629\topelement{$\pm$0.0462} & 0.0837\topelement{$\pm$0.0328} & \textbf{0.0417\topelement{$\pm$0.0370}} & 0.0645\topelement{$\pm$0.0478} & 0.0497\topelement{$\pm$0.0772} & 0.0650\topelement{$\pm$0.0745} & 0.0728\topelement{$\pm$0.0447} \\
& age &
0.1508\topelement{$\pm$0.0834} & 0.1557\topelement{$\pm$0.1059} & 0.1243\topelement{$\pm$0.0744} & \textbf{0.1179\topelement{$\pm$0.0960}} & 0.1530\topelement{$\pm$0.0442} & 0.1626\topelement{$\pm$0.0858}\ddag & 0.1550\topelement{$\pm$0.0894} & 0.1271\topelement{$\pm$0.0698} & 0.1546\topelement{$\pm$0.0759} & 0.1637\topelement{$\pm$0.0779} & 0.1439\topelement{$\pm$0.0799} & 0.1511\topelement{$\pm$0.0861} & 0.1467\topelement{$\pm$0.0882} \\
\bottomrule
\end{tabular}
}}
\end{table*}
\begin{table*}[t]
\centering\caption{%
Comparison of pruning methods with \poepabbr{} on three group fairness measures, using \gls{dt} as base learners and AdaBoostM1 to construct homogeneous ensembles. Notation follows Table~\ref{tab:fair,bag}. (a–c) Test-set results under \gls{dp}, \gls{eopp}, and \gls{pqp}. 
}\label{tab:fair,am1}
\vspace{-1.4em}
\toptabcol%
\subfloat[]{\label{tab:fair,am1,a}
\scalebox{.53}{%
\begin{tabular}{rr rrrrrrrrrr rrr}
\toprule
Dataset & $\text{Att}_\text{sen}$ & \multicolumn{1}{c}{Ensem} & \multicolumn{1}{c}{KP} & \multicolumn{1}{c}{OO} & \multicolumn{1}{c}{DREP} & \multicolumn{1}{c}{SEP} & \multicolumn{1}{c}{OEP} & \multicolumn{1}{c}{PEP} & \multicolumn{1}{c}{MRMC} & \multicolumn{1}{c}{Disc.EP} & \multicolumn{1}{c}{TSP.AP} & \multicolumn{1}{c}{\greedy} & \multicolumn{1}{c}{\ddismi} & \multicolumn{1}{c}{\poepabbr} \\
\midrule
ricci & race &
\textbf{0.3240\topelement{$\pm$0.2295}} & \textbf{0.3240\topelement{$\pm$0.2295}} & \textbf{0.3240\topelement{$\pm$0.2295}} & \textbf{0.3240\topelement{$\pm$0.2295}} & \textbf{0.3240\topelement{$\pm$0.2295}} & \textbf{0.3240\topelement{$\pm$0.2295}} & \textbf{0.3240\topelement{$\pm$0.2295}} & \textbf{0.3240\topelement{$\pm$0.2295}} & \textbf{0.3240\topelement{$\pm$0.2295}} & \textbf{0.3240\topelement{$\pm$0.2295}} & \textbf{0.3240\topelement{$\pm$0.2295}} & \textbf{0.3240\topelement{$\pm$0.2295}} & \textbf{0.3240\topelement{$\pm$0.2295}} \\
credit & sex &
0.0773\topelement{$\pm$0.0774} & 0.0622\topelement{$\pm$0.0462} & 0.1048\topelement{$\pm$0.0678} & 0.0849\topelement{$\pm$0.0566} & 0.0914\topelement{$\pm$0.0531} & 0.0753\topelement{$\pm$0.0846} & 0.0958\topelement{$\pm$0.0850} & \textbf{0.0446\topelement{$\pm$0.0322}} & 0.0771\topelement{$\pm$0.0619} & 0.0893\topelement{$\pm$0.0404} & 0.0706\topelement{$\pm$0.0320} & 0.0501\topelement{$\pm$0.0471} & 0.0830\topelement{$\pm$0.0428} \\
& age &
0.0879\topelement{$\pm$0.0900} & 0.0921\topelement{$\pm$0.0387} & 0.1048\topelement{$\pm$0.0767} & 0.1486\topelement{$\pm$0.1453} & 0.0640\topelement{$\pm$0.0458} & 0.0948\topelement{$\pm$0.0708} & 0.1063\topelement{$\pm$0.0840} & 0.1207\topelement{$\pm$0.0988} & 0.1173\topelement{$\pm$0.1392} & 0.0924\topelement{$\pm$0.0612} & 0.1286\topelement{$\pm$0.0638} & 0.0793\topelement{$\pm$0.0405} & \textbf{0.0629\topelement{$\pm$0.0647}} \\
\bottomrule
\end{tabular}
}}
\\ \vspace{-1em}
\subfloat[]{\label{tab:fair,am1,b}
\scalebox{.53}{
\begin{tabular}{rr rrrrrrrrrr rrr}
\toprule
Dataset & $\text{Att}_\text{sen}$ & \multicolumn{1}{c}{Ensem} & \multicolumn{1}{c}{KP} & \multicolumn{1}{c}{OO} & \multicolumn{1}{c}{DREP} & \multicolumn{1}{c}{SEP} & \multicolumn{1}{c}{OEP} & \multicolumn{1}{c}{PEP} & \multicolumn{1}{c}{MRMC} & \multicolumn{1}{c}{Disc.EP} & \multicolumn{1}{c}{TSP.AP} & \multicolumn{1}{c}{\greedy} & \multicolumn{1}{c}{\ddismi} & \multicolumn{1}{c}{\poepabbr} \\
\midrule
ricci & race &
\textbf{0.2000\topelement{$\pm$0.4472}} & \textbf{0.2000\topelement{$\pm$0.4472}} & \textbf{0.2000\topelement{$\pm$0.4472}} & \textbf{0.2000\topelement{$\pm$0.4472}} & \textbf{0.2000\topelement{$\pm$0.4472}} & \textbf{0.2000\topelement{$\pm$0.4472}} & \textbf{0.2000\topelement{$\pm$0.4472}} & \textbf{0.2000\topelement{$\pm$0.4472}} & \textbf{0.2000\topelement{$\pm$0.4472}} & \textbf{0.2000\topelement{$\pm$0.4472}} & \textbf{0.2000\topelement{$\pm$0.4472}} & \textbf{0.2000\topelement{$\pm$0.4472}} & \textbf{0.2000\topelement{$\pm$0.4472}} \\
credit & sex &
0.0749\topelement{$\pm$0.0407} & 0.0559\topelement{$\pm$0.0407} & 0.0553\topelement{$\pm$0.0411} & 0.0631\topelement{$\pm$0.0377} & 0.0720\topelement{$\pm$0.0525} & \textbf{0.0390\topelement{$\pm$0.0331}} & 0.0779\topelement{$\pm$0.0378} & 0.0665\topelement{$\pm$0.0224} & 0.0700\topelement{$\pm$0.0552} & 0.0504\topelement{$\pm$0.0360} & 0.0688\topelement{$\pm$0.0292} & 0.0659\topelement{$\pm$0.0343} & 0.0743\topelement{$\pm$0.0708} \\
& age &
0.0881\topelement{$\pm$0.0577} & 0.0752\topelement{$\pm$0.0804} & 0.1183\topelement{$\pm$0.0680} & 0.1205\topelement{$\pm$0.1357} & 0.0667\topelement{$\pm$0.0586} & 0.0941\topelement{$\pm$0.0686} & 0.0887\topelement{$\pm$0.0662} & 0.1035\topelement{$\pm$0.1028} & 0.1278\topelement{$\pm$0.1234} & 0.0727\topelement{$\pm$0.0532} & 0.1261\topelement{$\pm$0.0702} & \textbf{0.0588\topelement{$\pm$0.0441}} & 0.0697\topelement{$\pm$0.0489} \\
\bottomrule
\end{tabular}
}}
\\ \vspace{-1em}
\subfloat[]{\label{tab:fair,am1,c}
\scalebox{.53}{
\begin{tabular}{rr rrrrrrrrrr rrr}
\toprule
Dataset & $\text{Att}_\text{sen}$ & \multicolumn{1}{c}{Ensem} & \multicolumn{1}{c}{KP} & \multicolumn{1}{c}{OO} & \multicolumn{1}{c}{DREP} & \multicolumn{1}{c}{SEP} & \multicolumn{1}{c}{OEP} & \multicolumn{1}{c}{PEP} & \multicolumn{1}{c}{MRMC} & \multicolumn{1}{c}{Disc.EP} & \multicolumn{1}{c}{TSP.AP} & \multicolumn{1}{c}{\greedy} & \multicolumn{1}{c}{\ddismi} & \multicolumn{1}{c}{\poepabbr} \\
\midrule
ricci & race &
\textbf{0.2000\topelement{$\pm$0.4472}} & \textbf{0.2000\topelement{$\pm$0.4472}} & \textbf{0.2000\topelement{$\pm$0.4472}} & \textbf{0.2000\topelement{$\pm$0.4472}} & \textbf{0.2000\topelement{$\pm$0.4472}} & \textbf{0.2000\topelement{$\pm$0.4472}} & \textbf{0.2000\topelement{$\pm$0.4472}} & \textbf{0.2000\topelement{$\pm$0.4472}} & \textbf{0.2000\topelement{$\pm$0.4472}} & \textbf{0.2000\topelement{$\pm$0.4472}} & \textbf{0.2000\topelement{$\pm$0.4472}} & \textbf{0.2000\topelement{$\pm$0.4472}} & \textbf{0.2000\topelement{$\pm$0.4472}} \\
credit & sex &
0.0736\topelement{$\pm$0.0589} & 0.0800\topelement{$\pm$0.0446} & 0.0634\topelement{$\pm$0.0273} & 0.0688\topelement{$\pm$0.0597} & 0.0662\topelement{$\pm$0.0657} & 0.0722\topelement{$\pm$0.0366} & 0.0791\topelement{$\pm$0.0578} & 0.0742\topelement{$\pm$0.0697} & 0.0655\topelement{$\pm$0.0503} & 0.0640\topelement{$\pm$0.0464} & \textbf{0.0568\topelement{$\pm$0.0523}} & 0.0602\topelement{$\pm$0.0554} & 0.0603\topelement{$\pm$0.0670} \\
& age &
0.1211\topelement{$\pm$0.0584} & 0.1148\topelement{$\pm$0.0504} & 0.1026\topelement{$\pm$0.0612} & \textbf{0.0924\topelement{$\pm$0.0716}} & 0.1232\topelement{$\pm$0.0350} & 0.1288\topelement{$\pm$0.0725} & 0.1088\topelement{$\pm$0.0275} & 0.1206\topelement{$\pm$0.0418} & 0.1120\topelement{$\pm$0.0704} & 0.1168\topelement{$\pm$0.0324} & 0.1027\topelement{$\pm$0.0534} & 0.0968\topelement{$\pm$0.0534}\dag & 0.1334\topelement{$\pm$0.0454} \\
\bottomrule
\end{tabular}
}}
\end{table*}
\begin{table*}
\centering\caption{%
Comparison of pruning methods with \poepabbr{} on three group fairness measures, using \gls{dt} as base learners and SAMME to construct homogeneous ensemble classifiers. Notation follows Table~\ref{tab:fair,bag}. (a–c) Test-set results under \gls{dp}, \gls{eopp}, and \gls{pqp}. 
}\label{tab:fair,sam}
\vspace{-1.4em}
\toptabcol%
\subfloat[]{\label{tab:fair,sam,a}
\scalebox{.53}{%
\begin{tabular}{rr rrrrrrrrrr rrr}
\toprule
Dataset & $\text{Att}_\text{sen}$ & \multicolumn{1}{c}{Ensem} & \multicolumn{1}{c}{KP} & \multicolumn{1}{c}{OO} & \multicolumn{1}{c}{DREP} & \multicolumn{1}{c}{SEP} & \multicolumn{1}{c}{OEP} & \multicolumn{1}{c}{PEP} & \multicolumn{1}{c}{MRMC} & \multicolumn{1}{c}{Disc.EP} & \multicolumn{1}{c}{TSP.AP} & \multicolumn{1}{c}{\greedy} & \multicolumn{1}{c}{\ddismi} & \multicolumn{1}{c}{\poepabbr} \\
\midrule
ricci & race &
0.3153\topelement{$\pm$0.1663} & 0.3153\topelement{$\pm$0.1663} & 0.3153\topelement{$\pm$0.1663} & \textbf{0.3028\topelement{$\pm$0.1847}} & 0.3153\topelement{$\pm$0.1663} & 0.3153\topelement{$\pm$0.1663} & 0.3153\topelement{$\pm$0.1663} & 0.3153\topelement{$\pm$0.1663} & 0.3153\topelement{$\pm$0.1663} & 0.3153\topelement{$\pm$0.1663} & \textbf{0.3028\topelement{$\pm$0.1847}} & 0.3153\topelement{$\pm$0.1663} & 0.3153\topelement{$\pm$0.1663} \\
credit & sex &
0.0565\topelement{$\pm$0.0334} & 0.0876\topelement{$\pm$0.0227}\ddag & 0.0750\topelement{$\pm$0.0333} & 0.0657\topelement{$\pm$0.0146} & 0.1074\topelement{$\pm$0.0324} & 0.0741\topelement{$\pm$0.0266} & 0.0801\topelement{$\pm$0.0398} & 0.0538\topelement{$\pm$0.0510} & 0.0776\topelement{$\pm$0.0264} & 0.0855\topelement{$\pm$0.0493} & 0.0885\topelement{$\pm$0.0278} & 0.0680\topelement{$\pm$0.0260} & \textbf{0.0429\topelement{$\pm$0.0362}} \\
& age &
0.1218\topelement{$\pm$0.0424} & \textbf{0.0730\topelement{$\pm$0.0434}} & 0.1489\topelement{$\pm$0.0397} & 0.1120\topelement{$\pm$0.0598} & 0.1035\topelement{$\pm$0.0386} & 0.1242\topelement{$\pm$0.0468} & 0.1514\topelement{$\pm$0.0512} & 0.0906\topelement{$\pm$0.0577} & 0.2012\topelement{$\pm$0.1312}\ddag & 0.1837\topelement{$\pm$0.0708} & 0.1890\topelement{$\pm$0.0511} & 0.1660\topelement{$\pm$0.0389} & 0.1162\topelement{$\pm$0.0789} \\
\bottomrule
\end{tabular}
}}
\\ \vspace{-1em}
\subfloat[]{\label{tab:fair,sam,b}
\scalebox{.53}{
\begin{tabular}{rr rrrrrrrrrr rrr}
\toprule
Dataset & $\text{Att}_\text{sen}$ & \multicolumn{1}{c}{Ensem} & \multicolumn{1}{c}{KP} & \multicolumn{1}{c}{OO} & \multicolumn{1}{c}{DREP} & \multicolumn{1}{c}{SEP} & \multicolumn{1}{c}{OEP} & \multicolumn{1}{c}{PEP} & \multicolumn{1}{c}{MRMC} & \multicolumn{1}{c}{Disc.EP} & \multicolumn{1}{c}{TSP.AP} & \multicolumn{1}{c}{\greedy} & \multicolumn{1}{c}{\ddismi} & \multicolumn{1}{c}{\poepabbr} \\
\midrule
ricci & race &
\textbf{0.0250\topelement{$\pm$0.0559}} & \textbf{0.0250\topelement{$\pm$0.0559}} & \textbf{0.0250\topelement{$\pm$0.0559}} & \textbf{0.0250\topelement{$\pm$0.0559}} & \textbf{0.0250\topelement{$\pm$0.0559}} & \textbf{0.0250\topelement{$\pm$0.0559}} & \textbf{0.0250\topelement{$\pm$0.0559}} & \textbf{0.0250\topelement{$\pm$0.0559}} & \textbf{0.0250\topelement{$\pm$0.0559}} & \textbf{0.0250\topelement{$\pm$0.0559}} & \textbf{0.0250\topelement{$\pm$0.0559}} & \textbf{0.0250\topelement{$\pm$0.0559}} & \textbf{0.0250\topelement{$\pm$0.0559}} \\
credit & sex &
0.0514\topelement{$\pm$0.0160} & 0.0708\topelement{$\pm$0.0622} & 0.1054\topelement{$\pm$0.0472} & \textbf{0.0474\topelement{$\pm$0.0309}} & 0.0688\topelement{$\pm$0.0395} & 0.0814\topelement{$\pm$0.0448} & 0.0909\topelement{$\pm$0.0415} & 0.0785\topelement{$\pm$0.0303} & 0.0625\topelement{$\pm$0.0267} & 0.0654\topelement{$\pm$0.0438} & 0.0891\topelement{$\pm$0.0504} & 0.0636\topelement{$\pm$0.0221} & 0.0678\topelement{$\pm$0.0668} \\
& age &
0.0877\topelement{$\pm$0.0399} & 0.0765\topelement{$\pm$0.0349} & 0.1223\topelement{$\pm$0.0938} & 0.0945\topelement{$\pm$0.0721} & \textbf{0.0628\topelement{$\pm$0.0422}} & 0.0765\topelement{$\pm$0.0397} & 0.0853\topelement{$\pm$0.0710} & 0.0948\topelement{$\pm$0.0831} & 0.2432\topelement{$\pm$0.1287} & 0.1477\topelement{$\pm$0.0296} & 0.1359\topelement{$\pm$0.0847} & 0.1556\topelement{$\pm$0.0504} & 0.1325\topelement{$\pm$0.1069} \\
\bottomrule
\end{tabular}
}}
\\ \vspace{-1em}
\subfloat[]{\label{tab:fair,sam,c}
\scalebox{.53}{
\begin{tabular}{rr rrrrrrrrrr rrr}
\toprule
Dataset & $\text{Att}_\text{sen}$ & \multicolumn{1}{c}{Ensem} & \multicolumn{1}{c}{KP} & \multicolumn{1}{c}{OO} & \multicolumn{1}{c}{DREP} & \multicolumn{1}{c}{SEP} & \multicolumn{1}{c}{OEP} & \multicolumn{1}{c}{PEP} & \multicolumn{1}{c}{MRMC} & \multicolumn{1}{c}{Disc.EP} & \multicolumn{1}{c}{TSP.AP} & \multicolumn{1}{c}{\greedy} & \multicolumn{1}{c}{\ddismi} & \multicolumn{1}{c}{\poepabbr} \\
\midrule
ricci & race &
0.0222\topelement{$\pm$0.0497} & 0.0222\topelement{$\pm$0.0497} & 0.0222\topelement{$\pm$0.0497} & \textbf{0.0000\topelement{$\pm$0.0000}} & 0.0222\topelement{$\pm$0.0497} & 0.0222\topelement{$\pm$0.0497} & 0.0222\topelement{$\pm$0.0497} & 0.0222\topelement{$\pm$0.0497} & 0.0222\topelement{$\pm$0.0497} & 0.0222\topelement{$\pm$0.0497} & \textbf{0.0000\topelement{$\pm$0.0000}} & 0.0222\topelement{$\pm$0.0497} & 0.0222\topelement{$\pm$0.0497} \\
credit & sex &
0.0641\topelement{$\pm$0.0338} & 0.0751\topelement{$\pm$0.0522} & 0.0795\topelement{$\pm$0.0510} & 0.0676\topelement{$\pm$0.0526} & \textbf{0.0240\topelement{$\pm$0.0168}} & 0.0736\topelement{$\pm$0.0425} & 0.0484\topelement{$\pm$0.0320} & 0.0438\topelement{$\pm$0.0306} & 0.0632\topelement{$\pm$0.0450} & 0.0868\topelement{$\pm$0.0360}\ddag & 0.0602\topelement{$\pm$0.0653} & 0.0566\topelement{$\pm$0.0624} & 0.0587\topelement{$\pm$0.0488} \\
& age &
0.1581\topelement{$\pm$0.0971} & 0.1910\topelement{$\pm$0.0769} & 0.2142\topelement{$\pm$0.0783} & 0.1590\topelement{$\pm$0.0734} & \textbf{0.1520\topelement{$\pm$0.1215}} & 0.1593\topelement{$\pm$0.1057} & 0.1576\topelement{$\pm$0.0768} & 0.1716\topelement{$\pm$0.1017} & 0.2075\topelement{$\pm$0.0936} & 0.1877\topelement{$\pm$0.0887} & 0.1702\topelement{$\pm$0.0610} & 0.1824\topelement{$\pm$0.0677} & 0.1901\topelement{$\pm$0.1406} \\
\bottomrule
\end{tabular}
}}
\end{table*}

\paragraph{Implementation details} 
We use standard five-fold cross-validation in these experiments, that is to say, 
in each iteration, the entire dataset is split into two parts, with 80\% as the training set and 20\% as the test set. 
For baselines, an ensemble will be trained on the training set and then evaluated on the test set; For the proposed pruning method, an ensemble will be trained and pruned on the training set and then evaluated on the test set. 
Note that GPU is not necessarily required, and CPU would be sufficient to run our experiments. All experiments were run on our lab servers using a Docker image built from the Ubuntu (Linux/amd64) platform with installed miniconda3.

\section{Supplementary Empirical Results}
\label{appx:more}
\subsection{Additional results in \cref{expt:aware}}
\label{appx,expt:aware}
In this subsection, we report more results in Tables~\ref{tab:aware,acc;part1} and \ref{tab:aware,acc;part2} to validate the effectiveness of \poepabbr{}.

To be specific, Table~\ref{tab:aware,acc;part1}\subref{tab:aware,acc,a} 
contains the average test accuracy of each method and the corresponding standard deviation under 5-fold cross-validation on each dataset concerning each sensitive attribute; 
Table~\ref{tab:aware,acc;part1}\subref{tab:aware,acc,b} and Tables~\ref{tab:aware,acc;part2}\subref{tab:aware,acc,c} to \ref{tab:aware,acc;part2}\subref{tab:aware,fair,c} 
also contain the average values of $\text{f}_1$ score, \emph{\gls{dr}}, \gls{dp}, \gls{eopp}, and \gls{pqp} alongside the corresponding standard deviation on the test set accordingly.  
For instance, each row (one sensitive attribute of one of the datasets) in Table~\ref{tab:aware,acc;part2}\subref{tab:aware,acc,c} compares the test \gls{dr} with the same type of individual classifiers (that is, decision trees in this case), indicating the best results by bold fonts. 
Although \poepabbr{} is not the best among itself, AdaFair, and FairGBM, the average ranks of \poepabbr{} are still relatively high among all opponents from all criteria aspects except for the $\text{f}_1$ score. 
Furthermore, \poepabbr{} is better than the corresponding unpruned ensembles (\aka{} bagging) in most of the cases, which means that \poepabbr{} can indeed improve fairness and accuracy concurrently. 
Therefore, we can view the proposed \gls{dr} measure does its work and functions well in \poepabbr{}.\looseness=-1

\begin{figure}[t]
\begin{minipage}{.475\textwidth}
\centering 
\subfloat[]{
\includegraphics[height=24mm]{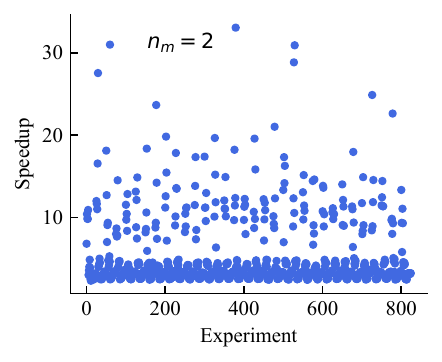}}
\hspace{3mm}
\subfloat[]{
\includegraphics[height=24mm]{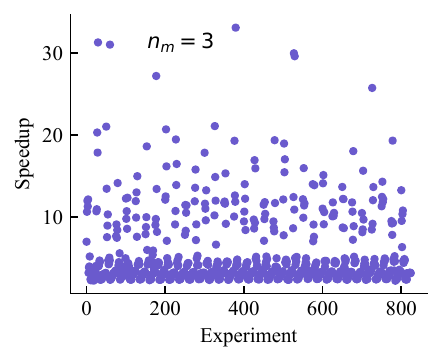}}
\\ \vspace{-4mm}
\subfloat[]{
\includegraphics[height=24mm]{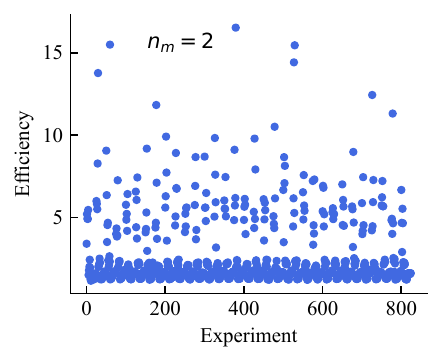}}
\hspace{3mm}
\subfloat[]{
\includegraphics[height=24mm]{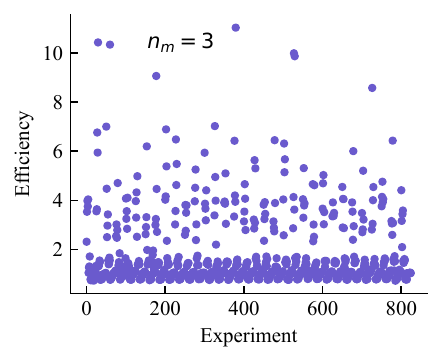}}
\vspace{-4mm}\caption{%
Comparison of speedup and efficiency between \greedy{} and \ddismi{}, where 
(a--b) indicates speedup, and (c--d) efficiency, with two and three machines, respectively. 
}\label{appx:fig:parl}
\end{minipage}
\end{figure}
\begin{figure*}[t]
\begin{minipage}{\textwidth}
\centering%
\subfloat[]{\label{fig:param,a}
\includegraphics[height=2.2cm]{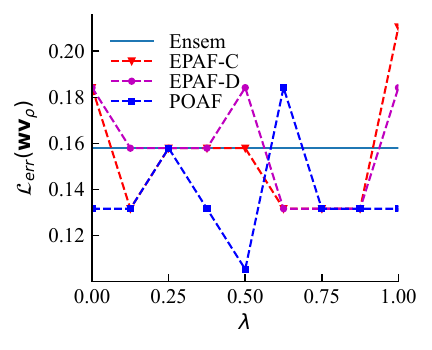}}
\subfloat[]{\label{fig:param,b}
\includegraphics[height=2.2cm]{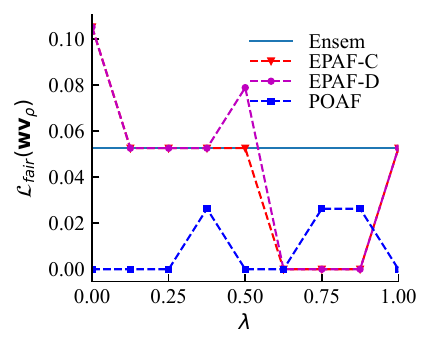}}
\subfloat[]{\label{fig:param,c}
\includegraphics[height=2.2cm]{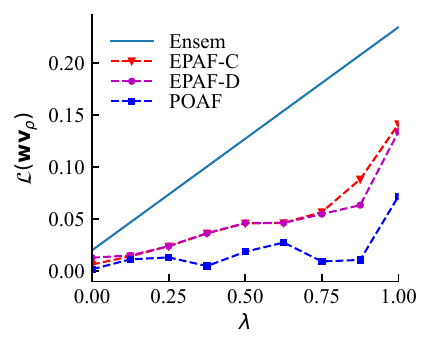}}
\subfloat[]{\label{fig:param,d}
\includegraphics[height=2.2cm]{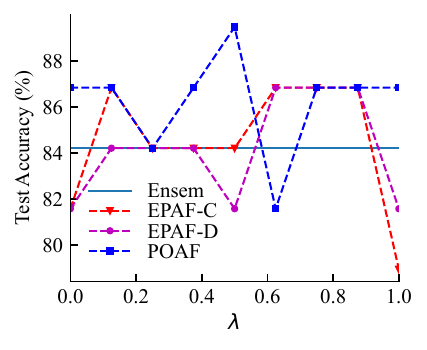}}
\\ \vspace{-1em}
\subfloat[]{\label{fig:param,e}
\includegraphics[height=2.2cm]{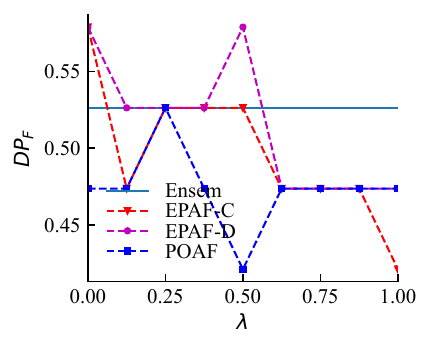}}
\subfloat[]{\label{fig:param,f}
\includegraphics[height=2.2cm]{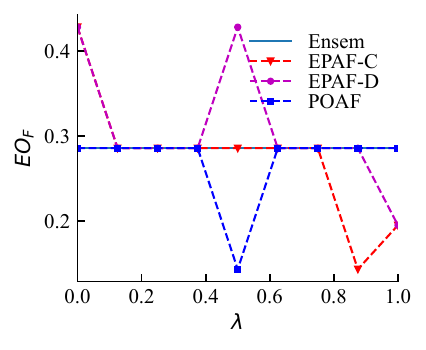}}
\subfloat[]{\label{fig:param,g}
\includegraphics[height=2.2cm]{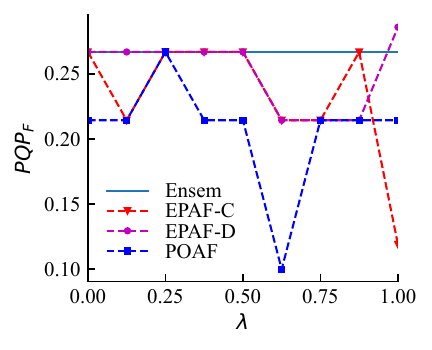}}
\subfloat[]{\label{fig:param,h}
\includegraphics[height=2.2cm]{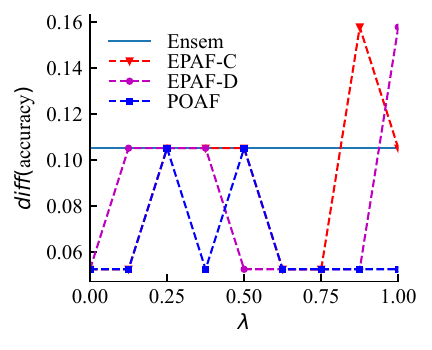}}
\vspace{-4mm}\caption{%
Effect of $\lambda$ value on the Ricci dataset using bagging with MLPs as individual classifiers. 
(a--c) Two sub-objectives used in the domination relationship (that is, $\justAccT(\mvrho)$ and $\justLt(\mvrho)$) and the objective function $\justObjT(\mvrho)$ in \eqref{eq:11}, respectively; the smaller the better. 
(d) Test accuracy (\%); the larger the better. 
(e--h) The test discrepancy between the privileged group and marginalised groups using three group fairness measures (\ie{} \gls{dp}, \gls{eopp}, and \gls{pqp}) and the test accuracy, respectively, wherein the closer the numerical value is to zero, the better result it shows.
}\label{fig:params}
\end{minipage}
\end{figure*}

\begin{figure}[t]
\begin{minipage}{.475\textwidth}
\centering
\subfloat[]{\label{subfig:us,1}
\includegraphics[scale=.45]{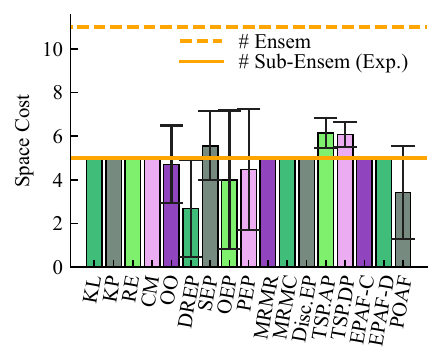}}
\hspace{1.5em}%
\subfloat[]{\label{subfig:us,2}
\includegraphics[scale=.45]{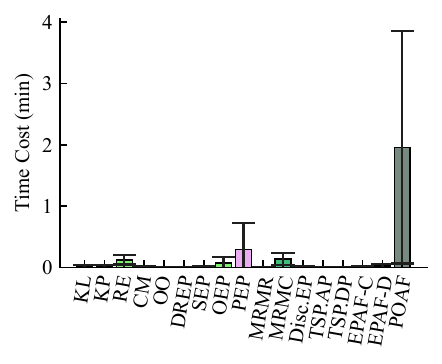}}
\vspace{-4mm}\caption{%
Comparison of pruning methods with \poepabbr{}, where (a--b) is the space and time cost. Note that the dashed and solid lines indicate the size of the original ensemble and the expected size of the pruned sub-ensemble, respectively.
}\label{fig:sota,us}
\end{minipage}
\begin{minipage}{.475\textwidth}
\centering 
\includegraphics[width=.8147\linewidth]{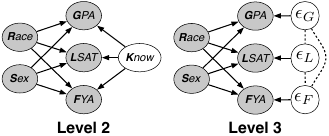}
\vspace{-3mm}
\caption{Fair prediction models \citep{kusner2017counterfactual}. 
(Left) \emph{Fair K}, a causal model for the problem of predicting law school success fairly; 
(Right) \emph{Fair add}, density plots of predicted $\mathrm{FYA}_a$ and $\mathrm{FYA}_{a'}$. 
}\label{fig:appx,cf}
\end{minipage}
\end{figure}

\subsection{Additional results in \cref{expt:prus}}
\label{appx,expt:prus}
Here we report more results in Tables~\ref{tab:dt,prus,rel1} to \ref{tab:fair,sam}, as well as Fig.~\ref{fig:sota,fair,am1} to \ref{fig:sens,adult}, to validate the effectiveness of \poepabbr{} as an ensemble pruning method.\looseness=-1

\paragraph{\poepabbr{} achieves the best fairness with acceptable accuracy performance}

Table~\ref{tab:dt,prus,rel6} contains the average test accuracy of each method and the corresponding standard deviation under five-fold cross-validation on each dataset; 
Table~\ref{tab:dt,prus,rel1} contains the average fairness quality of each method and the corresponding standard deviation on the test set accordingly as well. 
For instance, each row (dataset) in Table~\ref{tab:dt,prus,rel1} compares the test fairness quality with the same type of individual classifiers, indicating the best results by bold fonts. 
Besides, the fairness quality performance between two pruning methods is examined by two-tailed paired $t$-tests at a 5\% significance level to tell if they have significantly different results. 
Two methods end in a tie if there is no significant statistical difference; otherwise, the one with better values of the proposed fairness quality will win. 
The performance of each method is reported in the last two rows of Table~\ref{tab:dt,prus,rel1}, compared with \poepabbr{} in terms of the average rank and the number of datasets that \poepabbr{} has won, tied, or lost, respectively. 
As we can see in each sub-table of Table~\ref{tab:dt,prus,rel1}, the average ranks of \poepabbr{} outperform all baseline pruning methods including the original ensemble on the test fairness quality, which means that \poepabbr{} substantially achieves the best fairness performance. 
We may also notice that \poepabbr{} does not rank so high in Table~\ref{tab:dt,prus,rel6}, yet it still settles with ties in most of the cases, which means that \poepabbr{} could at least achieve acceptable accuracy performance, compared with other baseline methods. 
It is worth mentioning that the test fairness quality on the Ricci dataset in Table~\ref{tab:dt,prus,rel1} shows extraordinarily good results, of which the reason is probably that the size of Ricci is rather small.

\paragraph{\poepabbr{} achieves good fairness performance on three group fairness measures} 
We have reported Fig.~\ref{fig:past,merge} to demonstrate that \poepabbr{} can improve fairness in the ensembles without much accuracy degradation. 
Here we provide extra results in Tables~\ref{tab:fair,bag} to \ref{tab:fair,sam}, as well as Figures~\ref{fig:sota,fair,am1} to \ref{fig:sens,adult}, to demonstrate that the aforementioned observation is not transitory. 
As we may see from Table~\ref{tab:fair,bag}, it contains the average fairness values of each method on the test set and the corresponding standard deviation under 5-fold cross-validation on each sensitive attribute. 
For instance, each row (attribute of one dataset) in Table~\ref{tab:fair,bag} compares the discrepancy between the privileged group and marginalised groups of each method on the test set, using three group fairness measures in three sub-tables, respectively. 
Besides, two-tailed paired $t$-tests at a 5\% significance level are used to examine if two pruning methods have significantly different results on the group fairness measures. 
Two methods end in a tie if there is no significant statistical difference; otherwise, the one with the numerical result that is closer to zero will win, as it means that the treatment between the privileged group and marginalised groups is closer to equality. 
The performance of each method is reported in the last two rows of Table~\ref{tab:fair,bag}, compared with \poepabbr{} in terms of the average rank and the number of datasets that \poepabbr{} has won, tied, or lost, respectively. 
As shown in Tables~\ref{tab:fair,bag}, \ref{tab:fair,am1}\subref{tab:fair,am1,a}--\ref{tab:fair,am1}\subref{tab:fair,am1,b}, and \ref{tab:fair,sam}\subref{tab:fair,sam,a}--\ref{tab:fair,sam}\subref{tab:fair,sam,b}, \poepabbr{} achieves the best or nearly the best average ranking from the perspective of three group fairness measures, which indicates that \poepabbr{} could obtain fairer ensemble classifiers than most of the baseline pruning methods, not only in terms of our proposed \gls{dr}, but also in terms of \gls{dp}, \gls{eopp}, and \gls{pqp}. 
As for Tables~\ref{tab:fair,am1}\subref{tab:fair,am1,c} and \ref{tab:fair,sam}\subref{tab:fair,sam,c}, \greedy{} shows the potential to achieve fairer sub-ensembles as well in terms of \gls{pqp}.  
Besides, Fig.~\ref{fig:sota,fair,am1} shows that \poepabbr{} achieves competitive accuracy and high rankings on group fairness (especially, \gls{dp} and \gls{eopp}) compared with baselines. Similar observations are also presented in Fig.~\ref{fig:sota,fair,sam}, demonstrating that \poepabbr{} achieves the best fairness quality from the \gls{dp} and \gls{eopp} perspectives. 
Furthermore, Fig.~\ref{fig:sens,adult} depicts the discrepancy between the privileged group and marginalised groups on the Income dataset, 
indicating that \poepabbr{} works generally with the target of improving fairness despite the type of ensemble classifiers. 
To be specific, 
Fig.~\ref{fig:sens,adult} indicates that \poepabbr{} achieves the best fairness quality $\justLt(\mvrho)$ and objective $\justObjT(\mvrho)$ compared with baselines; 
Moreover, even from the perspective of other group fairness measures, \poepabbr{} still achieves the best results of fairness on each sensitive attribute in most cases.

\subsection{Effect of $\lambda$ value in \poepabbr}
\label{expt:lamb}

In this subsection, we study whether the choice of the hyper-parameter $\lambda$ will affect the effectiveness of \poepabbr{}. To be specific, we conduct experiments to investigate the effect of the regularisation factor $\lambda$ on two sub-objectives in the defined objective function of \eqref{eq:11}. In addition, we study whether \ddismi{} in Appendix~\ref{appx:alter} would work more efficiently than \greedy{}. The experimental results are reported in Figures~\ref{appx:fig:parl} to \ref{fig:sota,us}.

As we may see from Fig.~\ref{fig:params}\subref{fig:param,c}, \poepabbr{} beats both \greedy{} and \ddismi{} along with the original ensemble in comparison of the objective function \eqref{eq:11}, no matter which value is assigned to $\lambda$. 
Nevertheless, different values of $\lambda$ may directly affect the results of (sub-)ensembles in terms of sub-objectives and the discrepancy treatment between the privileged group and marginalised groups. 
Figures~\ref{fig:params}\subref{fig:param,a} and \ref{fig:params}\subref{fig:param,d} indicate that \poepabbr{} usually achieves the best accuracy performance when fairness and accuracy are considered equally important in the objective function. 
Figures~\ref{fig:params}\subref{fig:param,b}, \ref{fig:params}\subref{fig:param,e}, and \ref{fig:params}\subref{fig:param,f} depict that \poepabbr{} reaches the best fairness performance in the same situation, in terms of the proposed fairness quality (namely \emph{\gls{dr}}) in \cref{method:4} and two group fairness measures (namely \gls{dp} and \gls{eopp}).

\paragraph{\ddismi{} vs. \greedy{} over time cost}
\label{appx:expt,ut}

To verify whether \ddismi{} could achieve competitive performance quicker than \greedy{} or not, we employ different numbers of machines and compare their speedup and efficiency.\footnote{%
Speedup and efficiency are concepts in parallel programming, wherein speedup is defined as the ratio of the time cost for a serial program to that for the parallel program that accomplishes the same work. 
For a task-based parallel program, the parallel efficiency is calculated as the ratio of speedup to the number of cores. 
} %
Empirical results are reported in Fig.~\ref{appx:fig:parl}, wherein speedup is calculated as the ratio of the execution time of \greedy{} to that of \ddismi{}. 
The efficiency in Fig.~\ref{appx:fig:parl} can achieve 1, indicating that \ddismi{} scales linearly and even super-linearly in a few cases.\footnote{%
Sometimes a speedup of more than $n_m$ when using $n_m$ processors is observed in parallel computing, which is called super-linear speedup. 
} 
Additionally, the space cost between \poepabbr{} and baselines are reported in Fig.~\ref{fig:sota,us}\subref{subfig:us,1}, to depict the size of pruned sub-ensembles after using different pruning methods. 
It shows that part of the pruning methods cannot fix the size of the pruned sub-ensemble in advance, including OO, DREP, SEP, OEP, PEP, TSP.AP, TSP.DP, and \poepabbr{}.

\section{Details of the Case Study in \cref{expt:quality}}
\label{appx:cf}

We compare the proposed \emph{\gls{dr}} with counterfactual fairness using the same example (Law school success) given in the work of Kusner \etal{} \citep{kusner2017counterfactual}. 
The dataset includes the information of 21,790 law students across 163 law schools in the United States, such as their entrance exam scores (LSAT), their grade-point average (GPA) collected prior to law school, and their first-year average grade (FYA) \citep{wightman1998lsac}. 
The task is to predict if an applicant will have a high FYA, and meanwhile, the school intends to keep these predictions as fair as possible concerning an individual's race and sex.

There are four available models for comparison \citep{kusner2017counterfactual}, that is, (1) \emph{Full}, the standard technique of using all attributes to make predictions, including sensitive ones such as race and sex; (2) \emph{Unaware}, fairness through unawareness, where sensitive attributes are not used yet proxy attributes still persist; 
(3) \emph{Fair $K$} and \emph{fair add}, two models that are constructed to meet counterfactual fairness. 
Note that in the \emph{Fair K} model, a latent variable---a student's knowledge (K)---is postulated to affect GPA, LSAT, and FYA scores; 
in the \emph{Fair add} model, data are modelled using an additive error model, and independent error terms are used to make predictions. 
In other words, in the \emph{Fair K} model, the distributions are
\begin{subequations}
\topequation\small
\begin{align}
    \mathrm{GPA}&\sim \mathcal{N}(b_G +w_G^K K+ w_G^R R +w_G^S S, \sigma_G) \,,\\
    \mathrm{LSAT}&\sim \mathrm{Poisson}(\exp(b_L +w_L^K K+ w_L^R R+ w_L^S S)) \,,\\
    \mathrm{FYA}&\sim \mathcal{N}(w_F^K K+ w_F^R R+ w_F^S S, 1) \,,\\
    K&\sim \mathcal{N}(0,1) \,.
\end{align}%
\end{subequations}%
Meanwhile, the \emph{Fair add} model is expressed by 
\begin{subequations}
\topequation\small
\begin{align}
    \mathrm{GPA}&= b_G+ w_G^R R+ w_G^S S+\epsilon_G \,,\; \epsilon_G\sim p(\epsilon_G) \,,\\
    \mathrm{LSAT}&= b_L +w_L^R R+ w_L^S S+ \epsilon_L \,,\; \epsilon_L\sim p(\epsilon_L) \,,\\
    \mathrm{FYA}&= b_F +w_F^R R+ w_F^S S+ \epsilon_F \,,\; \epsilon_F\sim p(\epsilon_F) \,.
\end{align}%
\end{subequations}%
The error terms $\epsilon_G,\epsilon_L$ are estimated by first fitting two models that each use race and sex to individually predict GPA and LSAT. 
Then the residuals of each model are computed, for example, $\epsilon_G= \mathrm{GPA}-\hat{Y}_\mathrm{GPA}(R,S)$. 
These residual estimates of $\epsilon_G,\epsilon_L$ are used to predict FYA. 

The empirical results are reported in Fig.~\ref{fig:cf} and explained in \cref{expt:quality}.

%

\end{document}